\documentclass[11pt, a4paper, logo, onecolumn, copyright]{googledeepmind}

\usepackage[a4paper]{geometry}
\usepackage[authoryear, sort&compress, round]{natbib}
\usepackage{xspace}  %
\usepackage{siunitx}  %
\usepackage{amsthm, amsmath, amsfonts, amssymb}
\usepackage{placeins}
\usepackage[english]{babel}
\usepackage{cmap}
\usepackage[T1]{fontenc}
\usepackage{url}
\usepackage{titletoc}
\usepackage[title]{appendix}

\usepackage[colorlinks=true, allcolors=blue]{hyperref}
\usepackage[capitalise,noabbrev]{cleveref}
\usepackage{pifont}
\usepackage[dvipsnames]{xcolor}
\usepackage{textcomp} %
\usepackage[overload]{textcase}
\usepackage{graphicx}
\usepackage{colortbl}
\usepackage{booktabs}
\usepackage{changepage}
\usepackage{enumitem} %
\usepackage{tabularx}
\usepackage{datetime}
\usepackage{fancyhdr}  %
\usepackage{lastpage}  %
\usepackage{enumitem}
\usepackage[explicit]{titlesec}
\usepackage{bibentry}
\usepackage{mdframed}
\usepackage{caption}
\usepackage{needspace}
\usepackage{siunitx}
\usepackage{lscape}
\usepackage{float}

\newcommand{\ourmodel}{GenCast\xspace} %

\title{\ourmodel: Diffusion-based ensemble forecasting for medium-range weather} 
\correspondingauthor{pricei@google.com, matthjw@google.com, remilam@google.com, peterbattaglia@google.com}

\author[*,1]{Ilan~Price}
\author[*,1]{Alvaro~Sanchez-Gonzalez}
\author[*,1]{Ferran~Alet}
\author[1]{Tom~R.~Andersson}
\author[1]{Andrew El-Kadi}
\author[1]{Dominic~Masters}
\author[1]{Timo~Ewalds}
\author[1]{Jacklynn~Stott}
\author[1]{Shakir~Mohamed}
\author[1]{Peter~Battaglia}
\author[1]{Remi~Lam}
\author[1]{Matthew~Willson}

\affil[*]{Equal contribution}
\affil[1]{Google DeepMind}

\DeclareGraphicsExtensions{.pdf}

\newcommand{\quarterdegree}{\ang{0.25}\xspace}
\newcommand{\erafivenativedegree}{\ang{0.28125}\xspace}
\newcommand{\erafiveensnativedegree}{\ang{0.5625}\xspace}
\newcommand{\onedegree}{\ang{1}\xspace}

\newcommand{\hresfczero}{HRES-fc0\xspace}
\newcommand{\modelresolution}{\quarterdegree}

\newcommand{\gc}{GraphCast\xspace}
\newcommand{\gcens}{GraphCast-Perturbed\xspace}

\newcommand{\ngcm}{Neural GCM\xspace}

\newcommand{\varweight}{w}  %
\newcommand{\latitudeweight}{a}  %
\newcommand{\sll}{i}  %

\newcommand{\E}{\mathop{{}\mathbb{E}}}

\newcommand{\G}{G}  %

\newcommand{\numCRPStargets}{1320\xspace}

\newcommand{\winpercentage}{97.4}
\newcommand{\winpercentageafterthirtysixhours}{99.8}

\newcommand{\vll}[2]{\textsc{#1}#2}

\begin{abstract}
Weather forecasts are fundamentally uncertain, so predicting the range of probable weather scenarios is crucial for important decisions, from warning the public about hazardous weather, to planning renewable energy use. Here, we introduce \ourmodel, a probabilistic weather model with greater skill and speed than the top operational medium-range weather forecast in the world, the European Centre for Medium-Range Forecasts (ECMWF)'s ensemble forecast, ENS. Unlike traditional approaches, which are based on numerical weather prediction (NWP), \ourmodel is a machine learning weather prediction (MLWP) method, trained on decades of reanalysis data.
\ourmodel generates an ensemble of stochastic 15-day global forecasts, at 12-hour steps and \modelresolution latitude-longitude resolution, for over 80 surface and atmospheric variables, in 8~minutes. It has greater skill than ENS on \winpercentage\% of \numCRPStargets targets we evaluated, and better predicts extreme weather, tropical cyclones, and wind power production.
This work helps open the next chapter in operational weather forecasting, where critical weather-dependent decisions are made with greater accuracy and efficiency.
\end{abstract}

\begin{document}

\maketitle

\renewcommand{\thesection}{\arabic{section}}
\renewcommand{\thefigure}{\arabic{figure}}
\renewcommand{\thetable}{\arabic{table}}
\renewcommand{\theequation}{\arabic{equation}}

\section{Introduction}

Every day, people, governments, and other organisations around the world rely on accurate weather forecasts to make many key decisions---whether to carry an umbrella, when to flee an approaching tropical cyclone, planning the use of renewable energy in a power grid, or preparing for a heatwave.
But forecasts will always carry some uncertainty, because we can only partially observe the current weather, and even our best models of the weather are imperfect. The highly non-linear physics of weather means that small initial uncertainties and errors can rapidly grow into large uncertainties about the future, giving rise to the famous ``butterfly effect'' \citep{lorenz1996essence}. 
Making critical decisions often requires knowing not just a single probable scenario, but the range of possible scenarios and how likely they are to occur.

Traditional weather forecasting is based on numerical weather prediction (NWP) algorithms, which 
approximately solve the equations that model atmospheric dynamics. Deterministic NWP methods map the current estimate of the weather to a forecast of how the future weather will unfold over time.
This single forecast is only one possibility, so to model the probability distribution of different future weather scenarios \citep{palmer2006predictability,kalnay2003atmospheric}, weather agencies increasingly rely on ``ensemble forecasts'', which generate multiple NWP-based forecasts, each of which models a single possible scenario \citep{palmer2019ecmwf, roberts2023improver, ifs-manual-cy46r1-ens, yamaguchi2018introduction, Zhu2012b}. 
The European Centre for Medium Range Weather Forecasting (ECMWF)'s ENS~\citep{ifs-manual-cy46r1-ens} is the state-of-the-art NWP-based ensemble forecast within ECMWF's broader Integrated Forecast System (IFS), and will subsume their deterministic forecast, HRES, going forward~\citep{ecmwfPlansHighresolution}.
While ENS and other NWP-based ensemble forecasts are very effective at modelling the weather distribution, they are still prone to errors, are slow to run, and are time-consuming to engineer. 

\begin{figure}
  \centering
  \includegraphics[width=0.7\textwidth, trim={0 12.5cm 0 0},clip]{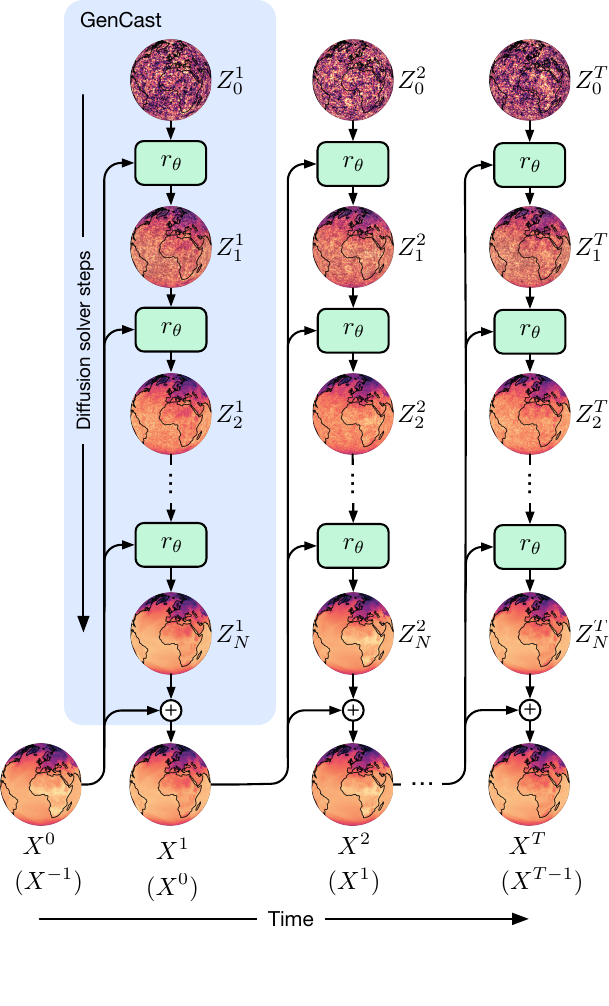}
  \caption{\textbf{Schematic of how \ourmodel produces a forecast.}
  The blue box shows how conditioning inputs, $(X^0, X^{-1})$, and an initial noise sample, $Z^1_0$, are refined by the neural network refinement function, $r_\theta$ (green box), which is parameterised by $\theta$. 
  The resulting $Z^1_1$ is the first refined candidate state, and this process repeats $N$ times. 
  The final $Z^1_N$ is then added as a residual to $X^0$, to produce the weather state at the next time step, $X^1$.
  This process then repeats, autoregressively, $T=30$ times, conditioning on $(X^{t},X^{t-1})$ and using a new initial noise sample $Z^t_0$ at each step, to produce the full weather trajectory sample (for visual clarity, we illustrate the previous state in parentheses, $(X^{t-1})$, below the current state, $X^{t}$).
  Each trajectory generated via independent $Z^{1:T}_0$ noise samples represents a sample from, $p\left(X^{1:T} \vert X^0, X^{-1}\right)$.}
  \label{fig:schematic}
\end{figure}

Recent advances in machine learning (ML)-based weather prediction (MLWP) have introduced an alternative to NWP-based forecasting methods, and have been shown to provide greater accuracy and efficiency for non-probabilistic forecasts~\citep{keisler2022forecasting, kurth2022fourcastnet, bi2023accurate, lam2023learning, chen2023fengwu,nguyen2023scaling, li2023fuxi}. 
Rather than forecasting a single weather trajectory, or a distribution of trajectories, these methods have largely focused on forecasting the mean of the probable trajectories, with relatively little emphasis on quantifying the uncertainty associated with a forecast. They are typically trained to minimise mean squared error (MSE), 
and as a result tend to produce blurry forecast states, especially at longer lead times, rather than a specific realisation of a possible weather state \citep{lam2023learning}.
There have been limited attempts to use traditional initial condition perturbation methods to produce ensembles with MLWP-based forecasts~\citep{kurth2022fourcastnet,bi2023accurate,li2023fuxi}, however they have not addressed the issue of blurring, and have not rivaled operational ensemble forecasts like ENS.
One exception is \ngcm~\citep{kochkov2023neural}, a hybrid NWP-MLWP method, which combines a traditional NWP's dynamical core with local ML-based parameterisations, and shows competitive performance with operational ensemble forecasts. However its ensembles are \ang{1.4} spatial resolution, which is an order of magnitude coarser than operational NWP-based forecasts, and it faces serious challenges scaling to operational settings.

\section{\ourmodel}

Here we introduce a probabilistic weather model---\ourmodel---which generates global 15-day ensemble forecasts at \modelresolution resolution that, for the first time, are more accurate than the top operational ensemble system, ECMWF's ENS. Generating a single 15-day \ourmodel forecast takes about 8 minutes on a Cloud TPUv5 device, and an ensemble of forecasts can be generated in parallel.

\ourmodel models the conditional probability distribution, $p\left(X^{t+1} \vert X^{t}, X^{t-1} \right)$, of the future weather state $X^{t+1}$ conditional on the current and previous weather states.
A forecast trajectory, $X^{1:T}$, of length $T$ is modeled by conditioning on the initial and previous states, $(X^{0}, X^{-1})$, and factoring the joint distribution over successive states,
\begin{align*}
p\left(X^{1:T} \vert X^{0}, X^{-1} \right) &= \prod_{t=0}^{T-1} p\left(X^{t+1} \vert X^{t}, X^{t-1}\right) \quad ,
\end{align*} 
each of which are sampled autoregressively.

The representation of the global weather state, $X$, is comprised of 6 surface variables, and 6 atmospheric variables at 13 vertical pressure levels on a \modelresolution latitude-longitude grid (see~\cref{tab:app:variables} for details).
The forecast horizon is 15 days, with 12 hours between successive steps $t$ and $t+1$, so $T=30$. 

\ourmodel is implemented as a conditional diffusion model \citep{karras2022elucidating,song2021scorebased,sohl2015deep}, a type of generative ML model used to generate new samples from a given data distribution, which underpins many of the recent advances in modelling natural images, sounds, and videos under the umbrella of ``generative AI'' \citep{yang2023diffusion, croitoru2023diffusion}. 
Diffusion models work through a process of iterative refinement. A future atmospheric state, $X^{t+1}$, is produced by iteratively refining a candidate state, $Z^{t+1}_0$, initialised purely from noise, conditioned on the previous two atmospheric states $(X^{t}, X^{t-1})$. The blue box in \cref{fig:schematic} shows how the first forecast step is generated from the initial conditions, and how the full trajectory, $X^{1:T}$, is generated autoregressively.
Because each time step in a forecast is initialised with noise (i.e., $Z^{t+1}_0$), the process can be repeated with different noise samples, to generate an ensemble of trajectories.
See~\cref{sec:sampling} for further details.

At each stage of the iterative refinement process, \ourmodel applies a neural network architecture comprised of an encoder, processor, and decoder. 
The encoder component maps the input $Z^{t+1}_n$, as well as the conditioning $(X^{t}, X^{t-1})$, from the native latitude-longitude grid to an internal learned representation defined on an 6-times-refined icosahedral mesh.
The processor component is a graph transformer~\citep{vaswani2017attention} in which each node attends to its k-hop neighbourhood on the mesh.
The decoder component maps from the internal mesh representation back to $Z^{t+1}_{n+1}$, which is defined on the latitude-longitude grid.

\ourmodel is trained on 40 years of ERA5 reanalysis data \citep{hersbach2020era5} from 1979 to 2018, using a standard diffusion model denoising objective~\citep{karras2022elucidating}. Crucially, while we only ever directly train \ourmodel on a single step prediction task, it can be rolled out autoregressively to generate 15-day ensemble forecasts.
\cref{sec:app:diffusion_model} provides full details of the \ourmodel architecture and training protocol.

When evaluating \ourmodel we initialise it with ERA5 reanalysis, together with perturbations derived from the ERA5 Ensemble of Data Assimilations (EDA)~\citep{ECMWF-EDA} which reflect uncertainty about the initial conditions. Further details are in \cref{sec:app:initial-conditions}.

\section{Realism of \ourmodel samples}

\begin{figure}
  \vspace{-2cm}
  \centering
\includegraphics[width=\textwidth,trim={0 0.5cm 0 0},clip]{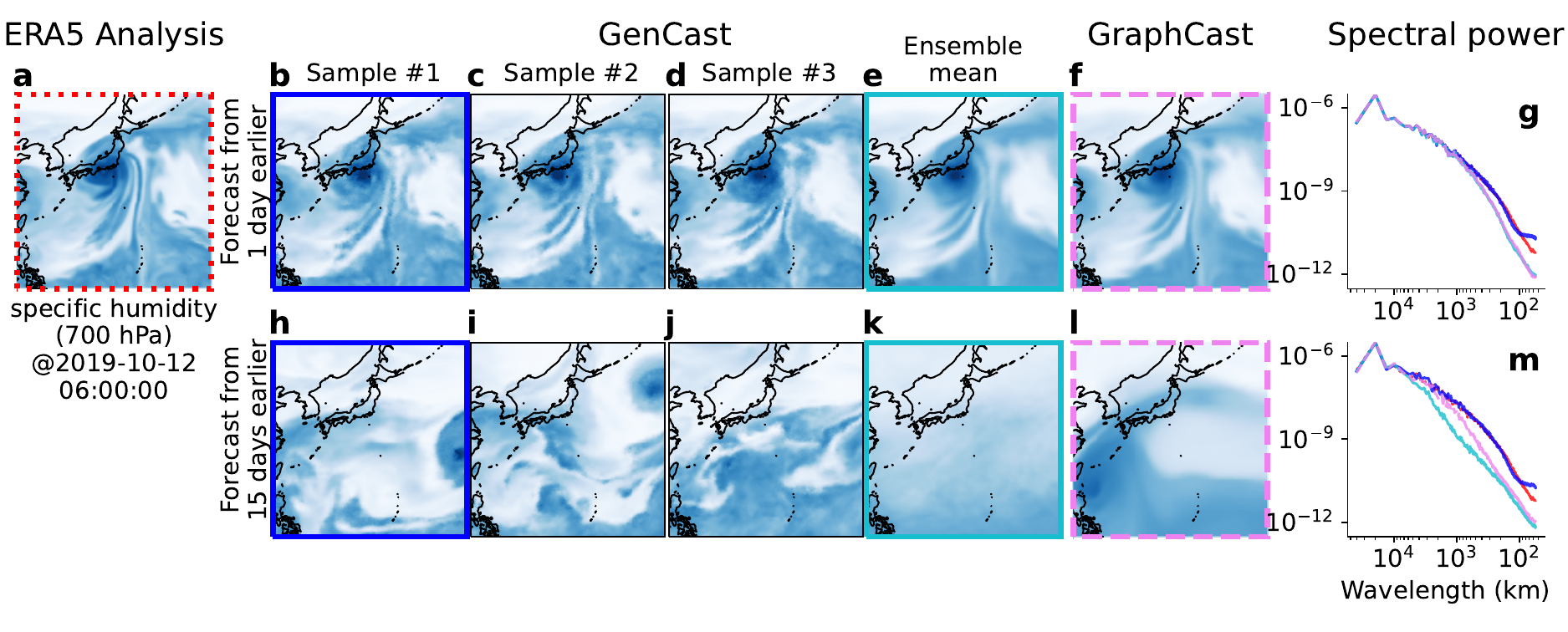}
  \includegraphics[width=\textwidth]{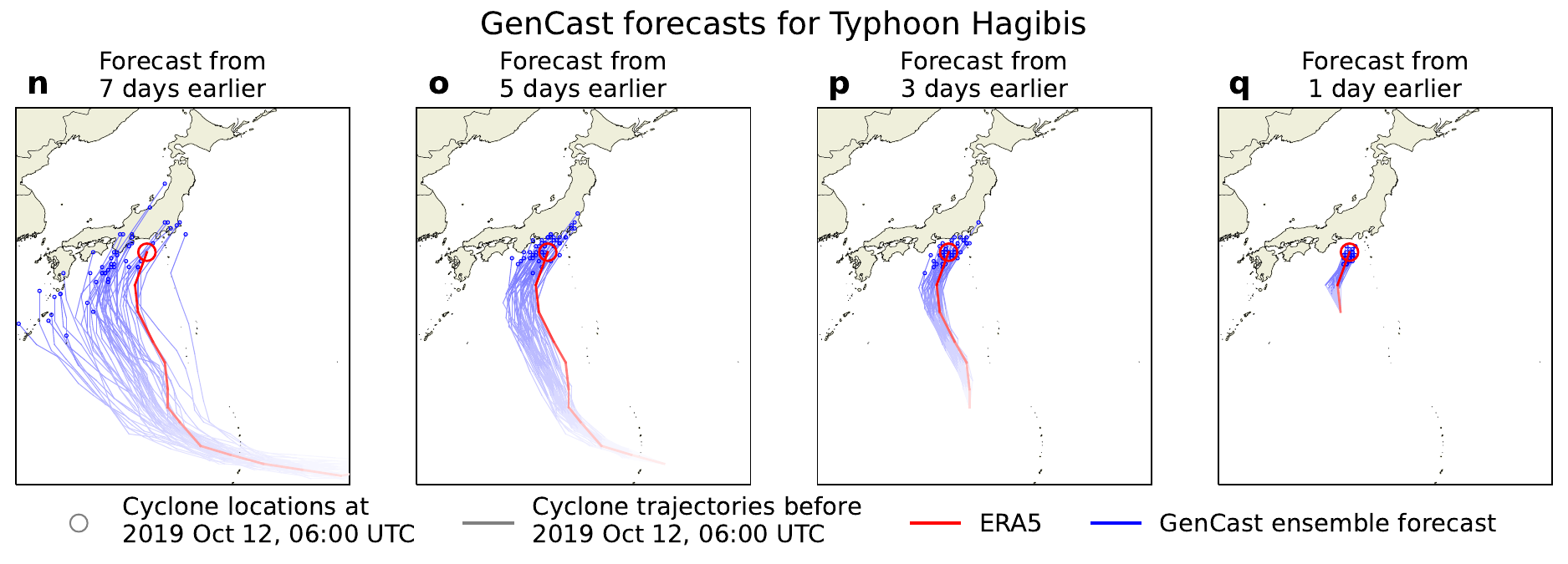}
  \caption{\small\textbf{Visualisation of forecasts and tropical cyclone tracks for the validity time of 12~October, 2019, 06~UTC, hours before Typhoon Hagibis made landfall in Japan.} 
  \textbf{(a)}~The ERA5 analysis state for specific humidity at 700hPa, at validity time 06 UTC, October 12, 2019, shows Typhoon Hagibis near the center of the frame. 
  \textbf{(b-d)}~Three sample \ourmodel forecast states, initialised one day earlier, show how the samples are sharp, and very similar to one another.
  \textbf{(e)}~The \ourmodel ensemble mean, obtained by computing the mean of 50 sample states like in (b-d), is somewhat blurry, showing how uncertainty results in a blurrier average state.
  \textbf{(f)}~The forecast state from (deterministic) \gc{}, initialised one day earlier like in (b-e), is blurry, similar to \ourmodel{}'s ensemble mean.
  \textbf{(g)}~The spatial power spectrum of the states in (a), (b), (e), and (f), where line colors match the frames of the panels, show how \ourmodel samples' spectra closely match ERA5's, while the blurrier \ourmodel ensemble mean and \gc states have less power at shorter wavelengths.
  \textbf{(h-m)}~These subplots are analogous to (b-g), except the forecasts are initialised 15 days earlier. The \ourmodel samples are still sharp (h-j), while the \ourmodel ensemble mean (k) and \gc (l) states are even blurrier than at the 1-day lead time. This is also reflected in the power spectrum (m), where the \ourmodel samples' spectra still closely match ERA5's, while the \ourmodel ensemble mean and \gc states have even less power in the shorter wavelengths relative to the 1-day lead time in (g). See~\cref{sec:app:visualizations_hagibis} for more variables and lead times.
  \textbf{(n-q)}~Typhoon Hagibis's trajectory based on ERA5 (in red) and the ensemble of tropical cyclone trajectories from \ourmodel (in blue) up to a validity time 4 hours before the cyclone made landfall on Japan. \ourmodel forecasts are shown at lead times of 7, 3, 5, and 1 day/s. The blue and red circles show cyclone locations at the validity time. 
  At long lead times the cyclone trajectories have substantial spread, while for the shorter lead times the predictive uncertainty collapses to a small range of trajectories. 
  See~\cref{sec:app:cyclone_tracks} for details and additional cyclone visualisations.}
  \label{fig:visualization}
\end{figure}

\cref{fig:visualization} illustrates \ourmodel's forecast samples, as well as post-processed tropical cyclone tracks, for Typhoon Hagibis, shortly before it made landfall in Japan on October 12, 2019.
\cref{fig:visualization}a-m contrasts \ourmodel with \gc~\citep{lam2023learning}, a top deterministic MLWP-based method. It shows that \ourmodel samples, at both 1- and 15-day lead times, are sharp and have spherical harmonic power spectra that closely match the ERA5 ground truth. As expected, the \ourmodel ensemble mean is blurry, like \gc's states, which lose power at the high frequencies (see also \cref{fig:headline_spectra_group_0} and \cref{fig:headline_spectra_group_1}). This reflects how \ourmodel's ensembles represent realistic samples from the predictive forecast distribution, while MSE-trained models like \gc are closer to the ensemble mean. Over and above the blurring issue, forecasts of the ensemble mean for each variable can be biased when computing nonlinear derived quantities such as wind speed (which is the L2-norm of the $(u,v)$ wind vector, $\sqrt{u^2 + v^2}$). This is because a nonlinear function of an average is not necessarily equal to the average of that nonlinear function. Computing nonlinear functions directly on the samples avoids this: as shown in \cref{sec:app:wind_speed_bias}, \ourmodel exhibits minimal bias on wind speed, unlike \gc which exhibits large negative bias.

\cref{fig:visualization}n-q shows how \ourmodel forecasts can be used in important downstream applications, such as predicting the paths of tropical cyclones.
We use Typhoon Hagibis as an example, which was the costliest tropical cyclone of 2019, causing widespread destruction across eastern Japan and over \$17 billion in damages.
When initialised 7 days before Typhoon Hagibis's landfall, \ourmodel's predicted trajectory has high uncertainty, covering a wide range of possible scenarios.
At shorter lead times, \ourmodel's uncertainty about the cyclone's path is lower, reflecting greater confidence about the landfall timing and location.
Figures~\ref{fig:app:visualization_cyclone_idai}--- \ref{fig:app:visualization_cyclone_dorian} visualise \ourmodel and ENS cyclone forecasts for other important cyclones in 2019.

\section{Baselines}\label{sec:baselines}
    
We compare \ourmodel to ENS, which we regridded from its native \ang{0.2} latitude-longitude resolution\footnote{ENS's resolution changed from \ang{0.2} to  \ang{0.1} in mid-2023, however our validation and test years are 2018 and 2019, respectively.} to \modelresolution. 
ENS contains 50 perturbed ensemble members, so we used 50-member \ourmodel ensembles to perform all evaluation. The public TIGGE archive~\citep{swinbank2016tigge} only makes all 50 ENS ensemble members available for surface variables and for atmospheric variables at 8 pressure levels in the troposphere. So these are the variables and levels we compare models on. 

For an ML baseline, we use \gc~\citep{lam2023learning} to generate ensemble forecasts (denoted \gcens). Since \gc is a deterministic model, all ensemble spread must come from differences in the initial conditions. We initialise \gc using ERA5 reanalysis plus the ERA5 EDA perturbations used for \ourmodel, however in order to improve dispersion and skill of the ensemble we found it necessary to supplement them with additional ad hoc Gaussian Process perturbations; full details are in \cref{sec:app:initial-conditions}.

For a fair comparison among models, we evaluate each model against its corresponding analysis. We follow the protocol described in \citet[Appendix Section 5.2]{lam2023learning}, in which we evaluate ECMWF's operational forecasts against \hresfczero (a dataset comprising the initial conditions used for their HRES deterministic forecast), and we evaluate ML models, that were trained and initialised using ERA5, against ERA5.

During our 2019 test period, ENS was initialised with analyses whose assimilation window had between 3 and 5 hours of look-ahead beyond the stated initialisation time \citep{continuous-da}. Again following \cite{lam2023learning}'s protocol, we initialise ML models using ERA5 at 06 UTC and 18 UTC, since these benefit from only 3 hours of look-ahead\footnote{An exception is sea surface temperature (SST), which in ERA5 is updated once per 24 hours based on a longer assimilation window; however this is a relatively slow-changing variable and we do not expect any extra look-ahead to provide a material advantage.}. In comparison, ERA5's 00 and 12 UTC times have 9 hours of look-ahead, which may translate into improved metrics for 00/12 UTC initialised ML model forecasts (as shown in \cref{fig:scorecard_00_12_init}).
Overall, the difference in assimilation windows used in our evaluation leaves ENS with a small advantage of up to 2 hours additional look-ahead over the ML models, for all variables except sea surface temperature (SST).

We follow standard verification practice \citep[Section 2.2.35]{wmo-manual-2019} in evaluating ensemble forecasts using a single deterministic analysis as ground truth. However we note that this favours under-dispersion at short lead times. For example, at a lead time approaching zero, scores computed against deterministic analysis would be optimized by setting all initial conditions equal to the deterministic analysis, ignoring initial condition uncertainty.

\begin{figure}[tb]
  \centering
  \includegraphics[width=\textwidth]{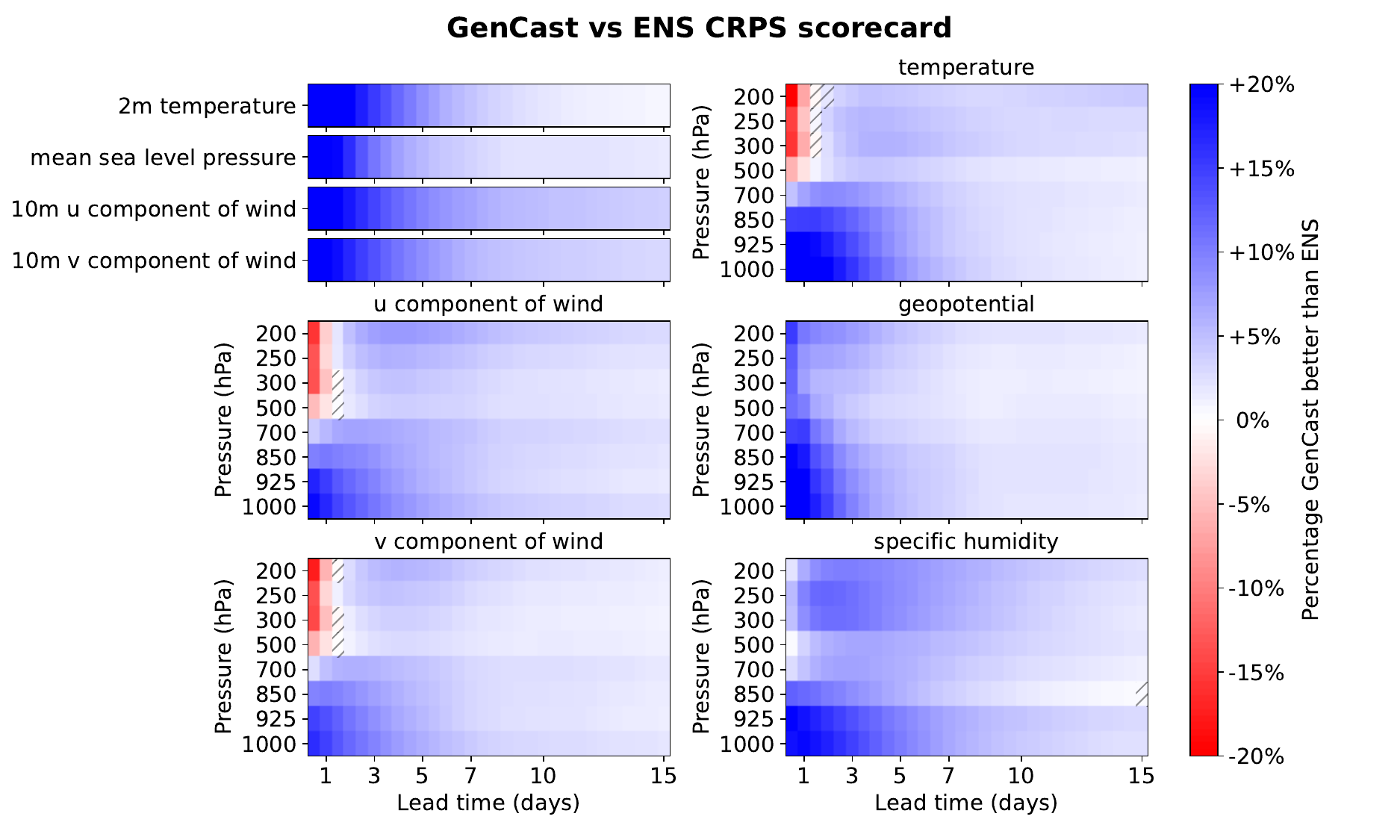}
  \caption{\textbf{CRPS scores for \ourmodel versus ENS in 2019.} The scorecard compares CRPS skill between \ourmodel{} and ENS across all variables and 8 pressure levels, where dark blue indicates \ourmodel{} is 20\% better than ENS, dark red indicates \ourmodel is 20\% worse, and white means they perform equally.  The results indicate that \ourmodel{} significantly outperforms ENS ($p < 0.05$) on \winpercentage\% of all reported variable, lead time, and level combinations. Hatched regions indicate where neither model is significantly better. }
  \label{fig:skill}
\end{figure}

\section{Ensemble skill}
\label{sec:skill}

To evaluate \ourmodel's performance compared to ENS and \gcens --- our primary NWP- and MLWP-based baselines, respectively --- we report Continuous Ranked Probability Scores (CRPS) for all evaluation variables, lead times, and pressure levels.
The CRPS \citep{gneiting2007strictly} is a standard measure of the skill of a probabilistic forecast. It measures how well the marginal distributions of the forecast represent the ground truth, and it is minimized, in expectation, by a forecast whose marginals reflect true predictive uncertainty. See~\cref{sec:app:crps} for CRPS's mathematical definition.

As shown in \cref{fig:skill}'s CRPS scorecard, \ourmodel's forecasts are significantly more skillful  ($p < 0.05$) than ENS's on \winpercentage\% of our \numCRPStargets variable, lead time, and vertical level combinations (and \winpercentageafterthirtysixhours\% of targets at lead times greater than 36 hours)\footnote{Note, due to our lack of confidence in the the quality of ERA5 precipitation data, we exclude precipitation results from our main results, and refer readers to \cref{sec:app:precip}.}.
Blue cells on the scorecard indicate a variable, lead time, and level combination where \ourmodel has better (i.e. lower) CRPS skill scores than ENS, while red cells indicate lower CRPS for ENS.
\ourmodel's largest improvements are often at shorter lead times up to around 3-5 days, for surface variables, as well as temperature and specific humidity at higher pressure levels, where CRPS skill scores range between 10-30\% better. 
When compared to \gcens, we found \ourmodel has better CRPS on 99.6\% of targets, as shown in \cref{sec:app:gcens_scorecards} and \cref{fig:app:gencast_vs_gc_perturbed}. \cref{fig:app:gc_perturbed_vs_ens} further shows that \gcens only outperforms ENS on CRPS in 27\% of targets, which are concentrated at shorter lead times. 

We also compared the Root Mean Squared Error (RMSE) of \ourmodel's and ENS's ensemble means. The Ensemble-Mean RMSE measures how the mean of an ensemble of forecasts matches ground truth. While RMSE is a common metric for deterministic forecasts, it does not account for uncertainty, which is central to probabilistic verification. Nonetheless, as shown in \cref{fig:app:emrmse_gencast_vs_ens}, \ourmodel's ensemble mean RMSE is as good or better than ENS's on 96\% of targets, and significantly better $(p<0.05)$ on 82\% of targets.

\section{Ensemble calibration}
\label{sec:calibration}

For a probabilistic forecast to be useful, it should be well-calibrated: it should know when it may be wrong, and have confidence when it is likely to be right.
This allows a decision-maker to hedge their choices in proportion to the forecast's confidence.
Two common tools in the weather community for evaluating calibration, on average, are spread/skill ratios and rank histograms.

Well-calibrated probabilistic forecasts exhibit uncertainty (as measured by ensemble spread) which is commensurate on average with the size of its errors~\citep{fortin2014should}. 
The degree to which this relationship holds can be quantified by the spread/skill ratio defined in \cref{sec:app:spread-skill-ratio}. This ratio should be 1 for a perfect ensemble forecast, with values greater than 1 suggestive of over-dispersion (an under-confident forecast), and values less than 1 suggestive of under-dispersion (over-confidence). 

Similarly, the members of an ideal ensemble forecast should be indistinguishable from ground truth values. Deviations from this property on average can be diagnosed using rank histograms \citep{talagrand1999evaluation}. The rank histogram should be flat if the truth tends to be indistinguishable from the ensemble members, $\cap$-shaped if the truth mostly ranks near the center of the ensemble (indicating the ensembles are over-dispersed), and $\cup$-shaped if the truth ranks mostly near the tails of the ensemble (indicating the ensembles are under-dispersed). 
See~\cref{sec:app:verification-metrics} for definitions and details.

Generally \ourmodel exhibits as good, or better, calibration than ENS according to these metrics (see \cref{sec:app:calibration}). 
\ourmodel's spread-skill scores are usually close to 1, as shown in \cref{fig:headline_spread_skill}.
\ourmodel also tends to have flat rank histograms, as shown in \cref{fig:headline_rank_hist_group_0} and \cref{fig:headline_rank_hist_group_1}.
In contrast, \gcens is consistently overconfident, showing spread-skill scores <1 and $\cup$-shaped rank histograms.

\section{Extreme weather events}

Deciding whether, and how, to prepare for extreme weather involves trading off the \textit{cost} of preparing for an event which might not occur, against the \textit{loss} incurred if no preparations are made and it does occur. In this cost/loss model~\citep{thompson1952operational}, the optimal decision is to prepare whenever the probability of the event exceeds the cost/loss ratio. Skillful forecasts of rare and extreme weather events are crucial for optimising these decisions, and the value of a forecast to a particular decision-maker depends on the situation's specific cost/loss ratio.

A standard metric for representing the potential value of a forecast, as a function of cost/loss ratio, is Relative Economic Value~(REV,~\cite{richardson2000skill}). 
REV quantifies the gain of using a particular forecast over using climatology as a naive baseline (which would yield REV of~0), normalised by the gain of using a perfect forecast which is correct 100\% of the time (which would yield REV of~1). See~\cref{sec:app:rev} for details.
We compute REV for a range of different cost/loss ratio trade-offs, showing the impact of a particular forecast across a wide range of decision scenarios. 

\subsection{Local surface extremes}

\begin{figure}[tb]
  \centering
  \includegraphics[width=\textwidth]{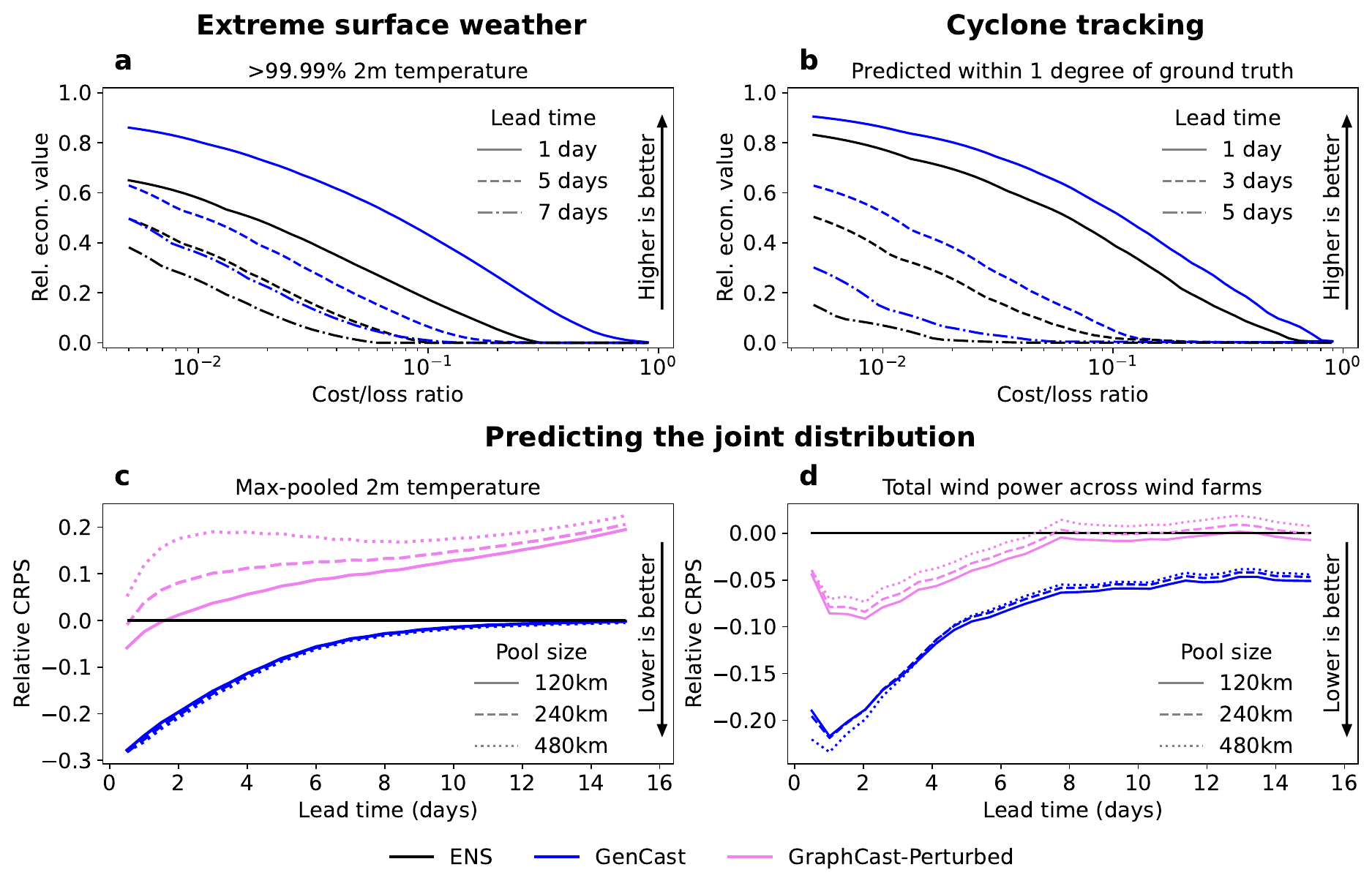}
  \caption{\textbf{Extreme weather, joint measures, and wind power.} (a) Relative economic value (REV) for predictions of the exceedance of the 99.99th percentile for \SI{2}{m} temperature, at lead times of 1, 5 and 7 days. (b) Relative economic value (REV) for predictions of the presence of a cyclone in any location at a given time, at lead times of 1, 3, and 5 days. (c) Relative CRPS of max-pooled 2m temperature, for different pooling region sizes. (d) Relative CRPS of the total wind power summed across wind farm locations in pooling regions of different sizes.}
  \label{fig:panel_applications}
\end{figure}

Extreme heat, cold, wind, and other severe surface weather pose hazards to lives and property, but can be anticipated and prepared for using ensemble forecasts.
To build intuition, consider a scenario in which an upcoming heatwave threatens to yield a 7-year high temperature tomorrow (exceeding the 99.99th percentile relative to climatology). Let us assume the cost of preparing by staging relief supplies is 1/20 of the loss associated with the public health impact if the population endures the heat unprepared, giving a cost/loss ratio is $0.05$.
\cref{fig:panel_applications}a shows that \ourmodel (blue curves) has 1.9$\times$ better REV than ENS (black curves) in this scenario. \ourmodel also shows a consistent advantage across other cost/loss values (x-axis), and for lead times of 5 and 7 days (dashed and dash-dot lines, respectively).
\cref{sec:app:extremes} shows that the increased value provided by \ourmodel also applies across other levels of extreme event (other exceedance percentiles), and for other variables including extreme low temperature, high surface wind speed, and low mean sea level pressure (see \cref{fig:app:supplementary_rev_2t_high,fig:app:supplementary_rev_2t_low,fig:app:supplementary_rev_wind_speed_high,fig:app:supplementary_rev_msl_low} for details and statistical tests).

Another method for assessing skill at extreme-weather prediction is to evaluate Brier skill scores (\cref{sec:app:brier}) for the exceedance of the same high and low percentiles of the climatological distribution.
\cref{fig:app:supplementary_brier} shows the Brier skill scores of \ourmodel compared to ENS for predicting the exceedance of the 99.99th, 99.9th, and 99th percentile events, for high \SI{2}{m} temperature and \SI{10}{m} wind speed, and for extremely low temperature and mean sea level pressure below the 0.01, 0.1 and 1st percentiles across all lead times. \ourmodel significantly ($p < 0.05$) improves on ENS across all of these variables and thresholds, with the exception of low mean sea level pressure where some improvements are not significant.

\subsection{Tropical cyclones}\label{sec:body:cyclones}

Tropical cyclones cause thousands of deaths and tens of billions of dollars in damages on average every year, and accurately predicting their emergence and trajectory is essential to prepare for and mitigate their effects~\citep{martinez2020forecast}.
However, given that it is impossible to predict definitively if a cyclone will appear and exactly where it will go, the ideal forecast should generate a distribution over possible trajectories, skillfully assigning probabilities for whether a particular location will be hit by a cyclone at a given time. 

To obtain forecasts for the presence and tracks of tropical cyclones, we apply the TempestExtremes tropical cyclone tracker~\citep{ullrich2021tempestextremes} to ensemble forecasts from both \ourmodel and ENS, for each initialisation time and each ensemble member. We then apply the same tracker to ERA5 and \hresfczero, generating ground truth tracks for \ourmodel and ENS, respectively. We define the presence or absence of a cyclone centre in a given location at a given time as a binary event, with a forecast probability equal to the proportion of ensemble members that predict a cyclone in that location at that time. See \cref{sec:app:cyclones} for details.

To assess the value of different models' tropical cyclone forecasts, consider a scenario where you need to decide whether to prepare for the potential landfall of a cyclone. As an example, assume that enduring a cyclone unprepared would incur a loss 100$\times$ greater than the cost of planning for evacuation or stockpiling emergency supplies like food, water, medicine and fuel (i.e., a cost/loss ratio of 0.01). 
\cref{fig:panel_applications}b shows that relying on \ourmodel instead of ENS for a cost/loss ratio of 0.01 yields $11\%,38\%,106\%$ higher relative economic values at 1 day, 3 days and 5 days lead time respectively. Across other cost/loss ratios, and for lead times up to 7 days, \ourmodel consistently yields greater REV (see \cref{sec:app:cyclones_results} for further tropical cyclone results). This suggests \ourmodel's forecasts can provide substantial value in deciding how to prepare for tropical cyclones.

\section{Skill in predicting the joint distribution}
\label{sec:joint}

Ensemble forecasting systems like \ourmodel sample from a high-dimensional joint spatiotemporal distribution, as opposed to modelling weather independently at each point in space and time.
Capturing this joint distribution accurately is key for applications where spatially coherent weather trajectories are needed, such as renewable energy prediction.
To assess this, we evaluate \ourmodel, \gcens, and ENS on two spatial structure prediction tasks: spatial pooling and regional wind power prediction.

\subsection{Spatially pooled evaluation}
\label{sec:joint_dense_pooled_metrics}

Per-grid-cell verification, like the CRPS scorecard (\cref{fig:skill}), evaluates marginal predictive distributions, but do not measure the accuracy of the forecasts' larger-scale spatial structures.
Here we use a neighbourhood verification approach~\citep{ebert2008fuzzy}, computing average-pooled and max-pooled versions of the CRPS scorecard from \cref{sec:skill}.
Forecasts and analysis targets are aggregated over circular spatial regions distributed to jointly cover the Earth's surface, and CRPS is computed on these pooled quantities for a range of pooling region sizes from \SI{120}{km} to \SI{3828}{km} (see \cref{sec:app:pooled_evaluation_description}).

\cref{fig:panel_applications}c shows the relative CRPS of \ourmodel compared to ENS and \gcens for max-pooled \SI{2}{m} temperature. \ourmodel shows $\sim$25\% improvement at short lead times, and plateauing from approximately 8 days lead time close to ENS performance. These improvements are consistent across spatial scales, unlike \gcens, which not only does substantially worse than ENS, but also sees its relative performance degrade as the pooling size increases. 

Aggregating over all 5400 pooled verification targets---across all variables, levels, lead times, and spatial scales---\ourmodel outperforms ENS on average-pooled CRPS in 98.1\% of targets and on max-pooled CRPS in 97.6\% of
targets, with relative performance increasing at larger scales (see \cref{fig:pool_scorecard_gencast_average_low_res}, \cref{fig:pool_scorecard_gencast_max_low_res}).
This contrasts with \gcens, which only outperforms ENS in 23.9\% of targets for average-pooled CRPS and 7.0\% of targets for max-pooled CRPS. %
This suggests that \ourmodel captures spatial structure substantially better than ENS and \gcens across all surface and atmospheric variables.

\subsection{Regional wind power forecasting}\label{sec:wind_farms}
In the electricity sector, power grid operators use regional wind power forecasts for tasks like unit commitment and reserve quantification~\citep{siebert2008development}, where leveraging forecast uncertainty can improve decision-making~\citep{matos2011setting, rachunok202assessment}.
However, forecast errors make it harder to ensure the balance of supply and demand.
Because of this, operators often rely on fossil fuel-based spinning reserves~\citep{siebert2008development}, which undermines wind power's potential for reducing carbon emissions~\citep{gielen2019role}.

To estimate \ourmodel's potential impact in wind energy applications, we conducted a simplified regional wind power forecasting experiment, where we interpolate \SI{10}{m} wind speed of forecasts and analyses at all 5344 wind farm locations from the Global Power Plant Database (\citealt{byers2018global}, \cref{fig:gppd_wind_farm_locs}).
\SI{10}{m} wind speeds are converted to wind power using a standard idealised power curve (see \cref{fig:wind_turbine_power_curve}) multiplied by each wind farm's nominal capacity.
Wind power (in megawatts) is then summed across arbitrary groupings of wind farms defined by the pooling regions from \cref{sec:joint_dense_pooled_metrics} with sizes \SI{120}{km}, \SI{240}{km}, and \SI{480}{km} (see \cref{sec:app:wind_power_evaluation} for further details).

\ourmodel outperforms ENS's CRPS by around $\sim$20\% up to lead times of 2 days, 10-20\% from 2-4 days, and retains statistically significant $(p<0.05)$ improvements out to 10 days (\cref{fig:panel_applications}d and \cref{fig:app:wind_farms_significance}). This is a substantially greater improvement than that provided by \gcens.
Though only a simplified experiment, 
these results suggest \ourmodel provides more skillful wind forecasts, which has potential value for the management and utilisation of wind energy.

\section{Conclusion}

Our results suggest that probabilistic weather forecasts based on MLWP are more skillful, and faster to generate, than the top NWP-based ensemble forecast, ECMWF's ENS. \ourmodel generates stable, skillful 15-day global forecasts, despite only being explicitly trained on 12-hour steps, which suggests that \ourmodel's predicted weather states are realistic samples of weather trajectories. Across a comprehensive probabilistic scorecard of variables, vertical levels, and lead times, as well as several important application settings, \ourmodel
has greater accuracy and decision-making value than ENS. \ourmodel's ensembles have well-calibrated uncertainty, allowing users to trust that the model generally has the right level of confidence in its predictions.

These results invite the question: what is required to deploy MLWP-based models in operation? 
While \ourmodel operates at \modelresolution, which approximately matches the \ang{0.2} resolution of ENS (which was upgraded to \ang{0.1} in mid-2023), it may be valuable to scale up to higher resolution. 
\ourmodel is trained and evaluated on ERA5, and while \gc has been shown to be similarly effective when trained on operational HRES inputs and targets \citep{rasp2023weatherbench}, \ourmodel will also need to be trained on operational ENS inputs.
As a diffusion model, \ourmodel is computationally more expensive than an equivalent deterministic MLWP architecture, because it requires multiple function evaluations required to sample each forecast timestep, and for applications that require very fast forecasts, distillation~\citep{salimans2022progressive} and other efficiency techniques should be explored. 
And importantly, \ourmodel relies on initial conditions from a traditional NWP ensemble data assimilation system, and therefore using \ourmodel for operational forecasts will still require NWP-based data assimilation to provide the input analysis.

Together our results open a new front in weather forecasting, promising greater accuracy, efficiency, and accessibility across a wide range of settings. More generally, our work demonstrates that cutting-edge generative AI methods can capture very high-dimensional and complex distributions over rich temporal dynamics, with sufficient accuracy and reliability to support effective decision-making in crucial applications.

\newpage

\section*{Acknowledgements}\label{sec:app:acknowledgements}

In alphabetical
order, we thank Alexis Boukouvalas, Matthew Chantry, Yuri Chervonyi, Sander Dieleman, Praneet Dutta, Carl Elkin, Sarah Elwes, Isla Finney, Meire Fortunato, Marta Garnelo, Xavier Glorot, Stephan Hoyer, Dmitrii Kochkov, Pushmeet Kohli, Paul Komarek, Amy Li, Sean Lovett, Rahul Mahrsee, Liam McCafferty, Piotr Mirowski, Kevin Murphy, Kate Musgrave, Charlie Nash, Nick Pezzotti, Luis Piloto, Stephan Rasp, Suman Ravuri, Daniel Rothenberg, Tim Salimans, Fei Sha, Kunal Shah, Jon Small, Duncan Smith, Octavian Voicu, David Wallis, Daniel Worrall, Flora Xue, and Janni Yuval for their advice and/or feedback on our work. We also thank ECMWF for providing invaluable datasets to the research community. 

\bibliography{main}

\begin{appendices}

\counterwithin{figure}{section}
\counterwithin{table}{section}
\counterwithin{equation}{section}
\renewcommand\thefigure{\thesection\arabic{figure}}
\renewcommand\thetable{\thesection\arabic{table}}
\renewcommand{\theequation}{\thesection\arabic{equation}}

\noindent{\huge\bfseries Supplementary material\par}

\startcontents[sections]
\printcontents[sections]{l}{1}{\setcounter{tocdepth}{3}}

\newpage

\section{Task definition and general approach}\label{sec:task-setup}
In its general formulation, the task of probabilistic forecasting from the present time $t=0$ into the future is to model the joint probability distribution, $p(\bar{X}^{0:T} \vert O^{\leq 0})$, where $T$ is the forecast horizon, $\bar{X}^i$ denotes the atmospheric state at time $i$, and $O^{\leq 0}$ are observations made up to the forecast initialisation time $t=0$. In NWP-based weather forecasts, $\bar{X}$ represents variables that are part of the underlying atmospheric fluid equations, on a discrete spatial grid or mesh. This joint distribution is assumed to be Markov, and is factored as,
\begin{align*}
    p(\bar{X}^{0:T} \vert O^{\le 0}) &= \underbrace{\vphantom{\prod_{t=1}^{T}} p(\bar{X}^{0} \vert O^{\le 0})}_{\texttt{State inference}} \  \underbrace{\vphantom{\prod_{t=1}^{T}} p(\bar{X}^{1:T} \vert \bar{X}^{0}) }_{\texttt{Forecast model}} =
    p(\bar{X}^{0} \vert O^{\le 0})%
    \  
    \prod_{t=1}^{T} p(\bar{X}^{t} \vert \bar{X}^{t- 1})
\end{align*}
Our innovation in this work is an MLWP-based \texttt{Forecast model}, and we simply adopt a traditional NWP-based \texttt{State inference} approach. We make a number of approximations to the above general formulation, as follows.

The standard approach to \texttt{State inference} is to approximate samples from $p(\bar{X}^0 | O^{\leq 0})$ --- the distribution over the current atmospheric state --- with an ensemble of NWP-based analysis\footnote{``Analysis'' refers to an estimate of the state of the atmosphere based on observations; ``reanalysis'' is an analysis computed retrospectively based on historical observations.} states, which are generated using a window of observations in a process known as ``data assimilation''. We too rely on ensemble analysis to approximate the \texttt{State inference} stage: we use ECMWF's ERA5 archive of reanalysis data \citep{hersbach2020era5}, including the ERA5 EDA (ensemble of data assimilations), to represent the initial conditions $\bar{X}^{0}$.  Details are in \cref{sec:app:initial-conditions}.

We approximate the full atmospheric state, $\bar{X}$, with $X$, an $84 \times 720 \times 1440$ array, which includes 6 surface variables, and 6 atmospheric variables at each of 13 vertical pressure levels (see \cref{tab:app:variables}), on a \quarterdegree latitude-longitude grid. Because $X$ is only a partial representation of $\bar{X}$, forecasts initialised with a single $X$ may not be able to generate forecasts as accurate as those initialised with $\bar{X}$, and in practice we found conditioning on the current \textit{and} previous state, $X^{0}$ and $X^{-1}$, respectively, to be more effective. Therefore, the initial conditions provided as input to our model are $(X^0, X^{-1})$. This means we approximate a single step of the \texttt{Forecast model} as $p(\bar{X}^{t} \vert \bar{X}^{t - 1}) \approx p(X^{t} \vert  X^{t - 1}, X^{t- 2})$.

We generate 15-day forecasts, at 12 hour steps, so $T=30$.
To sample a weather trajectory, $X^{1:T} \sim p(X^{1:T} \vert X^0, X^{-1})$, we first sample $(X^0, X^{-1})$ using ERA5 and ERA5 EDA. Then for each subsequent timestep $t=1,\dots,T$, we sample $X^{t} \sim p(X^{t} \vert X^{t- 1}, X^{t - 2})$, autoregressively, conditioned on the previous two steps $(X^{t-1}, X^{t-2})$. This process can repeated multiple times, and the multiple resulting sample trajectories can then be used to compute Monte Carlo estimates of probabilities and expectations of interest. This is a form of ensemble forecasting, which is a popular method in traditional NWP-based forecasting, because the joint distribution is extremely high-dimensional, and is thus intractable to represent explicitly.

\section{Data}

\subsection{ERA5 training and evaluation data for \ourmodel}

We trained \ourmodel on a dataset built from the ECMWF's ERA5 archive~\citep{hersbach2020era5}, a large corpus of global reanalysis data. The following description of the ERA5 dataset is adapted directly from \cite{lam2023learning}. ECMWF's ERA5~\citep{hersbach2020era5}\footnote{See ERA5 documentation: \url{https://confluence.ecmwf.int/display/CKB/ERA5}.} archive is a large corpus of data that represents the global weather from 1959 to the present, at 1 hour increments, for hundreds of static, surface, and atmospheric variables. The ERA5 archive is based on \textit{reanalysis}, which uses ECMWF's Integrated Forecast System (IFS) cycle 42r1 that was operational for most of 2016, with an ensemble 4D-Var data assimilation scheme. ERA5 assimilated 12-hour windows of observations, from 21-09 UTC and 09-21 UTC, as well as previous forecasts, into a dense representation of the weather's state, for each historical date and time.
The ERA5 archive consists of a deterministic reanalysis (ERA5) computed at \erafivenativedegree native latitude/longitude resolution, and an associated ensemble reanalysis (the ERA5 Ensemble of Data Assimilations or ERA5 EDA), which consists of 9 ensemble members and 1 control member computed at a lower \erafiveensnativedegree native resolution.

Our ERA5 dataset contains a subset of available variables in ECMWF's ERA5 archive (\cref{tab:app:variables}), on a \quarterdegree equiangular grid, at 13 pressure levels\footnote{We follow common practice of using pressure as our vertical coordinate, instead of altitude. A ``pressure level'' is a field of altitudes with equal pressure. E.g., ``pressure level 500 \unit{hPa}'' corresponds to the field of altitudes for which the pressure is 500 \unit{hPa}. The relationship between pressure and altitude is determined by the geopotential variable.} corresponding to the levels of the WeatherBench \citep{rasp2020weatherbench} benchmark:
50, 100, 150, 200, 250, 300, 400, 500, 600, 700, 850, 925, and 1000~\unit{hPa}. We subsampled the temporal resolution from 1 hour to 6 hours, corresponding to 00:00, 06:00, 12:00 and 18:00 UTC times each day. From this dataset, we extracted sequences at 12 hour temporal resolution (sequences of 00/12 UTC or 06/18 UTC times) to train \ourmodel.

Although its temporal resolution is hourly, ERA5 only assimilates observations in 12-hour windows, from 21 UTC--09 UTC and 09 UTC--21 UTC. This means that steps taken within a single 12h assimilation window have a different, less dispersed distribution to those which jump from one window into the next. By choosing a 12 hour time step we avoid training on this bimodal distribution, and ensure that our model always predicts a target from the next assimilation window.

For accumulated variables such as  precipitation, instead of subsampling the data in time, we accumulated values over the 12 hour period preceding each time. %

Our dataset covers the period of 1979 to 2019.
During the development phase of \ourmodel{}, we used dates from 1979 to 2017 for training and validated results on 2018. Before starting the test phase, we froze all model and training choices, retrained the model on data from 1979 to 2018, and evaluated results on 2019.
For the \gcens baseline, we trained the deterministic \gc baseline with the protocol described in \cite{lam2023learning} at \quarterdegree{} resolution, using data from 1979 to 2018, which is consistent with the approach to the data split for \ourmodel{}.

\subsection{Data preprocessing.}

\paragraph{Handling of ENS missing data.}\label{sec:app:ensnan}
When trying to download ENS forecasts for surface variables for the 2019-10-17 00:00 UTC initialisation, we were met with persistent errors, so we left that initialisation out of the ENS evaluation, and used neighboring information when performing paired statistical tests for differences in verification metrics (see \cref{sec:app:statistical-tests} for details).

\paragraph{Variables with NaNs.}\label{sec:app:varsnan}
ERA5 sea surface temperature (SST) data contains NaNs over land by default. As preprocessing of the SST training and evaluation data, values over land are replaced with the minimum sea surface temperature seen globally in a subset of ERA5.

\paragraph{Upsampling EDA analysis.}
\label{sec:app:edaupsample}
As described in \ref{sec:app:initial-conditions}, the initialisation of the ensemble forecasts for \ourmodel and \gcens utilises the ERA5 EDA dataset. Since ERA5 EDA exists only at \ang{0.5} resolution,  we bilinearly interpolate the EDA perturbations to \modelresolution before adding them to the ERA5 inputs. We noticed that for sea surface temperature (SST), which takes NaN values over land regions, after bilinear interpolation to \modelresolution, the NaN locations of the EDA data and the NaN locations of ERA5 did not match exactly. We built the sea surface temperature inputs according to the NaN locations of ERA5 at \modelresolution, and simply used ERA5 deterministic analysis SST for any points for which EDA SST values were not available.

\begin{center}
\begin{table}[htbp]
\centering
\hspace*{-0.8cm}
\begin{tabular}{c c c c c} 
 \toprule
 \textbf{Type} & \textbf{Variable name} & \textbf{Short} & \textbf{ECMWF} & \textbf{Role (accumulation} \\ [0.5ex] 
  &  &  \textbf{name} & \textbf{Parameter ID} & \textbf{period, if applicable)} \\ [0.5ex] 
 \midrule
 Atmospheric & Geopotential & z & 129 & Input/Predicted \\
Atmospheric & Specific humidity & q & 133 & Input/Predicted \\
Atmospheric & Temperature & t & 130 & Input/Predicted \\
Atmospheric & U component of wind & u & 131 & Input/Predicted \\
Atmospheric & V component of wind & v & 132 & Input/Predicted \\
Atmospheric & Vertical velocity & w & 135 & Input/Predicted \\
Single & 2 metre temperature & 2t & 167 & Input/Predicted 
\\
Single & 10 metre u wind component & 10u & 165 & Input/Predicted \\
Single & 10 metre v wind component & 10v & 166 & Input/Predicted \\
Single & Mean sea level pressure & msl & 151 & Input/Predicted \\
 Single & Sea Surface Temperature & sst & 34 & Input/Predicted \\
Single & Total precipitation & tp & 228 & Predicted (12h) \\
 \midrule
 \midrule
Static & Geopotential at surface & z & 129 & Input \\
Static & Land-sea mask & lsm & 172 & Input \\
Static & Latitude & n/a & n/a & Input \\
Static & Longitude & n/a & n/a & Input \\
Clock & Local time of day & n/a & n/a & Input \\
Clock & Elapsed year progress & n/a & n/a & Input \\
 \bottomrule
\end{tabular}
\caption{\small\textbf{ECMWF variables used in our datasets.} The ``Type'' column indicates whether the variable represents a \textit{static} property, a time-varying \textit{single}-level property (e.g., surface variables are included), or a time-varying \textit{atmospheric} property. The ``Variable name'' and ``Short name'' columns are ECMWF's labels. The ``ECMWF Parameter ID'' column is a ECMWF's numeric label, and can be used to construct the URL for ECMWF's description of the variable, by appending it as suffix to the following prefix, replacing ``ID'' with the numeric code: \texttt{https://apps.ecmwf.int/codes/grib/param-db/?id=ID}. The ``Role'' column indicates whether the variable is something our model takes as input and predicts, or only uses as input context (the double horizontal line separates predicted from input-only variables, to make the partitioning more visible).}
\label{tab:app:variables}
\end{table}
\end{center}

\subsection{Data Availability}

\begin{itemize}
    \item ERA5 and ERA5 EDA datasets were downloaded and are downloadable from the Climate Data Store (CDS) of the Copernicus Climate Change Service\\ (\url{https://cds.climate.copernicus.eu}). The results contain modified Copernicus Climate Change Service information 2020. Neither the European Commission nor ECMWF is responsible for any use that may be made of the Copernicus information or data it contains.

    \item ENS and HRES data were downloaded and are downloadable from the ECMWF as of April 2024 (\url{https://apps.ecmwf.int/datasets/data/tigge/}), and usable according to the license described at \url{https://apps.ecmwf.int/datasets/licences/tigge/}. The data forms part of the THORPEX Interactive Grand Global Ensemble (TIGGE) archive (\url{https://confluence.ecmwf.int/display/TIGGE}). TIGGE is an initiative of the World Weather Research Programme (WWRP).
    
    \item The Global Power Plant Database v1.3.0 was and can be downloaded from \url{https://datasets.wri.org/dataset/globalpowerplantdatabase}. The idealised wind turbine power curve was and can be downloaded from the National Renewable Energy Laboratory \url{https://github.com/NREL/turbine-models/blob/master/Onshore/WTK_Validation_IEC-2_normalized.csv}.
\end{itemize}

\section{Initial conditions}
\label{sec:app:initial-conditions}

We explored a number of methods for deriving ensemble initial conditions for our ML models. In this section we discuss these methods in more detail in order to motivate the choices we made. We also supply full details of the ad-hoc Gaussian Process perturbations described in \cref{sec:app:eda-and-gp-perturbations}.

\subsection{Deterministic analysis}

We tried initialising every ensemble member in the same way, using two consecutive deterministic analyses from ECMWF's ERA5 reanalysis dataset. This is a relatively crude approximation as it collapses initial condition uncertainty down to a pair of point estimates.
A deterministic model initialised this way will have no dispersion, but the method works relatively well for our diffusion model. A comparison of the CRPS results reported in the main results section to those achieved when initialising with deterministic analysis is included in \cref{sec:app:eda_ablation}.

\subsection{Ensemble of data assimilations (EDA) analysis}

Another possibility is to initialise each ensemble member using two consecutive ensemble analyses from a corresponding ensemble member of ECMWF's ERA5 EDA. This better approximates a posterior distribution conditional on observations. However it still tends to be under-dispersed \citep{buizza2008potential}, and the ERA5 EDA was computed using IFS at half the native resolution used for ERA5 itself \citep{hersbach2020era5}, reducing the quality of the analysis.

\subsection{Deterministic analysis plus EDA perturbations}

Following ECMWF \citep{buizza2008potential} we combine the high quality of the deterministic analysis with the distributional information of the ensemble analysis, by taking the deterministic ERA5 analysis and adding to it perturbations equal to the difference between ERA5 EDA members and their mean. This is analogous to the use of EDA perturbations in initialising ECMWF's ENS forecast system, although we do not include the additional singular value perturbations used at ECMWF to address under-dispersion of the EDA perturbations, largely because we were not easily able to obtain singular value perturbations for use with the ERA5 EDA dataset. Nevertheless this method yielded the best results for our diffusion model beyond the shortest lead times (\cref{fig:no_eda_vs_with_eda_scorecards}).

\subsection{Additional ad hoc Gaussian process perturbations}
\label{sec:app:eda-and-gp-perturbations}

The final method we explored takes the deterministic ERA5 analysis with EDA perturbations and adds to it perturbations sampled from a zero-mean Gaussian process on the sphere. This process uses the Gaussian-like stationary isotropic correlation function from \cite{weaver2001correlation}, with a horizontal decorrelation length-scale of $1200 \unit{km}$.

\begin{center}
\begin{table}[h]
\centering
\begin{tabular}{l}\toprule
\textbf{Variable} \\ \midrule
Geopotential \\
Temperature \\
U component of wind \\
V component of wind \\
2-metre temperature \\ \bottomrule
\end{tabular}
\caption{Variables to which Gaussian process perturbations are applied}
\label{tab:app:perturbed-variables}
\end{table}
\end{center}
We sample independent perturbations for each of the variables listed in \cref{tab:app:perturbed-variables};
other variables are not perturbed by the Gaussian process. We obtained better results perturbing only this subset than the full set of input variables, although we have not exhaustively investigated the best subset of variables to perturb.

The marginal standard deviations of the perturbations are equal to $0.085$ times those of 6-hour differences in the corresponding variables at each respective pressure (or surface) level.  Aside from these differences in scale, for a given variable the same perturbation is used for all levels, and for both input timesteps. This is equivalent to infinite vertical and temporal decorrelation lengthscales. We investigated using lower vertical decorrelation lengthscales but found they did not help.

We selected the scale factor $0.085$ and horizontal decorrelation length-scale $1200\unit{km}$ based on CRPS scores for the resulting forecasts over a range of variables and lead times, after sweeping over scale factors $(0.03, 0.05, 0.07, 0.085, 0.1, 0.3)$, and decorrelation length-scales ($30\unit{km}$, $480 \unit{km}$, $1200 \unit{km}$, $3000 \unit{km}$). Results were significantly worse at the shortest lengthscale we tried of $30\unit{km}$, but were not otherwise very sensitive to decorrelation lengthscale.

These perturbations are quite crude; in particular they are not flow-dependent and do not take any care to preserve physical invariants. Nevertheless we've found them to be surprisingly effective at longer lead times with the deterministic \gc model. In combination with EDA perturbations, they appear to play a similar role to the singular value perturbations used operationally at ECMWF to address the under-dispersion of EDA perturbations. They also allow us to generate ensembles of arbitrary size from deterministic \gc, without being limited by the size of the EDA ensemble.
However as noted, with the diffusion model they do not give any additional skill over EDA perturbations alone, in fact for the diffusion model, skill is reduced when these perturbations are added with all but the smallest scale factors.

\section{Diffusion model specification}\label{sec:app:diffusion_model}

Our key innovation is to model $p(X^{t} | X^{t-1}, X^{t-2})$ with a learned neural network-based diffusion model \citep{karras2022elucidating,song2021scorebased, sohl2015deep}, which allows us to draw samples from it. Rather than sampling $X^{t}$ directly, our approach is to sample a residual $Z^{t}$ with respect to the most recent weather state  $X^{t-1}$, where the residuals have been normalised to unit variance on a per-variable and per-level basis as was done for \gc in \cite{lam2023learning}.
$X^{t}$ is then computed as $X^{t} = X^{t-1} + S Z^{t}$, where $S$ is a diagonal matrix which inverts the normalisation\footnote{An exception to this is precipitation, for which we set $X^{t} = S Z^{t}$ without adding the previous state.}.

We broadly follow the diffusion framework presented in \cite{karras2022elucidating}, and refer the reader to their paper for a more rigorous introduction to diffusion, as well as a detailed treatment of the available modelling decisions. We adopt their choices of noise schedule, noise scaling, loss weighting by noise level, and preconditioning. However we make changes to the noise distribution, the training-time distribution of noise levels, and add additional loss weightings, all of which are described below. These changes improve performance on the task of probabilistic weather forecasting.

\subsection{Sampling process}\label{sec:sampling}

The sampling process begins by drawing an initial sample $Z^{t}_0$ from a noise distribution on the sphere $p_{noise}(\cdot | \sigma_0)$ that is described in \cref{sec:noise-sphere}, at a high initial noise level $\sigma_0$. After $N$ steps of transformation we end up at $Z^{t}_N:=Z^{t}$, our sample from the target distribution at noise level $\sigma_N = 0$. To take us from one to the other, we apply an ODE solver to the probability flow ODE described by \cite{song2021scorebased,karras2022elucidating}. Each step of this solver is denoted by $r_\theta$ (see \cref{fig:schematic}), with
\begin{align}
    Z^{t}_{i+1} &= r_\theta\bigl(Z^{t}_i; X^{t-1}, X^{t-2}, \sigma_{i+1}, \sigma_i\bigr)
\end{align}
taking us from noise level $\sigma_i$ to the next (lower) noise level $\sigma_{i+1}$, conditioned on $(X^{t-1}, X^{t-2})$.

We use the second-order DPMSolver++2S solver \citep{lu2022dpm}, augmented with the stochastic churn (again making use of $p_{noise}$ from \cref{sec:noise-sphere}) and noise inflation techniques used in \cite{karras2022elucidating} to inject further stochasticity into the sampling process. In conditioning on previous timesteps we follow the Conditional Denoising Estimator approach outlined and motivated by \cite{batzolis2021conditional}.

Each step $r_\theta$ of the solver makes use of a learned denoiser $D_\theta$ with parameters $\theta$, described in detail below (\cref{sec:architecture}). We take $N=20$ solver steps per generated forecast timestep. As we are using a second-order solver, each step $r_\theta$ requires two function evaluations of the denoiser $D_\theta$ (with the exception of the last step which requires only a single evaluation). This results in 39 function evaluations in total. See \cref{sec:app:sampler-hyperparameters} for further details including a full list of sampling hyperparameters.

\subsection{Noise distribution on the sphere}
\label{sec:noise-sphere}

At the core of a diffusion model is the addition and removal of noise, drawn from some distribution $p_{noise}(\cdot | \sigma)$ parameterized by noise level $\sigma$. When using diffusion to generate natural images \citep{ho2020denoising}, $p_{noise}$ is usually chosen to be i.i.d. Gaussian, and much of the early theory around diffusion models was developed for the case of Gaussian white noise. A true white noise process on the sphere is isotropic or rotation-invariant, and is characterised by a flat spherical harmonic power spectrum in expectation. However these properties do not hold if we attempt to approximate it at finite resolution by sampling i.i.d. noise on the cells of our equiangular latitude-longitude grid. This is due to the greater density of cells near to the poles which results in more power at higher frequencies in the spherical harmonic domain.

Empirically we didn't find this to be a fatal problem; nonetheless we found we can obtain a small but consistent improvement using a different approach to noise sampling which is sensitive to the spherical geometry. We sample isotropic Gaussian noise in the spherical harmonic domain, with an expected power spectrum that is flat over the range of wavenumbers that our grid is able to resolve, and truncated thereafter. We then project it onto our discrete grid using the inverse spherical harmonic transform \citep{driscoll1994-discrete-sht}. The resulting per-grid-cell noise values are not independent especially near to the poles, but are approximately independent at the resolution resolved at the equator, and display the desired properties of isotropy and flat power spectrum.

\subsection{Denoiser architecture}\label{sec:architecture}

To recap, our diffusion sampling process involves taking multiple solver steps $r_\theta$, and each solver step calls a denoiser $D_\theta$ as part of its computation. We parameterise the denoiser $D_\theta$ following \cite{karras2022elucidating} as a preconditioned version of a neural network function $f_\theta$.
\begin{align}
    D_\theta\bigl(Z^{t}_\sigma; \; X^{t-1}, X^{t-2}, \; \sigma \bigr)
    &:= c_{skip}(\sigma)\cdot Z^{t}_\sigma \; + \;  c_{out}(\sigma)\cdot f_{\theta}\Bigl(c_{in}(\sigma) Z^{t}_\sigma; \; X^{t-1}, X^{t-2}, \; c_{noise}(\sigma)\Bigr) . \label{eq:dtheta}
\end{align}
Here $Z^{t}_\sigma$ denotes a noise-corrupted version of the target $Z^{t}$ at noise level $\sigma$, and $c_{in}$, $c_{out}$, $c_{skip}$ and $c_{noise}$ are preconditioning functions taken from Table 1 in \cite{karras2022elucidating}, with $\sigma_{data}=1$ due to the normalisation of the targets.

The architecture used for $f_\theta$ is related to the \gc architecture described in \cite{lam2023learning}.  To be precise, the Encoder and Decoder architectures stay the same, and those inputs to the encoder corresponding to the previous two timesteps are normalised to zero mean and unit variance in the same way. However, unlike in \gc, which uses a similar message-passing GNN for the Processor architecture as in the Encoder and Decoder, in \ourmodel the Processor is a graph transformer model operating on a spherical mesh which computes neighbourhood-based self-attention.  Unlike the multimesh used in \gc, the mesh in \ourmodel is a 6-times refined icosahedral mesh as defined in \cite{lam2023learning}, with 41,162 nodes and 246,960 edges. The Processor consists of 16 consecutive standard transformer blocks \citep{vaswani2017attention, nguyen2019transformers}, with feature dimension equal to 512. The 4-head self-attention mechanism in each block is such that each node in the mesh attends to itself and to  all other nodes which are within its 32-hop neighbourhood on the mesh. 

To condition on prior timesteps $X^{t-1}, X^{t-2}$, we concatenate these along the channel dimension with the input to be denoised, and feed this as input to the model. Conditioning on noise level $\sigma$ is achieved by replacing all layer-norm layers in the architecture with conditional layer-norm \citep{chen2021adaspeech} layers. We transform log noise levels into a vector of sine/cosine Fourier features at 32 frequencies with base period 16, then pass them through a 2-layer MLP to obtain 16-dimensional noise-level encodings. Each of the conditional layer-norm layers applies a further linear layer to output replacements for the standard scale and offset parameters of layer norm, conditioned on these noise-level encodings.

\subsection{Training the denoiser}
\label{sec:training-denoiser}

At training time we apply the denoiser to a version of the target $Z^{t}$ which has been corrupted by adding noise $\epsilon \sim p_{noise}(\cdot | \sigma)$ at noise level $\sigma$:
\begin{align}
\label{eq:denoiser}
    Y^{t} &= D_\theta\bigl(Z^{t} + \epsilon; \; X^{t-1}, X^{t-2}, \; \sigma\bigr).
\end{align}
We train its output, denoted $Y^{t}$, to predict the expectation of the noise-free target $Z^{t}$ by minimizing the following mean-squared-error objective weighted per elevation level and by latitude-longitude cell area,
\begin{align}\label{eq: training objective}
    \sum_{t \in \mathcal{D}_{\text{train}} }
    \E \left[ \lambda(\sigma) \frac{1}{|\G||J|}\sum_{\sll \in \G} \sum_{j \in J}
    \varweight_j 
    \latitudeweight_{\sll} 
    (Y_{\sll,j}^{t} - Z_{\sll, j}^{t})^2 \right],
\end{align}
where
\begin{itemize}
    \item $t$ indexes the different timesteps in the training set $\mathcal{D}_{\text{train}}$,
    \item $j \in J$ indexes the variable, and for atmospheric variables the pressure level, i.e.
    $J=\{\vll{z}{1000}, \vll{z}{850},$\\$\dots, \vll{2t}, \vll{msl}\}$,
    \item $\sll \in \G$ indexes the location (latitude and longitude coordinates) in the grid,
    \item $\varweight_j$ is the per-variable-level loss weight, set as in \cite{lam2023learning} with the additional sea surface temperature variable weighted at $0.01$.
    \item $\latitudeweight_{\sll}$ is the area of the latitude-longitude grid cell, which varies with latitude, and is normalised to unit mean over the grid,
    \item $\lambda(\sigma)$ is the per-noise-level loss weight from \cite{karras2022elucidating},
    \item the expectation is taken over $\sigma \sim p_{train}$, $\epsilon \sim p_{noise}(\cdot; \sigma)$.
\end{itemize}

Instead of using the log-normal distribution for $p_{train}$ that is suggested in \cite{karras2022elucidating}, we construct a distribution whose quantiles match the noise-level schedule used for sample generation, assigning higher probability to noise levels which are closer together during sampling. Specifically its inverse CDF is:
$$F^{-1}(u) = \Bigl(\sigma_{max}^{\frac1\rho} + u(\sigma_{min}^{\frac1\rho} - \sigma_{max}^{\frac1\rho})\Bigr)^\rho$$
and we sample from it by drawing $u \sim U[0,1]$. At training time we use the same $\rho$ as at sampling time, but a slightly wider range for $[\sigma_{min}, \sigma_{max}]$, values are in \cref{tab:app:sampler-settings}.

As in \cite{lam2023learning} we weight the squared error made at each latitude-longitude grid cell by a per-variable-level loss weight, as well as the normalised area of that grid cell; this is also a departure from \cite{karras2022elucidating}.

Unlike \gc, which is fine-tuned by back-propagating gradients through 12 step trajectories (3 days with 6 hour steps) produced by feeding the model its own inputs during training, \ourmodel is only ever trained using targets that consist of the next 12 hour state, without ever being provided its own predictions on previous steps as inputs.

\subsubsection{Resolution training schedule}

The results reported in this paper were generated by a model which was trained in a two-stage process. Stage one was a pre-training stage, taking 2 million training steps. During this stage, the ground truth dataset was bi-linearly downsampled from \modelresolution to \onedegree, and the denoiser architecture used a 5-refined icosahedral mesh. This training stage takes a little over 3.5 days using 32 TPUv5 instances. After this training phase was complete, Stage 2 was conducted, fine-tuning the model to \modelresolution, taking 64 000 further training steps. This takes just under 1.5 days using 32 TPUv5 instances.  During Stage two, the ground truth data is kept at \modelresolution, and the denoiser architecture is updated to take in \modelresolution data and output \modelresolution outputs, and to operate on a 6-refined icosahedral mesh. The GNN and graph-transformer architectures are such that the same model weights can operate on the higher data and mesh resolutions without any alterations. We do, however, make one minor modification before beginning the fine-tuning stage, in order to decrease the shock to the model of operating on higher resolution data. In the Encoder GNN, which performs message passing between the grid and mesh nodes, when the data resolution increases from \onedegree to \modelresolution, the number of messages being received by each mesh node increases by a factor of 16. To approximately preserve the scale of the incoming signal to all mesh nodes at the start of finetuning, we therefore divide the sum of these message vectors by 16. The optimisation hyperparameters used for both stages of training are detailed in \cref{table:app:optimizer-hyperparams}.

\begin{table}[ht!]
\centering
\begin{tabular}{lc}
\toprule
\textbf{Optimiser}    & \textbf{AdamW} \citep{loshchilov2018decoupled}    \\ \midrule
LR decay schedule & Cosine \\
Stage 1: Batch size & 32    \\
Stage 1: Warm-up steps & 1e3    \\
Stage 1: Total train steps & 2e6 \\
Stage 1: Peak LR      & 1e-3   \\
Stage 1: Weight decay      & 0.1 \\
Stage 2: Batch size & 32    \\
Stage 2: Warm-up steps & 5e3    \\
Stage 2: Total train steps & 64000 \\
Stage 2: Peak LR      & 1e-4   \\
Stage 2: Weight decay      & 0.1 \\ \bottomrule
\end{tabular}
\caption{Diffusion model training hyperparameters.}
\label{table:app:optimizer-hyperparams}
\end{table}

\subsection{Sampler hyperparameters}
\label{sec:app:sampler-hyperparameters}

To draw samples we use DPMSolver++2S \citep{lu2022dpm} as a drop-in replacement for the second-order Heun solver used in \cite{karras2022elucidating}. It is also a second-order ODE solver requiring $2N-1$ function evaluations for $N$ noise levels (two per step, one fewer for the final Euler step). We augment it with the stochastic churn and noise inflation described in Algorithm 2 of \cite{karras2022elucidating}.

At sampling time we adopt the noise level schedule specified by \cite{karras2022elucidating}:

$$\sigma_i := \Bigl(\sigma_{max}^{\frac1\rho} + \frac{i}{N-1}(\sigma_{min}^{\frac1\rho} - \sigma_{max}^{\frac1\rho})\Bigr)^\rho \quad \text{for } i\in \{0 \ldots N-1\}.$$

Settings for the parameters of this schedule ($\rho, \sigma_{min}, \sigma_{max}$, $N$) as well for stochastic churn and noise inflation are given in \cref{tab:app:sampler-settings}.

\begin{center}
\begin{table}[h]
\centering
\begin{tabular}{cccc} 
 \toprule
    \textbf{Name} & \textbf{Notation} & \textbf{Value, sampling} & \textbf{Value, training}\\
 \midrule
    Maximum noise level & $\sigma_{max}$ & 80 & 88 \\ 
    Minimum noise level & $\sigma_{min}$ & 0.03 & 0.02 \\
    Shape of noise distribution & $\rho$ & 7 & 7 \\
    Number of noise levels & $N$ & 20 & \\
    Stochastic churn rate & $S_{churn}$ & 2.5 & \\
    Churn maximum noise level & $S_{tmax}$ & 80 & \\
    Churn minimum noise level & $S_{tmin}$ & 0.75 & \\
    Noise level inflation factor & $S_{noise}$ & 1.05 & \\
 \bottomrule
\end{tabular}
\caption{\small\textbf{Settings used at sampling time, and their equivalents at training time where applicable}. Notation aligns with that used in \cite{karras2022elucidating}.}
\label{tab:app:sampler-settings}
\end{table}
\end{center}

\section{Supplementary evaluation details}

\subsection{ENS initialisation and evaluation times}\label{sec:app:ens_init_times}

As discussed in \cref{sec:baselines}, we only evaluate \ourmodel on forecasts initialised at 06/18 UTC, since using 00/12-initialised forecasts gives \ourmodel an additional advantage (see \cref{fig:scorecard_00_12_init}) due to the longer data-assimilation lookahead. Ideally we would compare all models at the same 06/18 UTC initialisation times, however ENS forecasts from 06/18 UTC are only archived up to 6-day lead times and are not free for public download. Hence we evaluate ENS on forecasts initialised at 00/12 UTC. For globally averaged metrics this should not matter, and in fact \cite{lam2023learning} found that 00/12 UTC initialisation tends to give a small advantage in RMSE to the deterministic HRES forecast over the 06/18 UTC initialisation, and we expect a similar minor advantage to apply to ENS.  
However, the regional wind power evaluation (\cref{sec:wind_farms}) is sensitive to the diurnal cycle since wind power capacity is sparsely and non-uniformly distributed around the world. Thus in this case it is important to compare forecasts by ENS and \ourmodel at the same set of validity times. We therefore evaluate ENS (initialised at 00/12 UTC) at the same 06/18 UTC targets as \ourmodel.
However, \ourmodel produces 06/18 UTC forecasts at lead times of 12 hours, 24 hours, 36 hours, ..., while for ENS we only obtain 06/18 UTC forecasts at lead times of 6 hours, 18 hours, 30 hours, and so on.
In order to estimate ENS's 06/18 UTC regional wind power CRPS at the same lead times as \ourmodel, we linearly interpolate ENS's CRPS curve. In \cref{sec:app:lead_time_averaging} we validate this approach on 2018 data where we did get access to ENS 06/18 UTC initialisations, showing that this lead time interpolation in fact overestimates the performance of ENS, in particular at short lead times.

\subsection{Verification metrics}
\label{sec:app:verification-metrics}

In the following, for a particular variable, level and lead time,

\begin{itemize}
    \item $x^m_{i,k}$ denotes the value of the $m$th of $M$ ensemble members in a forecast from initialisation time indexed by $k=1\ldots K$, at latitude and longitude indexed by $i \in G$.
    \item $y_{i,k}$ denotes the corresponding verification target.
    \item $\bar{x}_{i,k} = \frac1M \sum_m x^m_{i,k}$ denotes the ensemble mean.
    \item $a_i$ denotes the area of the latitude-longitude grid cell, which varies by latitude and is normalized to unit mean over the grid.
\end{itemize}

\subsubsection{CRPS}
\label{sec:app:crps}

The Continuous Ranked Probability Score (CRPS, see e.g. \cite{gneiting2007strictly}) is estimated for our ensembles as follows:

\begin{align}\label{eq:app:crps}
    \text{CRPS} &:= \frac1K \sum_k \frac1{|G|} \sum_i a_i \Big( \frac1M \sum_m |x^m_{i,k} - y_{i,k}| - \frac{1}{2M^2} \sum_{m,m'} |x^m_{i,k} - x^{m'}_{i,k}| \Big).
\end{align}
For CRPS, smaller is better.

Here we use the traditional estimator of CRPS, which estimates the expected CRPS of the empirical distribution of a finite ensemble of size $M$. An alternative is the `fair' CRPS estimator \citep{zamo2018estimation}, which estimates the CRPS that would be attained in the limit of an infinite ensemble. In our case this would not be appropriate as we make use of precomputed NWP ensemble analyses of finite size, and we do not wish to assume that they can be scaled up arbitrarily. Although we can scale up our own ensembles more cheaply, in doing so we must repeatedly re-use the same set of NWP ensemble analysis members, and the fair CRPS estimator would not correctly model this reuse.

\subsubsection{Ensemble mean RMSE}
\label{sec:app:emrmse}
This is defined as:
\begin{align}\label{eq:app:emrmse}
    \text{EnsembleMeanRMSE} &:= \sqrt{
    \frac1K \sum_k \frac1{|G|} \sum_i a_i \left(y_{i,k} - \bar{x}_{i,k}\right)^2
    },
\end{align}
and again smaller is better.

\subsubsection{Spread/skill ratio}
\label{sec:app:spread-skill-ratio}

Following \cite{fortin2014should} we define spread as the root mean estimate of ensemble variance given below, and skill as the Ensemble-Mean RMSE from \cref{eq:app:emrmse}:
\begin{align}\label{eq:app:spread-skill}
    \text{Spread} &:= \sqrt{
    \frac1K \sum_k \frac1{|G|} \sum_i a_i \frac1{M-1} \sum_m \left(x^m_{i,k} - \bar{x}_{i,k}\right)^2
    }, \\
    \text{Skill} &:= \text{EnsembleMeanRMSE}.
\end{align}

Under these definitions and the assumption of a perfect forecast where ensemble members and ground truth $\{x_{i,k}^1, \ldots, x_{i,k}^M, y_{i,k}\}$ are all exchangeable, \cite{fortin2014should} Equation 15 shows that
\begin{align}
    \text{Skill} &\approx \sqrt{\frac{M+1}{M}} \text{Spread},
\end{align}
which motivates the following definition of spread/skill ratio including a correction for ensemble size:
\begin{align}\label{eq:app:spread-skill-ratio}
    \text{SpreadSkillRatio} &:= \sqrt{\frac{M+1}{M}} \frac{\text{Spread}}{\text{Skill}}.
\end{align}
Under the perfect forecast assumption then, we expect to see $\text{SpreadSkillRatio} \approx 1$. While diagnosis of under- or over-dispersion is confounded by forecast bias \citep{hamill2001interpretation,wilks2011reliability}, if we assume such bias is relatively small, we can associate under-dispersion on average with spread/skill~<~1 and over-dispersion on average with spread/skill~>~1.

\subsubsection{Rank histogram}
\label{sec:app:rank-histogram}
The rank histogram \citep{talagrand1999evaluation} measures where the ground truth value tends to fall with respect to the ensemble distribution. More precisely, for each evaluation time and for each grid cell, we record the rank of the ground truth among the forecast ensemble members, from 1 to $M+1$, where $M$ is the ensemble size, and plot the histogram of these ranks. For perfect ensemble forecasts in which ensemble members and ground truth are all exchangeable, we expect to see a flat rank histogram, since the true value should fall between any pair of proximal sorted ensemble values with equal probability.  If we assume relatively small bias, rank histograms can be used to diagnose under- or over-dispersion \citep{hamill2001interpretation,wilks2011reliability}. Under-dispersion will tend to result in the rank of the ground truth falling towards or beyond the outer bounds of the ensembles' values, resulting in a rank histogram with a $\cup$-shape. Conversely, an over-dispersed but unbiased ensemble will result in the ground truth falling predominantly near the center of the range of ensemble values, causing the rank histogram to have a $\cap$-shape. Peaks on either side of the rank histogram can suggest over-- or under- prediction bias.

\subsubsection{Brier skill score}
\label{sec:app:brier}

Suppose now that predictions $x^m_{i,k} \in \{0,1\}$ and targets $y_{i,k}\in \{0,1\}$ are binary variables, such as whether or not one is close to a cyclone, or whether surface temperature exceeds the 99.9th percentile of climatology.

Our ensemble means $\bar{x}_{i,k}$ now correspond to empirical predictive probabilities for the events in question, and are scored by the Brier score as follows:
\begin{align}\label{eq:app:brier}
    \text{BrierScore} &:= \frac1K \sum_k \frac1{|G|} \sum_i a_i \left(\bar{x}_{i,k} - y_{i,k}\right)^2
\end{align}
As with CRPS this is chosen to score the empirical probabilities of an ensemble of a specific finite size. While estimators exist for the Brier score attained by the true underlying probability of an infinite ensemble, these are not appropriate in our case for the same reason described above.

To obtain a Brier skill score, this is then normalized relative to the Brier score attained by predicting a fixed (and location-independent) climatological probability of the event, estimated using the evaluation set:

\begin{align}\label{eq:app:brier_score}
    p_{clim} &:= \frac1K \sum_k \frac1{|G|} \sum_i a_i y_{i,k} \\
    \text{BrierScore}_{clim} &:= \frac1K \sum_k \frac1{|G|} \sum_i a_i \left(p_{clim} - y_{i,k}\right)^2 \\
    \text{BrierSkillScore} &:= 1 -  \frac{\text{BrierScore}}{\text{BrierScore}_{clim}}
\end{align}

Brier skill score is 0 for the climatological forecast and 1 for a perfect forecast, so larger is better.

\subsubsection{Relative Economic Value}
\label{sec:app:rev}

The Brier score evaluates predictive probabilities without regard for the specific decision rule that will be used to act on those probabilities. In doing so it implicitly places much weight on decision thresholds that are not relevant in practice, especially in decision-making about extreme events, where it is often worth making preparations even given a small probability of the event in question.

To remedy this situation, Relative Economic Value has been proposed by  \cite{Richardson_2006,richardson2000skill,wilks2001skill}, based on the well-studied cost-loss ratio decision model \citep{thompson1952operational,murphy1977value} for forecasts of binary events. This model allows us to assess the value of a forecast to a range of users facing different decision problems---from those who will act on a relatively small probability of a severe event, to those who will only act once the event is predicted with confidence.

In this model, a user must decide whether or not to prepare for an adverse weather event. Facing the event unprepared incurs an expense $L$ (the `loss'). However, this loss can be avoided by preparing for the event with an expense $C$ (the `cost'), as reflected in \cref{tab:app:cost-loss-matrix}:

\begin{table}[ht!]
\centering
\begin{tabular}{rcc}\toprule
     & Event doesn't happen & Event happens \\
    \cmidrule(r){2-2} \cmidrule(l){3-3}
    No preparation made & $0$ & $L$  \\
    Preparation made & $C$ & $C$  \\
    \bottomrule
\end{tabular}
\caption{\small\textbf{Expenses under the cost-loss decision model.}} \label{tab:app:cost-loss-matrix}
\end{table}

The optimal strategy is to take action whenever the probability of the event exceeds $C/L$. This `cost-loss ratio' is thus sufficient to characterise the decision problem faced by a particular user. Since it may vary significantly for different users, we display results for a range of cost-loss ratios. For severe weather events we will focus in particular on small cost-loss ratios (on the order of 0.01 to 0.2 for example) since these are more typical in practice, see \cite{katz1997economic} for a number of examples.

Under this model the expected expense of a decision system can be computed in terms of its confusion matrix, consisting of the proportions of true and false positives and negatives incurred (denoted TP, TN, FP, FN).

To make binary decisions based on our ensemble, we select a probability threshold $q$, and make a positive prediction whenever our empirical predictive probability $\bar{x}_{i,m}$ exceeds $q$, obtaining binary forecasts:
\begin{align}
x_{i,k} &:= \mathbb{I}[\bar{x}_{i,k} > q].
\end{align}
We then compute the confusion matrix weighted by grid cell area:
\begin{align}
\begin{bmatrix}
\text{TN} & \text{FN} \\
\text{FP} & \text{TP}
\end{bmatrix} &:= \frac1K \sum_k \frac1{|G|} \sum_i a_i
\begin{bmatrix}
(1-x_{i,k})( 1-y_{i,k}) & (1-x_{i,k}) y_{i,k} \\
x_{i,k} (1-y_{i,k}) & x_{i,k} y_{i,k}
\end{bmatrix}
\end{align}
Multiplying the entries of this confusion matrix with the expenses in  \cref{tab:app:cost-loss-matrix} and summing, we can then compute the expected expense incurred as:
\begin{align}
    E_{forecast} &= (\text{TP} + \text{FP})\cdot C + \text{FN} \cdot L.
\end{align}
To derive relative economic value (REV), we normalize this relative to the expense $E_{clim}$ incurred by a constant forecast (whose value is chosen based only on the climatological base rate of the event), and the expense $E_{perfect}$ incurred by a perfect forecast. In other words, REV compares how much expense you save using the forecasts instead of relying on climatology, relative to how much it would have been possible to save if you had the perfect forecast.

Never preparing will incur expense equal to either $(\text{TP} + \text{FN}) \cdot L$ (because $\text{TP} + \text{FN}$ is the base rate of the event), whereas always preparing will incur an expense $C$. Choosing the better of these strategies yields
\begin{align}
    E_{clim} &= \min\{(\text{TP} + \text{FN}) \cdot L, C\},
\end{align}
while with a perfect forecast, one only needs to prepare when the event is actually going to occur, incurring an expense 
\begin{align}
    E_{perfect} &= (\text{TP} + \text{FN}) \cdot C.
\end{align}
REV is then defined as 
\begin{align}
    \text{REV}(C/L, q) &:= \frac{E_{clim} - E_{forecast}}{E_{clim} - E_{perfect}}     \label{eq:app:rev},
\end{align}
and is 0 for a forecast based on climatology alone and 1 for a perfect forecast, so larger is better. Note that by dividing each term in \Cref{eq:app:rev} by $L$, the definition of REV depends on $C$ and $L$ only through the cost-loss ratio $C/L$.

It also depends on the probability threshold $q$ that was chosen earlier. When probabilities are perfectly calibrated, it is optimal to set $q=C/L$. In practice ensemble forecast systems are rarely perfectly calibrated, and we are more interested in the potential REV obtainable by a system after its probabilities have been optimally recalibrated. We follow \cite{richardson2000skill} in approximating this by computing REV at every possible threshold for the empirical predictive probability of our size-$M$ ensemble, and taking the maximum REV over these:
\begin{align}
    \text{REV}^*(C/L) &:= \max_{j=0, \ldots, M+1} \text{REV}\Big(C/L, q=\frac{j-1/2}{M}\Big)
\end{align}

It is this maximum or potential REV that we report throughout; for the sake of conciseness we refer to it just as REV.
Note that it is impossible to do worse than climatology on this metric, since both options available to the climatological forecast (never preparing and always preparing) are included in the maximization above. Thus $0 \leq \text{REV}^* \leq 1$ always holds, and it is common to see values equal to zero when a forecast does not improve on climatology.

\label{sec:app:statistical-tests}
\subsection{Statistical tests}

We compare \ourmodel with ENS on a number of verification metrics (detailed in \cref{sec:app:verification-metrics}) computed on our 2019 evaluation set.
For each relevant metric (and where applicable at each lead time, level, quantile and cost-loss ratio), we test the null hypothesis of no difference in the metric between \ourmodel and ENS, against the two-sided alternative. Specifically we are testing for differences in the values of the metrics that would be attained in the limit of infinite years of evaluation data, assuming stationarity of the climate.

Each of our verification metrics $V$ can be viewed as a function of the mean of a time-series of statistics given for every initialisation time $t$, where these statistics have already been spatially averaged. The statistics used for each metric are given in \cref{tab:app:metrics-statistics}.

\begin{table}[ht!]
\centering
\begin{tabular}{lll}%
    \toprule
    \textbf{Verification metric} & \textbf{Statistic time-series} & \textbf{Proxy used for block length selection}\\
    \midrule
    CRPS &  CRPS  &  CRPS \\
    Ensemble mean RMSE & Ensemble mean MSE & Ensemble mean MSE \\ 
    Brier skill score & Brier score, base rate &  BSS with constant denominator \\ 
    REV$^*$ & Confusion matrix & REV with constant denominator and \\
    & & $q$ maximizing overall REV. \\ 
    \bottomrule
\end{tabular}
\caption{\small\textbf{Time-series of statistics used by each verification metric.}} \label{tab:app:metrics-statistics}
\end{table}

To perform our test, we compute paired time-series of these statistics for both \ourmodel and ENS. We then resample the paired time-series $10000$ times using the stationary block bootstrap of \cite{politis1994stationary} implemented in \cite{arch-library}. We compute $V_{\ourmodel} - V_{ENS}$ for each resample, and use these values to construct a $(1-\alpha)*100$\% confidence interval for the difference using the bias-corrected and accelerated (`bca') method of \cite{efron2020automatic}. We reject the null hypothesis when this interval does not contain zero.

\paragraph{Block length selection}
In order to account for temporal dependence it is important to select an appropriate mean block length for the stationary block bootstrap. For this we use the automatic block length selection described in \cite{politis2004automatic,patton2009correction} and implemented by \cite{arch-library}. Block length selection is performed separately for each lead time, variable and level, and for each setting of the metric in question (such as cost-loss ratio or event threshold). Since the selection mechanism takes as input a univariate time-series and our paired time-series of statistics are multivariate, we compute a suitable per-initialisation-time proxy $V_k$ whose temporal mean is equal to or closely related to $V$, described in \cref{tab:app:metrics-statistics}. We then use differences $V_{\ourmodel,k} - V_{ENS,k}$ for block length selection.

\paragraph{Alignment of \ourmodel and ENS time-series}
As motivated in \cref{sec:app:ens_init_times}, ENS is initialised at 00/12 UTC, and \ourmodel at 06/18 UTC, meaning that forecast initialisation times do not match between the two time-series of statistics.
In order to align the two time-series when performing a paired test, we take one of two approaches:
\begin{itemize}
    \item Pair statistics for \ourmodel with statistics for ENS that are based on a forecast initialisation time 6 hours earlier, and a validity time 6 hours earlier too (so maintaining the same lead time).
    \item Pair statistics for \ourmodel with the mean of two statistics for ENS: one taken from a forecast initialised 6 hours earlier and so with 6 hours additional lead time; another taken from a forecast initialised 6 hours later and so with 6 hours less lead time. Both forecasts thus have the same validity time as the corresponding \ourmodel forecast, and the lead time is the same on average.
    This method is used only in the results on regional wind power forecasting, where a similar adjustment is performed for the reported metrics themselves; these adjustments are described and motivated in more detail in \cref{sec:app:ens_init_times}.
\end{itemize}

\paragraph{Missing data.}
For ENS, a single initialisation time is missing, as described in \cref{sec:app:ensnan}. The statistics from \cref{tab:app:metrics-statistics} are imputed for this initialisation time based on a linear interpolation of the previous and next values.

\subsection{Local surface extremes evaluation}

We evaluate \ourmodel and ENS on the task of predicting when surface weather variables exceed high (99.99th, 99.9th and 99th) and low (0.01st, 0.1st and 1st) climatological percentiles. These percentiles are computed per latitude/longitude using 7 recent years of 6-hourly data from 2016 - 2022, taken from the corresponding ground truth dataset for each model (ERA5 for \ourmodel and \hresfczero for ENS). For each latitude/longitude, they correspond to return periods of approximately 7 years, 8 months and 25 days respectively.

\subsection{Cyclone tracking}\label{sec:app:cyclones}
\subsubsection{Cyclone evaluation}\label{sec:app:cyclones:evaluation}

A common approach for evaluating \textit{deterministic} cyclone trajectory forecasts is to initialise the tracker from cyclones' initial locations and to continue the tracks until they predict the disappearance of the cyclone.
Then the track error is computed from these pairs of forecast cyclone tracks and ground truth cyclone tracks until one of them terminates~\citep{magnusson2021tropical}.
However, this approach does not measure ability to accurately predict cyclogenesis nor cyclone termination.
Furthermore, this approach cannot be applied to ensemble forecasts because the temporal duration of cyclones can differ between ensemble members. %

We use a probabilistic cyclone evaluation which overcomes these issues.
First, we apply a cyclone tracker that processes forecasts independently from ground truth and captures cyclogenesis.
We then extract strike probabilities from these cyclone tracks.
Strike probability is a common output in probabilistic cyclone modelling~\citep{magnusson2021tropical}, measuring the chance that a given location on earth has a cyclone passing within a certain distance at a given time. 
We compute REV from these probabilities, which penalises incorrectly predicting the existence or non-existence of a cyclone, as well as errors in predictions of their positions. This probabilistic approach is well aligned with decision making, where it is vital to accurately know the risk of being affected by an incoming cyclone.

More specifically, we apply the same cyclone tracker, TempestExtremes (\cref{sec:app:cyclones:tempest_extremes}), to each forecast generated by ENS and \ourmodel, downsampling ENS from a 6-hourly to 12-hour resolution for a fair comparison with \ourmodel.
From these tracks we generate \ang{1} resolution probability heatmaps for each time step, where the probability in each \ang{1} cell is the fraction of ensemble members predicting a cyclone centre within that cell. We choose \ang{1} as it corresponds to 111km at the equator, which is close to 120km, a common radius used for defining cyclone strike probability~\citep{magnusson2021tropical}. We then run the tracker on chunks of HRES-fc0 and ERA5 of the same length as the forecast trajectories, yielding binary ground truth maps for each initialisation time and lead time. Finally, Relative Economic Value (REV) is computed for each model's forecasts using their respective ground truth.

As mentioned in the IBTrACS paper~\citep{knapp2010ibtracs}, different cyclone trackers apply different detection criteria in the definition of cyclones, which can affect results substantially, confounding the evaluation of the accuracy of the underlying gridded predictions. Our use of the TempestExtremes cyclone tracker controls for this.
TempestExtremes is an open-source tracker which requires physically consistent cyclone behaviour in the raw forecasts~(\cref{sec:app:cyclones:tempest_extremes}).
We apply this same tracker in the same way to all models and ground truth analysis datasets (without tuning the tracker to optimise performance on a particular model).
Our evaluation therefore minimises bias toward any particular model, isolating the relative quality of the raw forecasts (as opposed to the quality or sensitivity of the tracker or source of ground truth).

\begin{figure}[h]
    \centering
    \includegraphics[width=\textwidth]{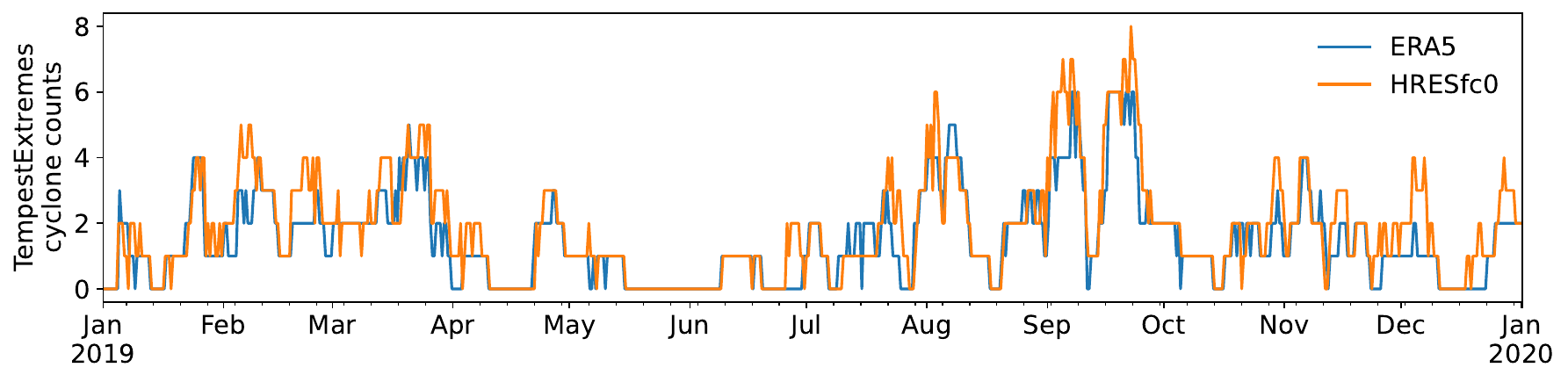}
    \caption{Per-timestep TempestExtremes cyclone counts for ERA5 and HRES-fc0. The two time-series exhibit high correlation, but HRES-fc0 has $23\%$ more cyclones than ERA5. The TempestExtremes tracker is applied to these analysis datasets as described in section~\ref{sec:app:cyclones:tempest_extremes}. We then arbitrarily picked a lead time of 4 days to extract cyclone count (all lead times 0-10 days yield very similar results).}
    \label{fig:cyclone_counts}
\end{figure}

We note that due to differences between ERA5 and HRES-fc0, running the TempestExtremes cyclone tracker on each dataset produces different cyclone locations and counts.
On average across 2019, ERA5 has 1.51 cyclones per timestep and HRES-fc0 has 1.85 cyclones per timestep, which gives HRES-fc0 a base rate that is 23\% greater than ERA5. \cref{fig:cyclone_counts} plots the total cyclone counts tracked in the ground truth dataset at each time of the year, for both ERA5 and HRES-fc0. However, REV accounts for this difference in base rates by virtue of its normalisations with respect to climatology and the perfect forecast (see \cref{sec:app:rev}), and is thus a fair metric to use when comparing methods evaluated against different ground truths. Furthermore, \ourmodel still outperforms ENS beyond 1~day lead times even when using HRES-fc0 ground truth, which puts \ourmodel at a disadvantage (\cref{fig:cyclones_revs_vs_fc0}).

\subsubsection{Cyclone tracker}\label{sec:app:cyclones:tempest_extremes}

To extract cyclone trajectories from gridded forecasts and analysis datasets, we use the TempestExtremes v2.1 cyclone tracking algorithm \citep{ullrich2021tempestextremes}.
TempestExtremes is open-source on GitHub\footnote{\url{https://github.com/ClimateGlobalChange/tempestextremes}} and has been used in a wide range of cyclone studies (\citealt{ullrich2021tempestextremes}, Section 3.1). %
The algorithm has two stages described below (readers are referred to Section 3.2 of \citealt{ullrich2021tempestextremes} and \citealt{ullrich2017tempestextremes} for full details).

The first stage, \textit{DetectNodes}, finds candidate tropical cyclones where minima in mean sea level pressure (MSL) are co-located with upper-level warm cores:
\begin{itemize}
\itemsep0em 
    \item Initial candidate locations are determined by local minima in MSL. Candidates within \ang{6}\xspace of another stronger MSL minimum are eliminated.
    \item Each MSL minimum must be surrounded by a closed contour of MSL that is 200 hPa greater than the minimum. The minima must be sufficiently compact with this closed contour falling within a \ang{5.5}\xspace great circle distance of the MSL minimum.
    \item A warm core criterion checks that candidate storms are co-located with maxima in the geopotential thickness field between \SI{500}{hPa} and \SI{300}{hPa}, Z500 - Z300. The maxima must be enclosed by a closed contour of thickness that is \SI{58.8}{m^2s^{-2}} less than the maximum. The maxima must be sufficiently compact with this closed contour falling within a \ang{6.5} great circle distance from the maximum.
\end{itemize}

The second stage, \textit{StitchNodes}, detects plausible cyclone trajectories from these candidate cyclone locations by linking them together in time with the following criteria:
\begin{itemize}
\itemsep0em 
    \item Candidates cannot move more than an \ang{8}\xspace great circle distance between subsequent time slices (the `stitch range').
    \item Trajectories are allowed a maximum gap of 24 hours between candidate nodes; greater gaps result in a trajectory terminating (and potentially a new trajectory forming).
    \item To filter out short-lived storms that are not tropical cyclones, trajectories must last for at least 54 hours.
    \item A threshold analysis is performed on candidate trajectories to ensure cyclones are sufficiently intense and in the right locations on Earth: each trajectory must have at least 10 time slices with wind speed greater than \SI{10}{ms^{-1}}, elevation below \SI{150}{m}, and latitude between \ang{-50}\xspace and \ang{50}\xspace.
\end{itemize}

The algorithm's default hyper-parameters were chosen so that when TempestExtremes is applied to 6-hourly reanalysis datasets the resulting tracks closely match observed tracks \citep{zarzycki2017assessing}, using the IBTrACS dataset as ground truth \citep{knapp2010ibtracs}. To account for the 12-hourly (instead of 6-hourly) temporal resolution in our evaluation, we changed two of the tracker's \textit{StitchNodes} hyperparameters.
Firstly, we halved the number of time slices for the criteria checks from 10 to 5.
Secondly, because cyclones can travel further in a 12 hour interval than in 6 hours, we increased the stitch range from an \ang{8}\xspace to a \ang{12}\xspace great circle distance.
This value was chosen from visual inspection to trade-off fast-moving cyclones being cut off (if the stitch range is too small) and trajectories jumping between separate storms (if the stitch range is too large).
Increasing the stitch range to \ang{12}\xspace can sometimes result in nearby but separate storms being connected as part of the same cyclone trajectory (for example, in \cref{fig:app:visualization_cyclone_dorian}g). However, we found this to be rare, and do not expect it to bias the results towards one particular model.
All other TempestExtremes hyperparameters were left as their default values, and the same hyperparameters were used for each model and each analysis dataset.

Since TempestExtremes performs a global optimisation when stitching nodes, the tracker output at a particular lead time depends on raw predictions at nearby lead times.
We prepend 10 days of the model's respective ground truth (ERA5 or HRES-fc0) to each forecast before running the cyclone tracker. 
This avoids cyclones being dropped when forecasts are initialised close to the end of a cyclone's lifetime, due to the short duration of cyclone within the forecast period not passing the tracker's criteria.
Similarly, we only report results up to lead times of 9 days despite providing 15 days of predictions to the tracker, because the tracker may drop cyclones that begin close to the end of the forecast period.

\subsection{Spatially pooled evaluation}\label{sec:app:pooled_evaluation_description}

\begin{figure}
    \centering
    \includegraphics[width=\textwidth]{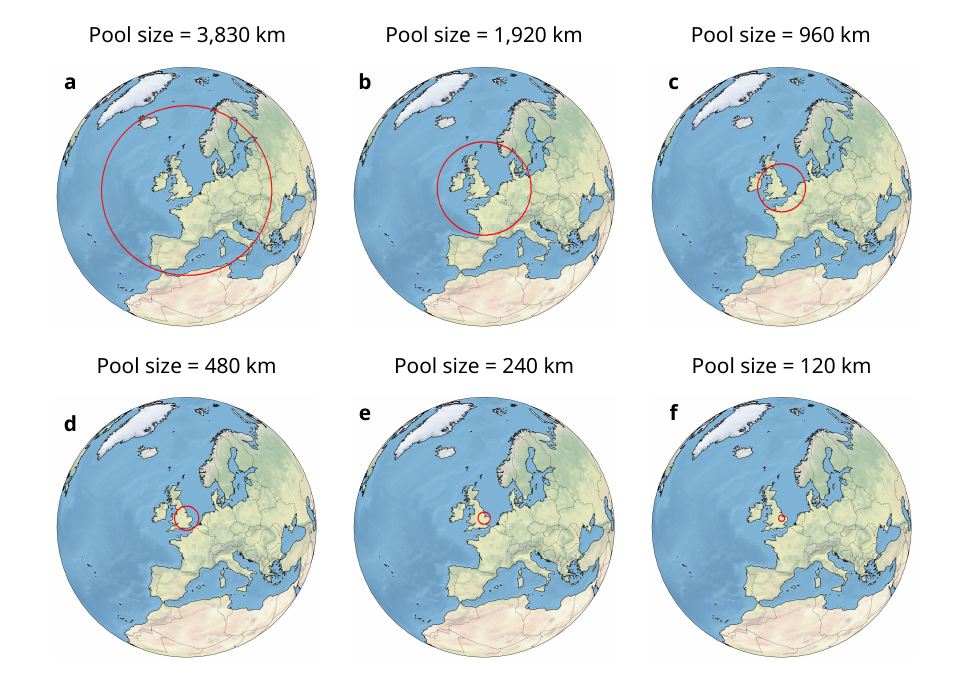}
    \caption{Illustration of pool sizes. The sizes refer to the length of the shortest route (i.e.~geodesic distance) from one side of the pooling region through the centre to the opposite side.}
    \label{fig:pool_sizes}
\end{figure}

The meteorological literature has produced a wide range of methods to verify spatial structure in weather forecasts \citep{gilleland2009intercomparison}.
We use a neighbourhood (or `fuzzy') approach,
where forecasts and targets are first aggregated over regions of a particular spatial scale and standard skill scores computed on these pooled counterparts, with the process repeated at a range of spatial scales \citep{ebert2008fuzzy, ravuri2021skilful}.
Neighbourhood methods often perform pooling using squares on a 2D latitude-longitude grid, which can lead to undesirable behaviour towards the poles and is not appropriate for evaluating global weather forecasts.
Instead, we perform pooling on the surface of a sphere. We define pool centres as the nodes of a $k$-times refined icosahedral mesh, which are distributed approximately uniformly over the Earth's surface. Pooling regions are defined within a fixed geodesic distance of each pool centre, with radii set to the mean distance between mesh nodes.
To capture performance at different spatial scales, we do this separately for 6 mesh refinement levels ($k = 7, 6, \dots, 2$),
resulting in a wide range of pool sizes: \SI{120}{km}, \SI{241}{km}, \SI{481}{km}, \SI{962}{km}, \SI{1922}{km}, and \SI{3828}{km} (\cref{fig:pool_sizes})\footnote{For simplicity, we round these pool sizes to the nearest 10 in text and figures.}, ranging from mesoscales to planetary scales.

\begin{figure}
    \centering
    \includegraphics[width=\textwidth]{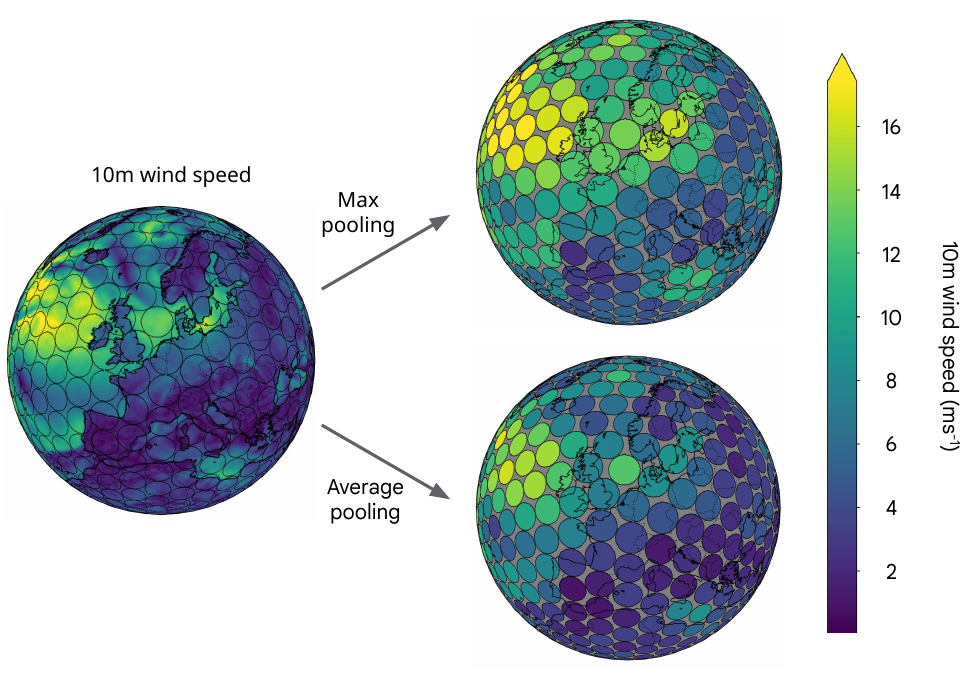}
    \caption{Illustration of max-pooling and average-pooling on an icosahedral mesh. The pool size is \SI{480}{km}. The overlap between pooling regions has been removed for visualisation purposes (\cref{sec:app:pooled_evaluation_description}).}
    \label{fig:mesh_pooling_illustration}
\end{figure}

We use two standard aggregation methods from the neighbourhood verification literature: average-pooling and max-pooling (\cref{fig:mesh_pooling_illustration}), which assess how well models predict statistics of the spatial distributions.
Grid cells are weighted by their area when average-pooling.
We then compute globally averaged CRPS from the pooled forecasts and ground truth. 
To account for slight non-uniformities in the distribution of pooling centres, we weight each pooling region by the area of the pooling centre's Voronoi cell. 

Average-pooling behaves like a low-pass filter, focusing the evaluation on increasingly large spatial structures as pooling size increases, and reducing the double-penalty problem at scales smaller than the pooling size. Max-pooling emphasises the positive tail of the spatial distribution, which may include high-impact events such as extreme wind or temperature that would otherwise be blurred by average-pooling.
In some cases, forecast users perform max-pooling to account for possible location error, such as the location of a weather front for wind power ramp prediction \citep{drew2017importance}.

\begin{figure}[H]
    \centering
    \includegraphics[width=\textwidth]{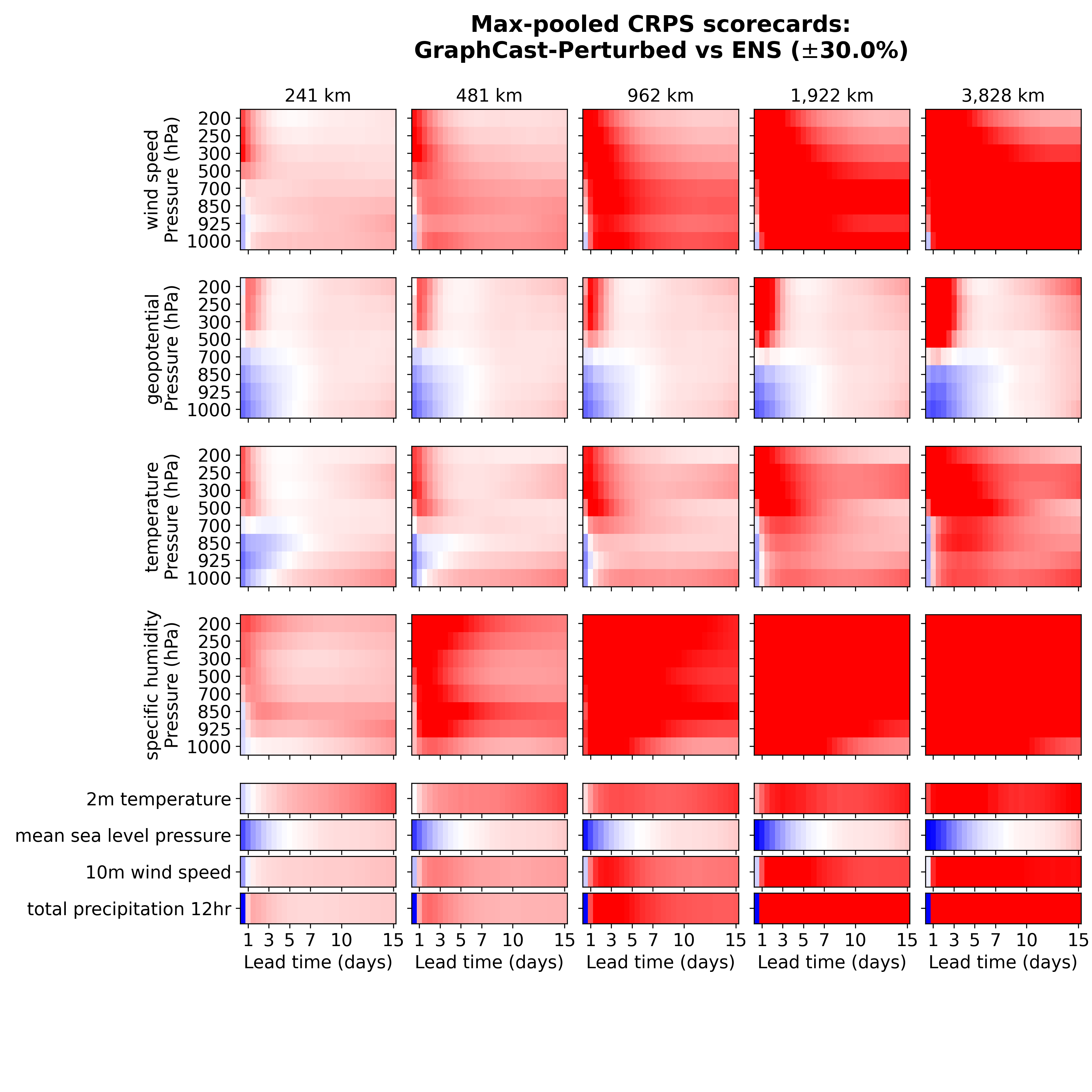}
    \caption{Max-pooled CRPS scorecard comparing \gcens and ENS at varying spatial scales, dark blue (resp. red) means \gcens is 30\% better (resp. worse) than ENS, and white means they perform equally.}
    \label{fig:pool_scorecard_graphcast_perturbed_max_low_res}
\end{figure}

Average- and max-pooled CRPS scores are computed on key surface variables at \ang{0.25}\xspace: \SI{2}{m} temperature, \SI{10}{m} wind speed, 12h accumulated precipitation, and mean sea level pressure at \modelresolution (\cref{fig:panel_applications}a, \cref{fig:pool_skillscore_average}, \cref{fig:pool_skillscore_max}).
We also compute pooled CRPS scorecards for wind speed, geopotential, temperature, and specific humidity at all pressure levels (\cref{fig:pool_scorecard_gencast_average_low_res}--\cref{fig:pool_scorecard_graphcast_perturbed_max_low_res}).
To reduce the computational cost of computing multiple pooled scorecards, forecasts and targets were subsampled to \ang{1}\xspace before pooling.
We skipped the smallest pool size because \SI{120}{km} corresponds approximately \ang{1}\xspace at the equator, making it similar to a univariate evaluation of the subsampled forecasts.

\subsection{Regional wind power forecasting evaluation}\label{sec:app:wind_power_evaluation}

\begin{figure}[H]
    \centering
    \includegraphics[width=0.5\textwidth]{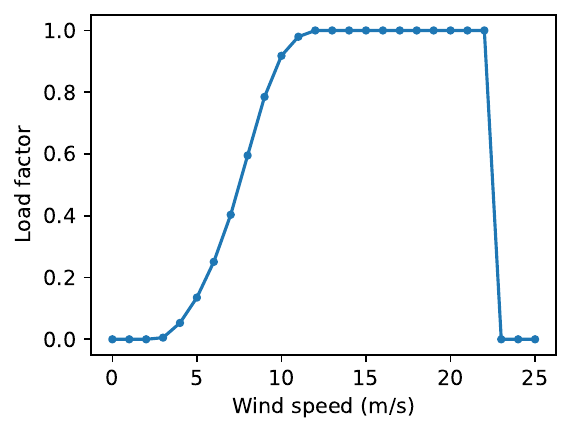}
    \caption{International Electrotechnical Commission Class II turbine power curve from the WIND Toolkit \citep{king2014validation}.}
    \label{fig:wind_turbine_power_curve}
\end{figure}

\begin{landscape}
\begin{figure}
    \centering
    \includegraphics[height=0.75\textheight]{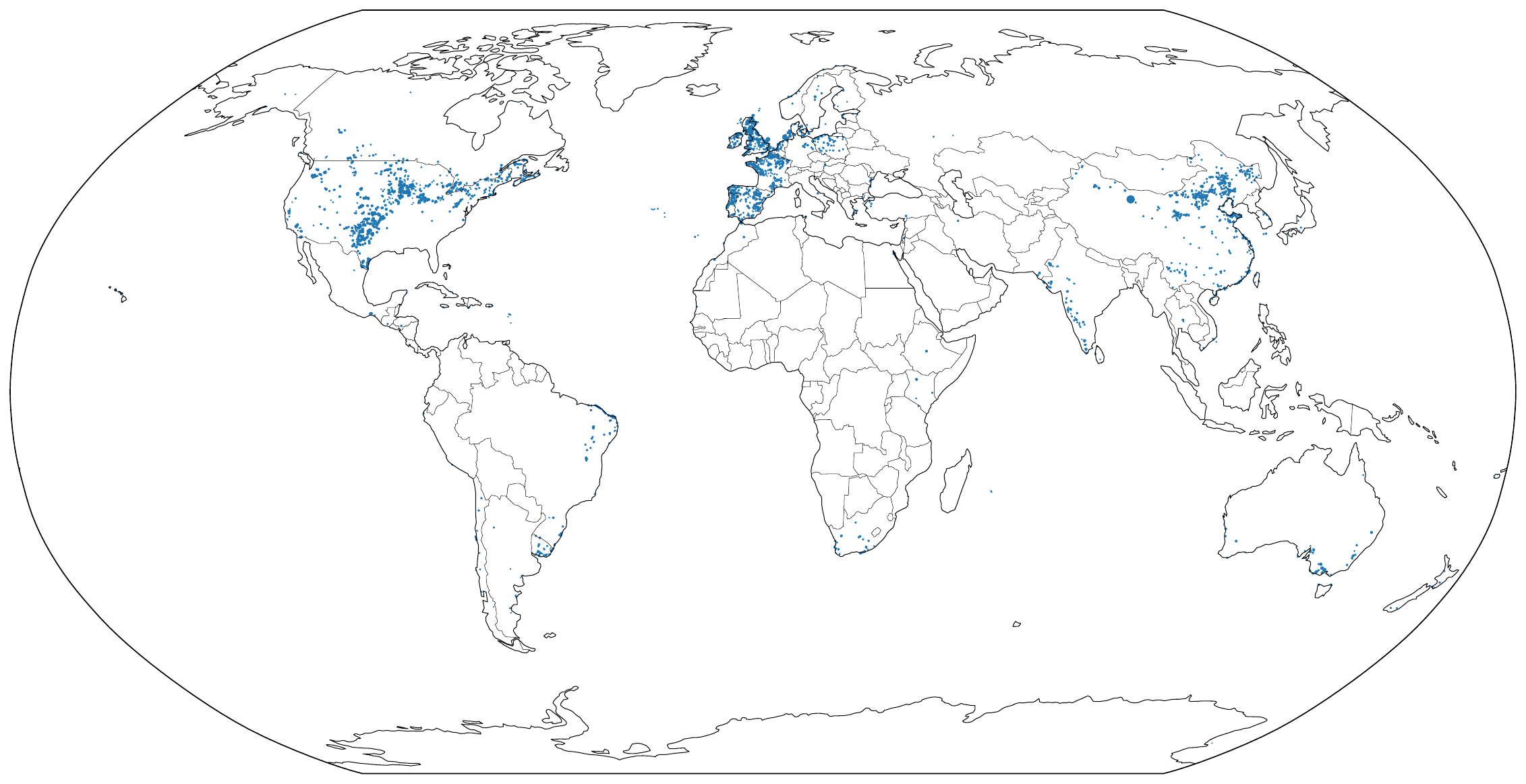}
    \caption{Wind farm locations in the Global Power Plant Database \citep{byers2018global} shown as blue markers, with marker size scaling with wind farm nominal capacity.}
    \label{fig:gppd_wind_farm_locs}
\end{figure}
\end{landscape}

For the regional wind power forecasting experiment, we use all 5344 wind farm locations and their nominal capacities from the Global Power Plant Database (GPPD, \citealt{byers2018global},  \cref{fig:gppd_wind_farm_locs})\footnote{GPPD captures $\sim$40\% of all global wind farm capacity as of 2020 \citep{byers2018global}.}.
We first bilinearly interpolate \SI{10}{m} wind speed forecasts and analysis states at each wind farm location.
We then map \SI{10}{m} wind speed to load factor---the ratio between actual wind turbine power output and maximum power output---using an idealised International Electrotechnical Comission Class II 2MW turbine power curve from the WIND Toolkit \citep{king2014validation}.
This power curve has a cut-in speed of \SI{3}{ms^{-1}}, maximum output at \SI{14}{ms^{-1}}, and curtailment at \SI{25}{ms^{-1}} (\cref{fig:wind_turbine_power_curve}).
Load factor is then multiplied by nominal capacity to obtain idealised power generation in megawatts at each wind farm.
This assumes no turbines are switched off due to maintenance or oversupply. Accounting for these additional complexities is is beyond the scope of this study.

To generate random groupings of wind farms across the globe at a range of spatial scales,
we use a similar procedure to the pooled evaluation in \cref{sec:app:pooled_evaluation_description}.
Pooling centres are defined on a 7-times refined icosahedaral mesh and
separate evaluations performed using pool sizes of \SI{120}{km}, \SI{240}{km}, and \SI{480}{km}.
The \SI{120}{km} scale contains 3648 groups with a mean capacity of \SI{272}{MW}, the \SI{240}{km} scale contains 7759 groups with a mean capacity of \SI{513}{MW}, and the \SI{480}{km} scale contains 15913 groups with a mean capacity of \SI{996}{MW}. 
Power output is summed over wind farm sites within each group and CRPS is computed for this derived quantity.
We then compute average CRPS across all wind farm groups.
By using power as the target variable, more weight is applied to pools containing more wind farm capacity in the global average CRPS.

\FloatBarrier

\clearpage

\section{Supplementary results}\label{sec:app:supplementary_results}

\subsection{CRPS line plots}\label{sec:app:headline_crps}

\begin{figure}[ht!]
    \centering
    \includegraphics[width=\textwidth]{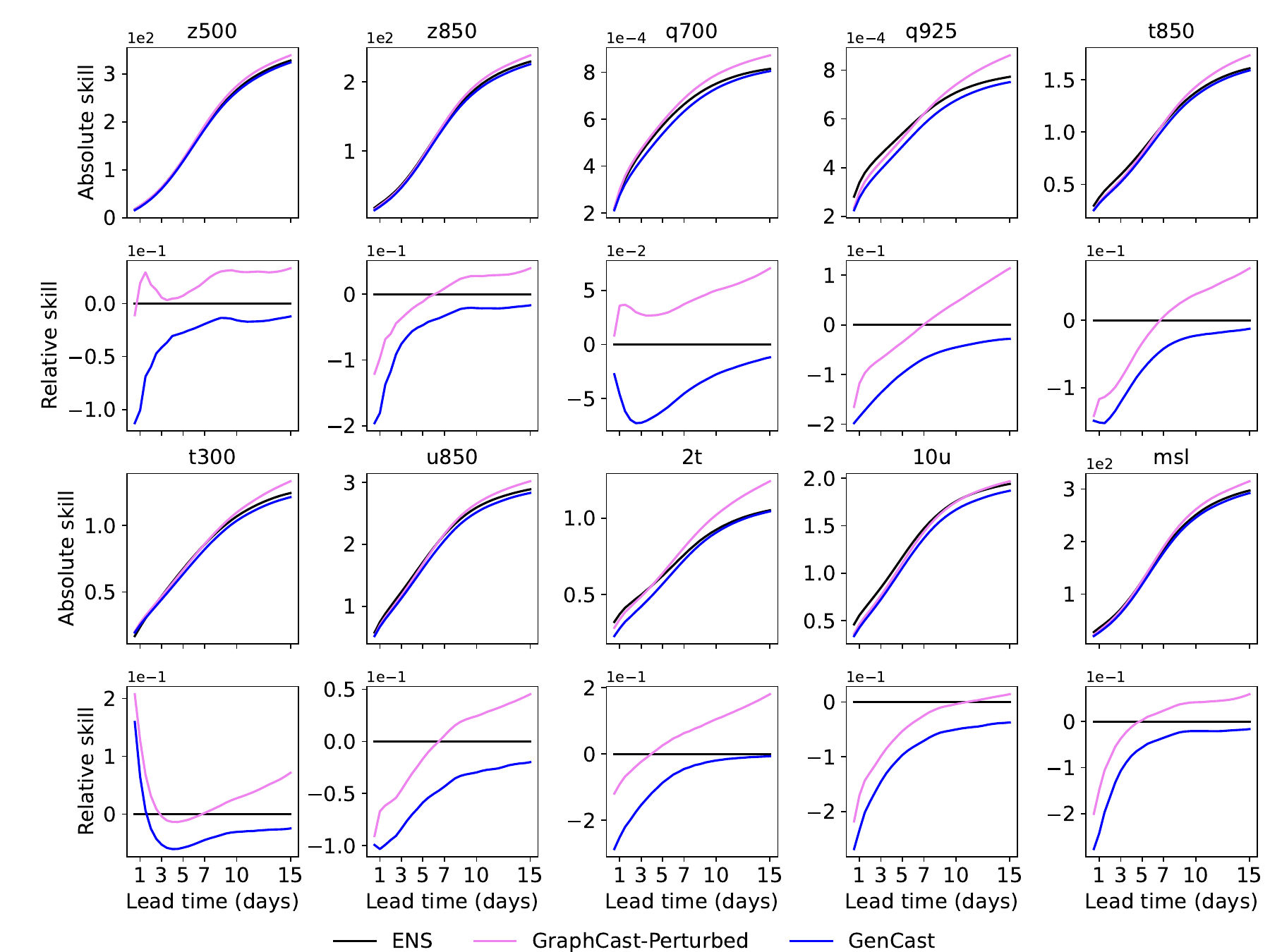}
    \caption{Absolute and relative CRPS model comparison for z500, z850, q700, q925, t850, t300, u850, 2t, 10u and msl.}
    \label{fig:headline_crps}
\end{figure}

We provide line plots on a set of representative variables for easier quantitative comparisons across models (\cref{fig:headline_crps}).

\subsection{Ensemble calibration}\label{sec:app:calibration}

\begin{figure}[H]
    \centering
    \includegraphics[width=\textwidth]{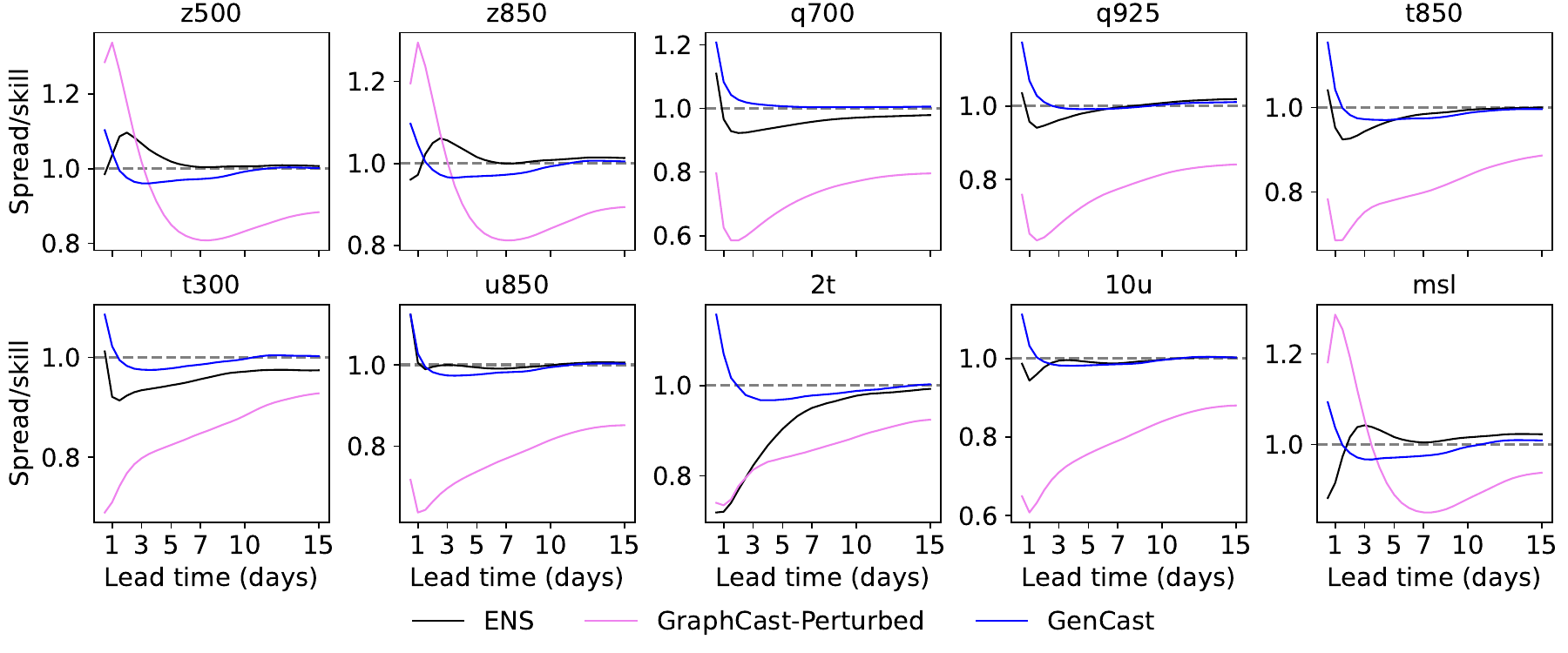}
    \caption{Spread/skill ratio model comparison for z500, z850, q700, q925, t850, t300, u850, 2t, 10u and msl.}
    \label{fig:headline_spread_skill}
\end{figure}

We find that, generally, \ourmodel's spread/skill and rank histograms tend to be as good, or better, than ENS's, and much better than that of \gcens.  \cref{fig:headline_spread_skill} compares spread/skill scores across lead times for a set of different variables and pressure levels. \ourmodel exhibits spread-skill scores close to 1 from 2-3 days onward. As noted in \cref{sec:baselines}, over-dispersion at short lead times (as exhibited by \ourmodel) is an expected consequence of evaluating forecasts with uncertain initial conditions against a deterministic analysis ground truth. Spread/skill for ENS is also mostly quite close to one, with some exceptions like t300, q700, and 2t where it exhibits some under-dispersion. \gcens shows substantial under-dispersion across almost all variables and lead times.

\begin{figure}[H]
    \centering
    \includegraphics[width=\textwidth]{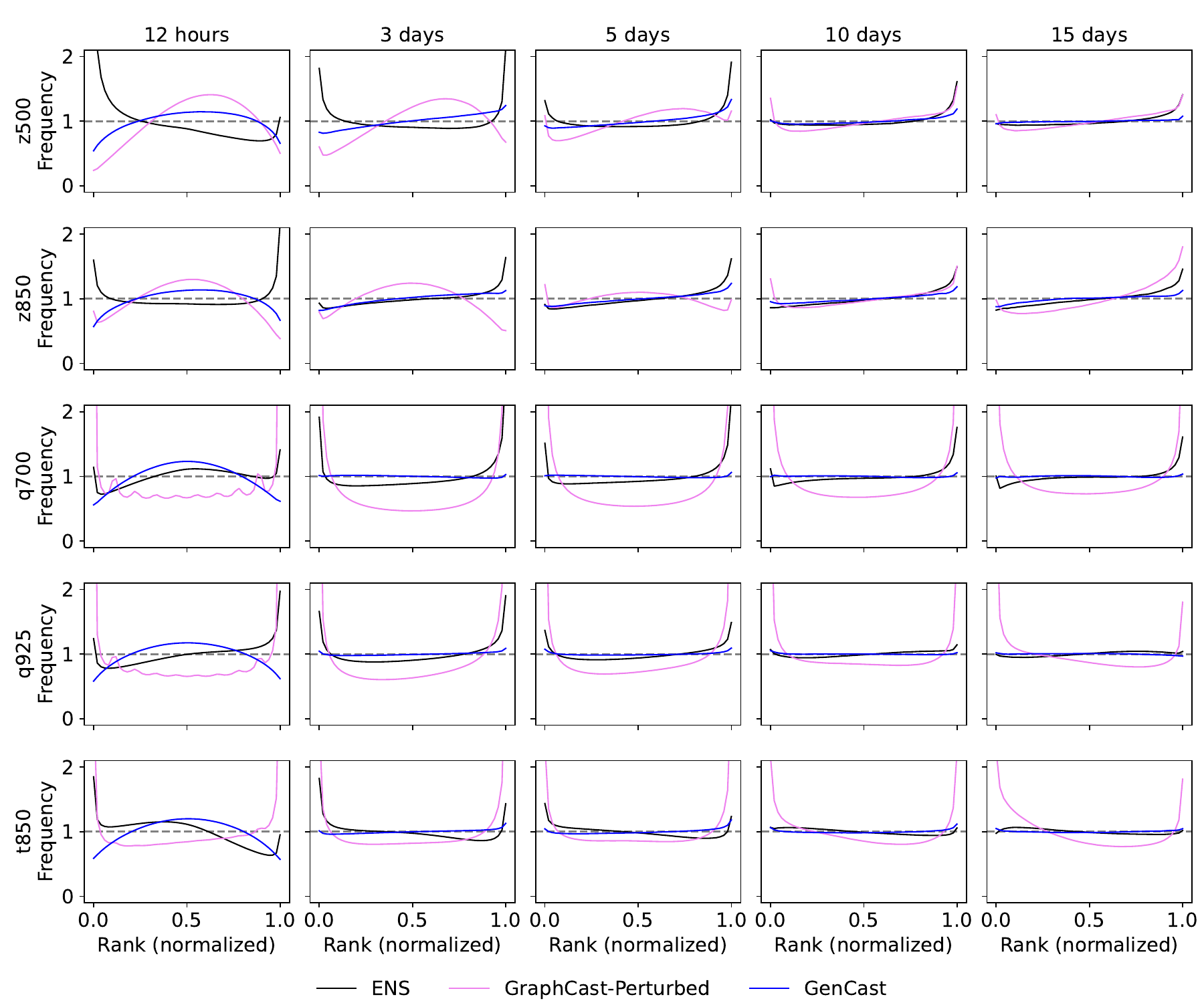}
    \caption{Rank histogram model comparison for z500, z850, q700, q925 and t850.}
    \label{fig:headline_rank_hist_group_0}
\end{figure}

\begin{figure}[H]
    \centering
    \includegraphics[width=\textwidth]{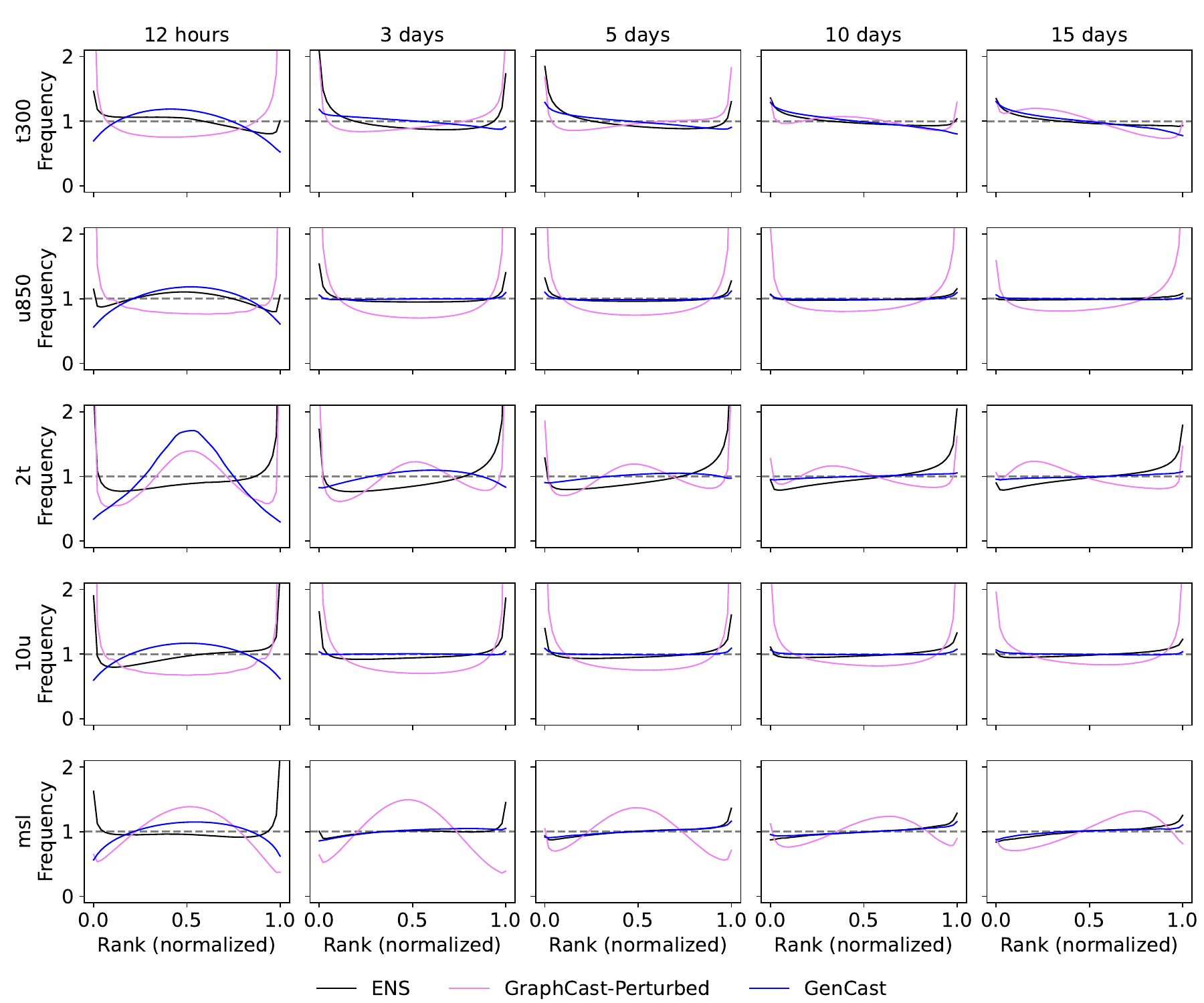}
    \caption{Rank histograms model comparison for t300, u850, 2t, 10u and msl.}
    \label{fig:headline_rank_hist_group_1}
\end{figure}

The rank histograms in \cref{fig:headline_rank_hist_group_0} and \cref{fig:headline_rank_hist_group_1} show that \ourmodel generally has very flat rank histograms from between 1 and 3 days lead time onward, in most cases flatter than those of ENS, and in all cases flatter than \gcens.

\subsection{Ensemble-Mean RMSE}\label{app:sec:emrmse_results}
\begin{figure}[H]
    \centering
    \includegraphics[width=\textwidth]{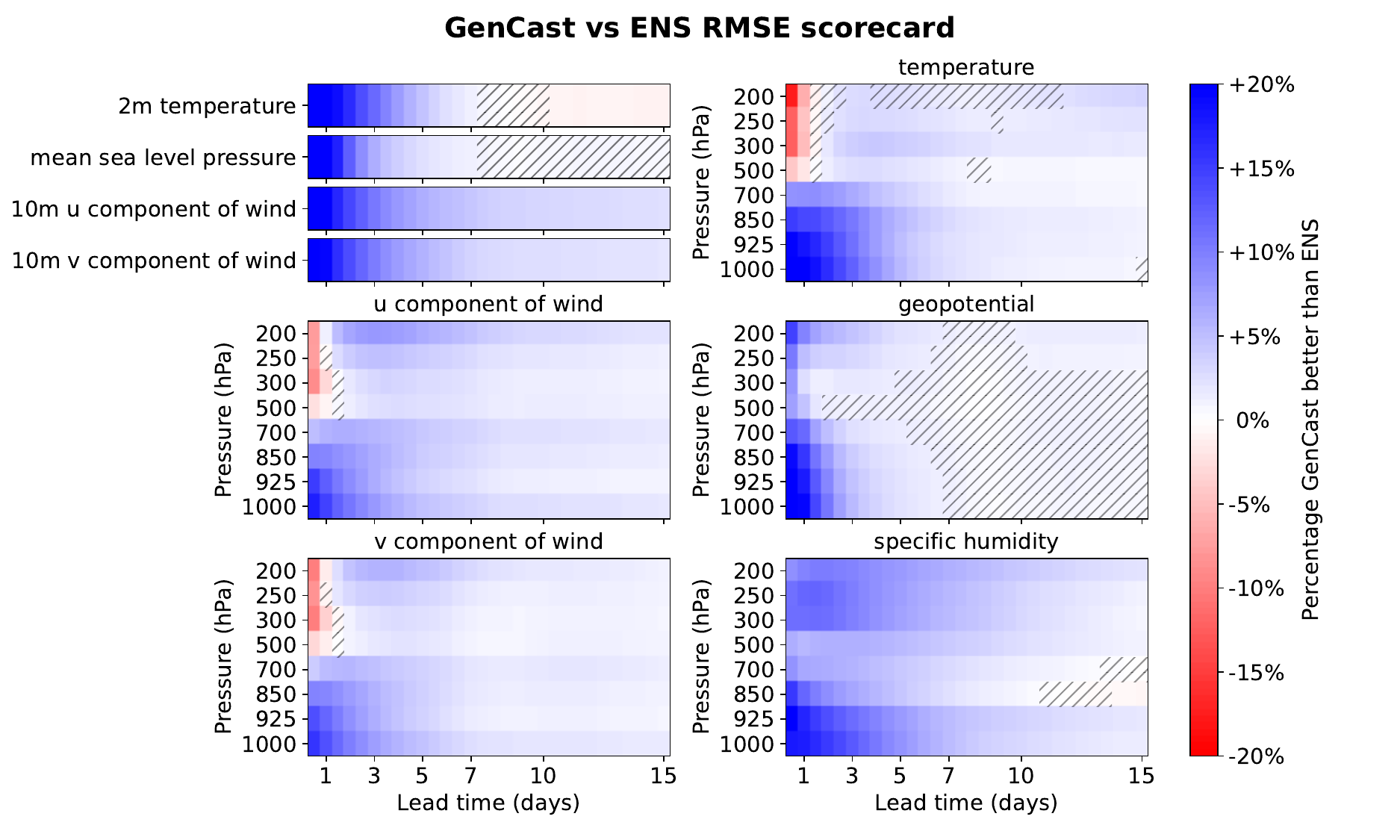}
    \caption{Ensemble-mean RMSE scorecard comparing \ourmodel to ENS. Analogous to \cref{fig:skill} but displaying RMSE instead of CRPS. Hatched regions indicate where neither model is significantly better ($p > 0.05$).}
    \label{fig:app:emrmse_gencast_vs_ens}
\end{figure}

\begin{figure}[H]
    \centering
    \includegraphics[width=\textwidth]{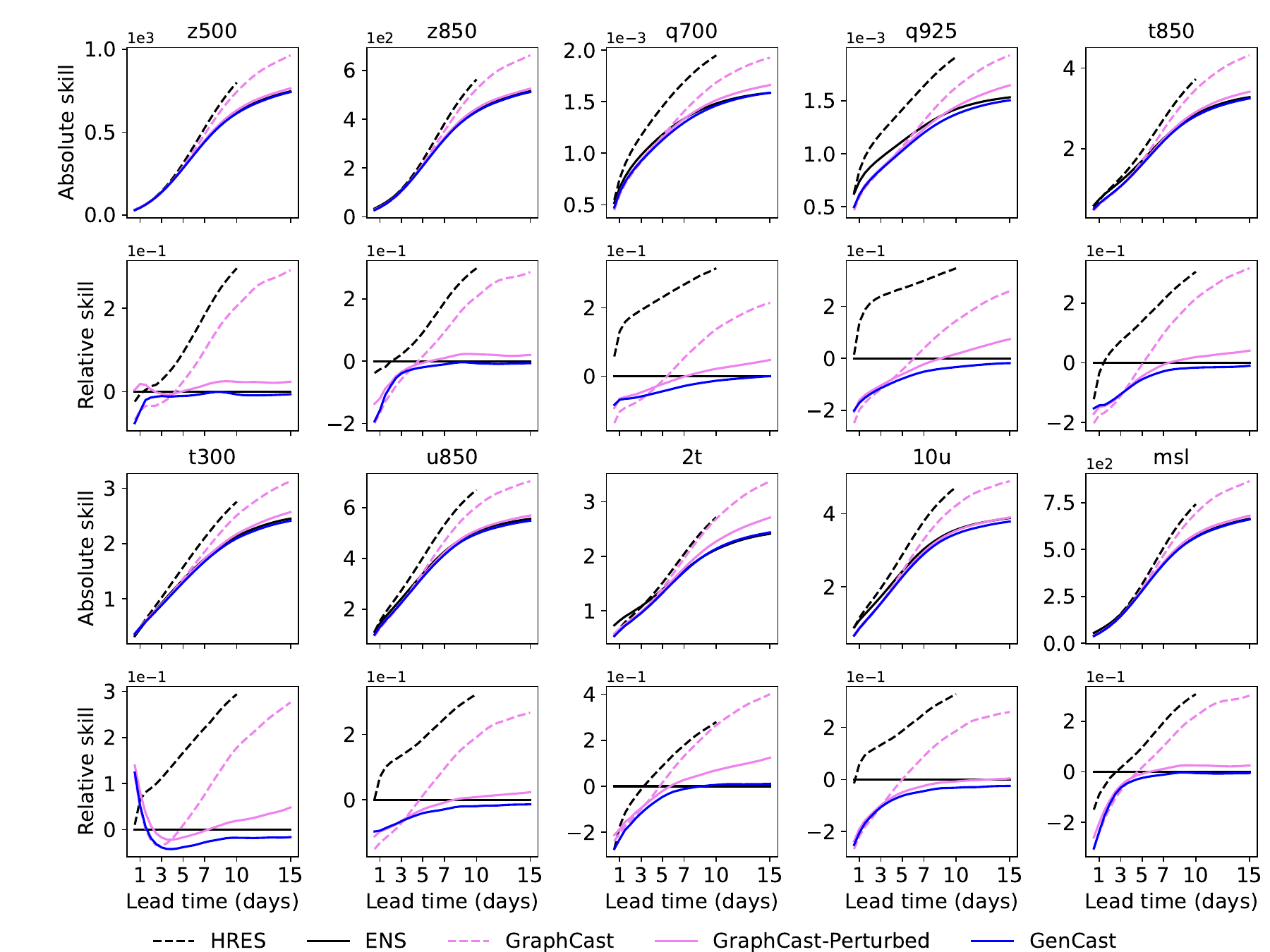}
    \caption{Absolute and relative RMSE (Ensemble mean) model comparison for z500, z850, q700, q925, t850, t300, u850, 2t, 10u and msl.}
    \label{fig:headline_rmse}
\end{figure}

Scorecards comparing the Ensemble-Mean RMSE are shown in \cref{fig:app:emrmse_gencast_vs_ens}. The relative performance of \ourmodel compared to ENS is similar on Ensemble-Mean RMSE to what is observed for CRPS.  \ourmodel is as good or better than ENS in 96\% of cases, and significantly better ($p < 0.05$) in 82\% of cases. We also present some line plots for a subset of representative variables for easier quantitative comparison (\cref{fig:headline_rmse}).

\subsection{Precipitation}\label{sec:app:precip}

\begin{figure}[ht!]
    \centering
    \includegraphics[width=\textwidth]{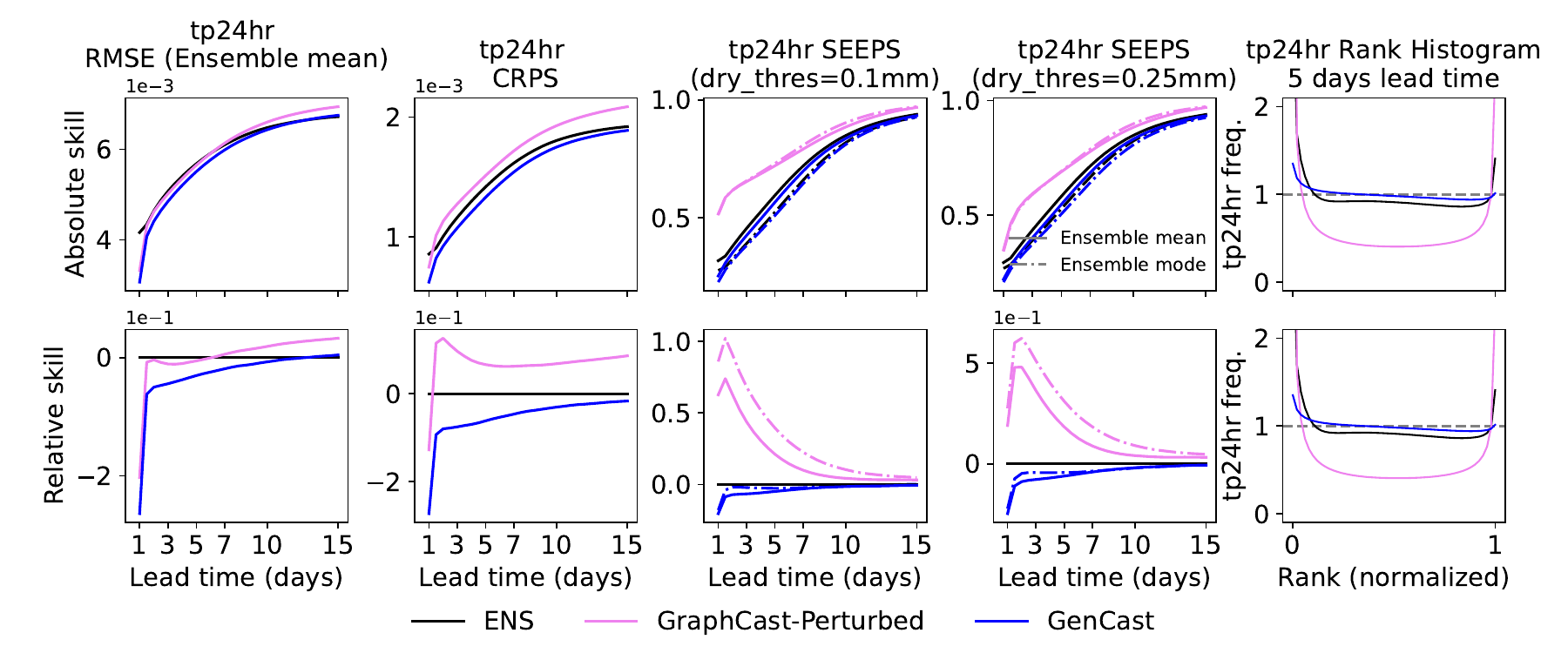}
    \caption{Comparing preliminary results of model performance on predicted total accumulated precipitation over 24 hours, evaluated on Ensemble-Mean RMSE, CRPS, SEEPS with dry thresholds 0.1 and 0.25, and via rank-histogram. All metrics were calculated globally and over the full test period, except for SEEPS which excludes very dry regions according to the criteria in \cite{lam2023learning}. For ensembles, we evaluate SEEPS both on the mean of the ensemble, as well as using the mode of the dry, light-rain, and heavy-rain categories predicted by the different ensemble members (i.e. majority vote). Relative and absolute plots are shown in the top and bottom rows respectively. \ourmodel shows promising results, more often than not outperforming ENS and with a significantly flatter rank histogram.}
    \label{fig:precip}
\end{figure}

\begin{figure}[ht!]
    \centering
    \includegraphics[width=\textwidth]{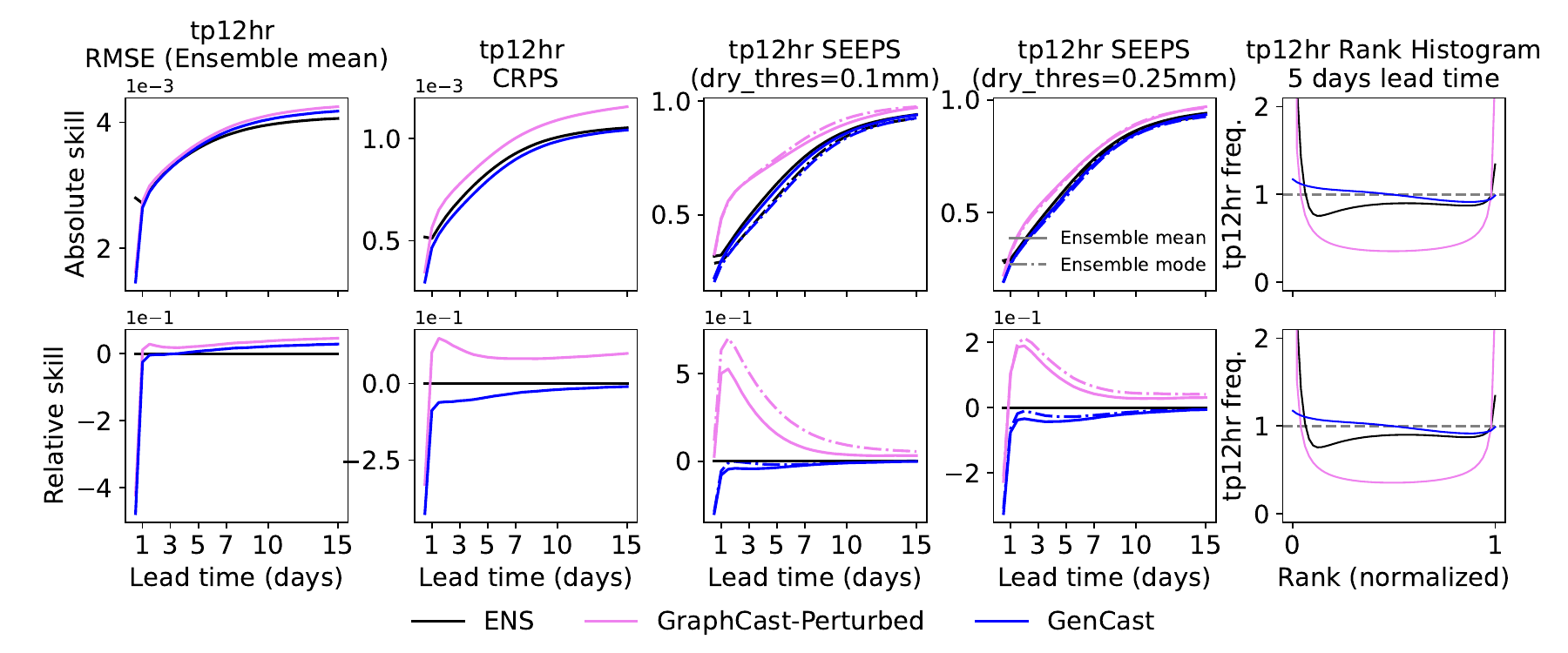}
    \caption{Results analogous to \cref{fig:precip} on 12 hour precipitation.}
    \label{fig:precip12h}
\end{figure}

In line with the rest of the paper, we evaluate precipitation forecasting skill using RMSE, CRPS and rank histograms. Since some of these metrics may not be representative of actual skill due to precipitation being highly sparse and very non-Gaussian, we also evaluated it using Stable Equitable Error in Probability Space (SEEPS) \citep{rodwell2010new,haiden2012intercomparison,north2013assessment}, using the same methodology as in \citet{lam2023learning}. Because SEEPS is a metric for deterministic categorical forecasts (dry, light-rain, and heavy-rain) we evaluated the ensembles both computing the category of the ensemble mean, as well as the mode of the ensemble category members (i.e. majority vote).  We repeat the caveats mentioned in the main text that we lack full confidence in the quality of ERA5 precipitation data, and that we have not tailored our evaluation to precipitation specifically beyond adding SEEPS as a further metric. Results for 24 hour and 12 hour accumulated precipitation are shown in \cref{fig:precip}  and \cref{fig:precip12h} respectively.

\clearpage

\subsection{Spectrum for additional variables}\label{sec:app:spectra}

\begin{figure}[H]
    \centering
    \includegraphics[width=\textwidth]{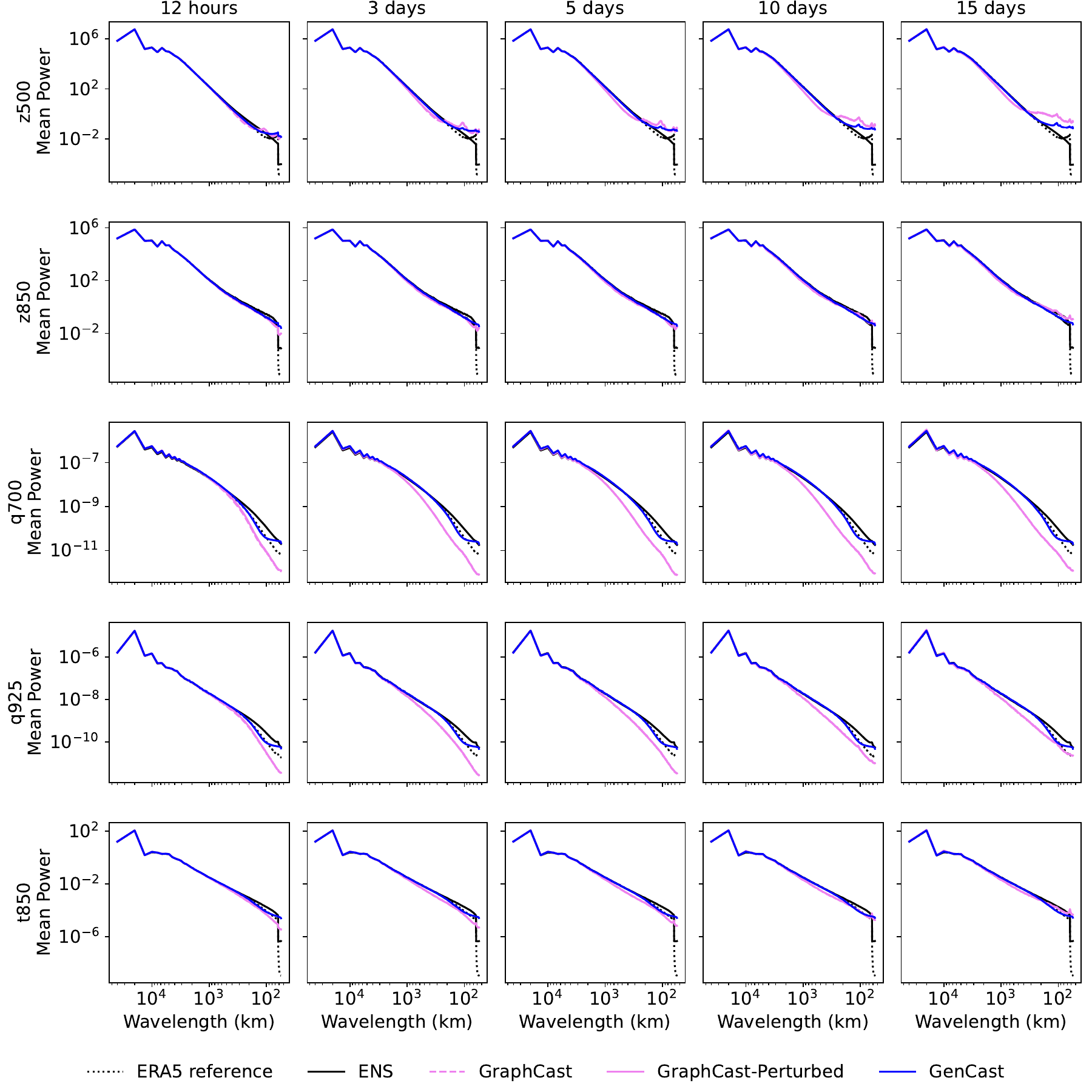}
    \caption{Power spectrum plots for z500, z850, q700, q925 and t850.}
    \label{fig:headline_spectra_group_0}
\end{figure}

\begin{figure}[H]
    \centering
    \includegraphics[width=\textwidth]{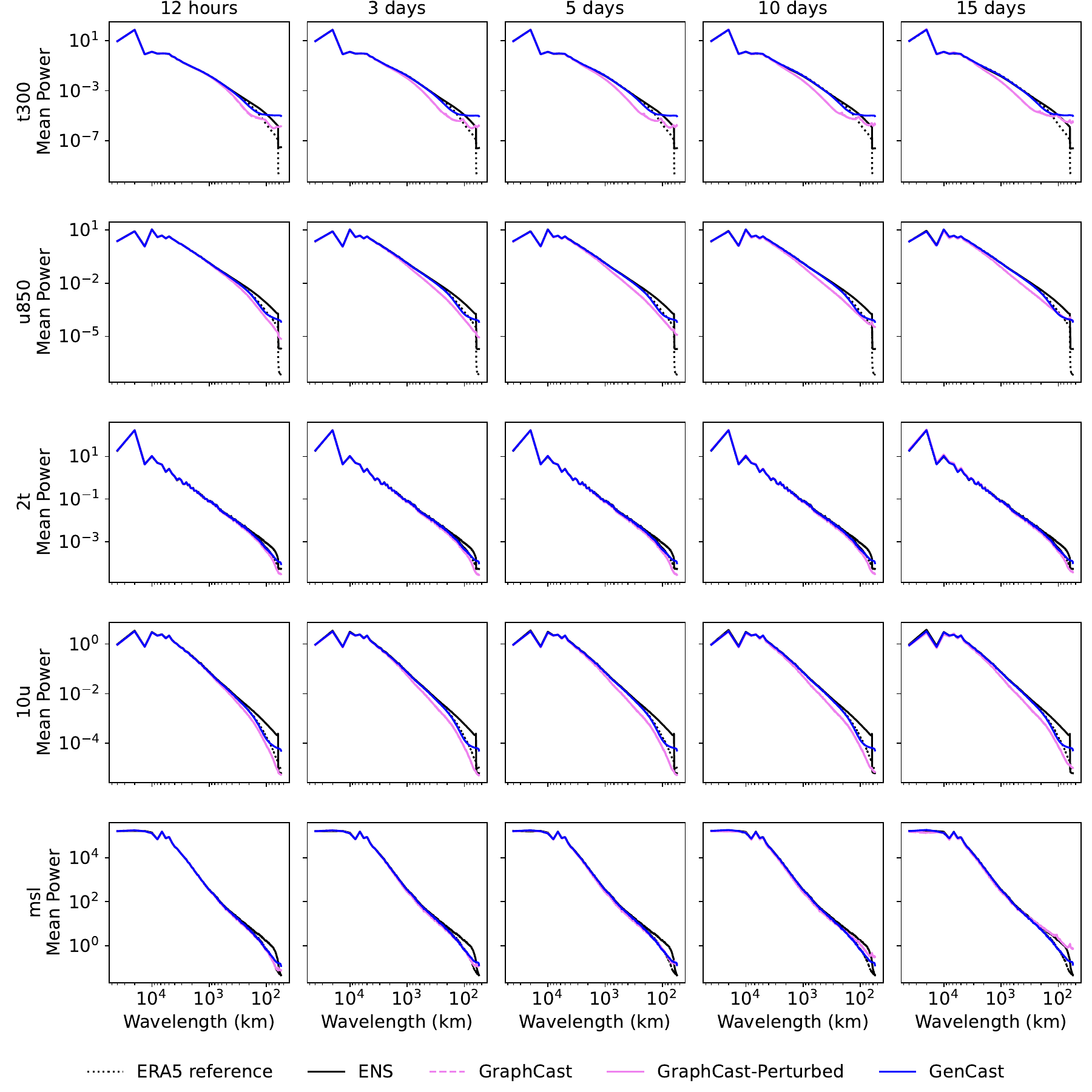}
    \caption{Power spectrum plots for t300, u850, 2t, 10u and msl.}
    \label{fig:headline_spectra_group_1}
\end{figure}

We provide additional spectral results to supplement those the main paper, showing 10 representative variables: z500, z850, q700, q925, t850, t300, u850, 2t, 10u and msl (\cref{fig:headline_spectra_group_0}, \cref{fig:headline_spectra_group_1}).

\newpage

\subsection{\gcens scorecards}\label{sec:app:gcens_scorecards}

\begin{figure}[H]
    \centering
    \includegraphics[width=\textwidth]{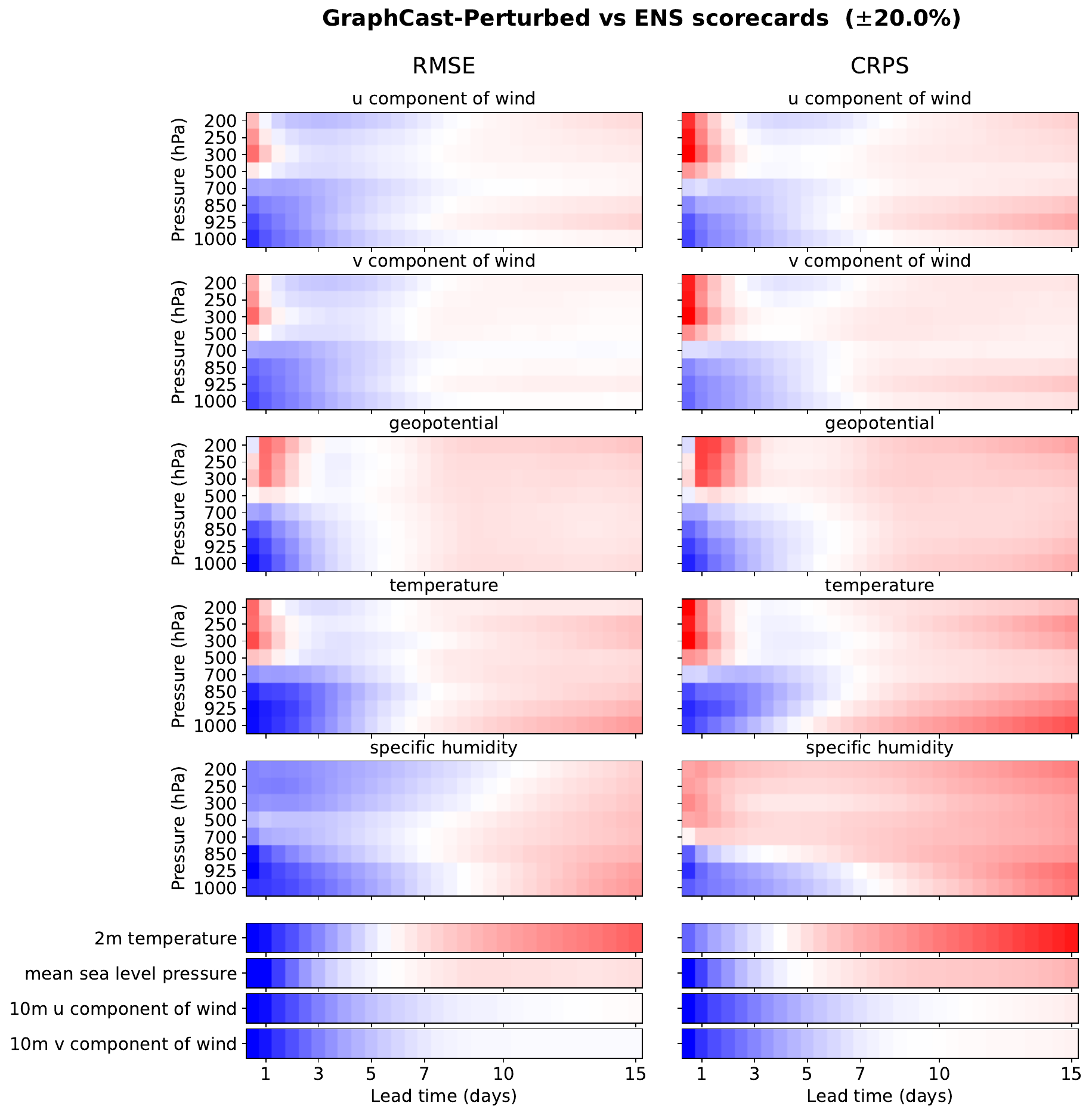}
    \caption{RMSE and CRPS scorecard comparing \gcens to ENS, dark blue (resp. red) means \gcens is 20\% better (resp. worse) than ENS, and white means they perform equally. Analogous to \cref{fig:skill} and \ref{fig:app:emrmse_gencast_vs_ens}. }
    \label{fig:app:gc_perturbed_vs_ens}
\end{figure}

\begin{figure}[H]
    \centering
    \includegraphics[width=\textwidth]{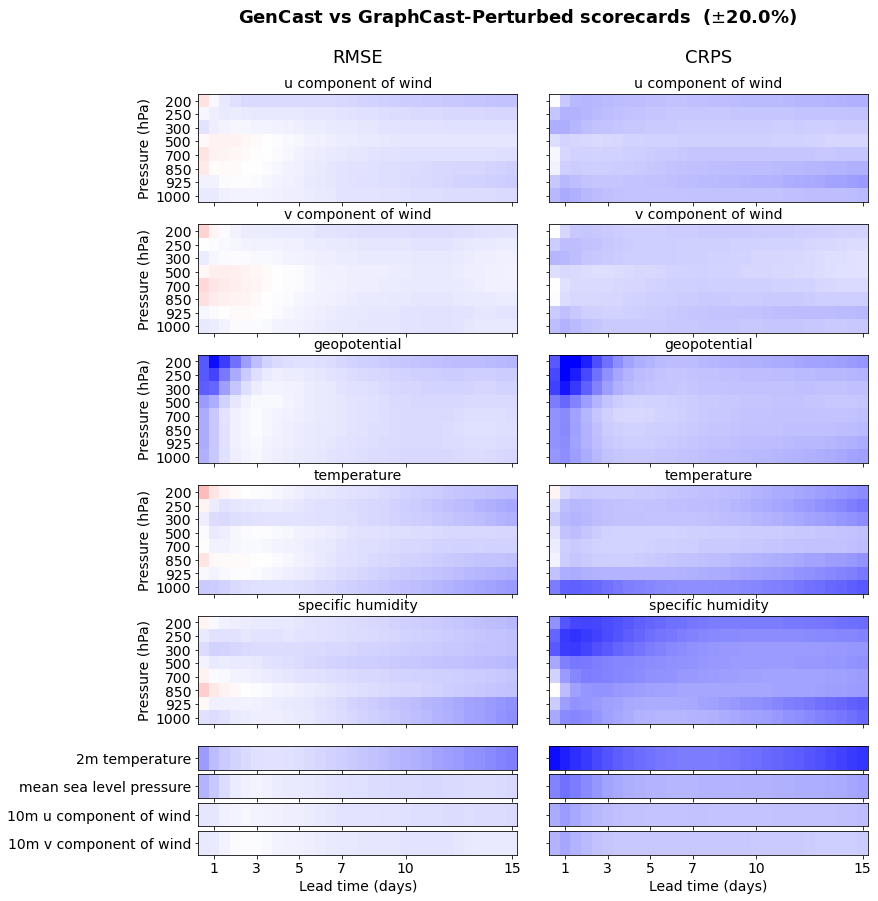}
    \caption{RMSE and CRPS scorecard comparing \ourmodel to \gcens. Analogous to \cref{fig:skill}, \ref{fig:app:emrmse_gencast_vs_ens} and \ref{fig:app:gc_perturbed_vs_ens}.
    }
    \label{fig:app:gencast_vs_gc_perturbed}
\end{figure}

\cref{fig:app:gc_perturbed_vs_ens} shows RMSE and CRPS scorecards comparing \gcens to ENS. We find that for long lead times, ENS is consistently better than \gcens, sometimes by as much as 20\%.  \cref{fig:app:gencast_vs_gc_perturbed} shows RMSE and CRPS scorecards comparing \ourmodel to \gcens. Except for RMSE at shorter lead times for a small number of variables, \ourmodel has better performance than \gcens.

\subsection{Extreme surface weather}\label{sec:app:extremes}

\begin{figure}[H]
  \centering
  \includegraphics[width=\textwidth]{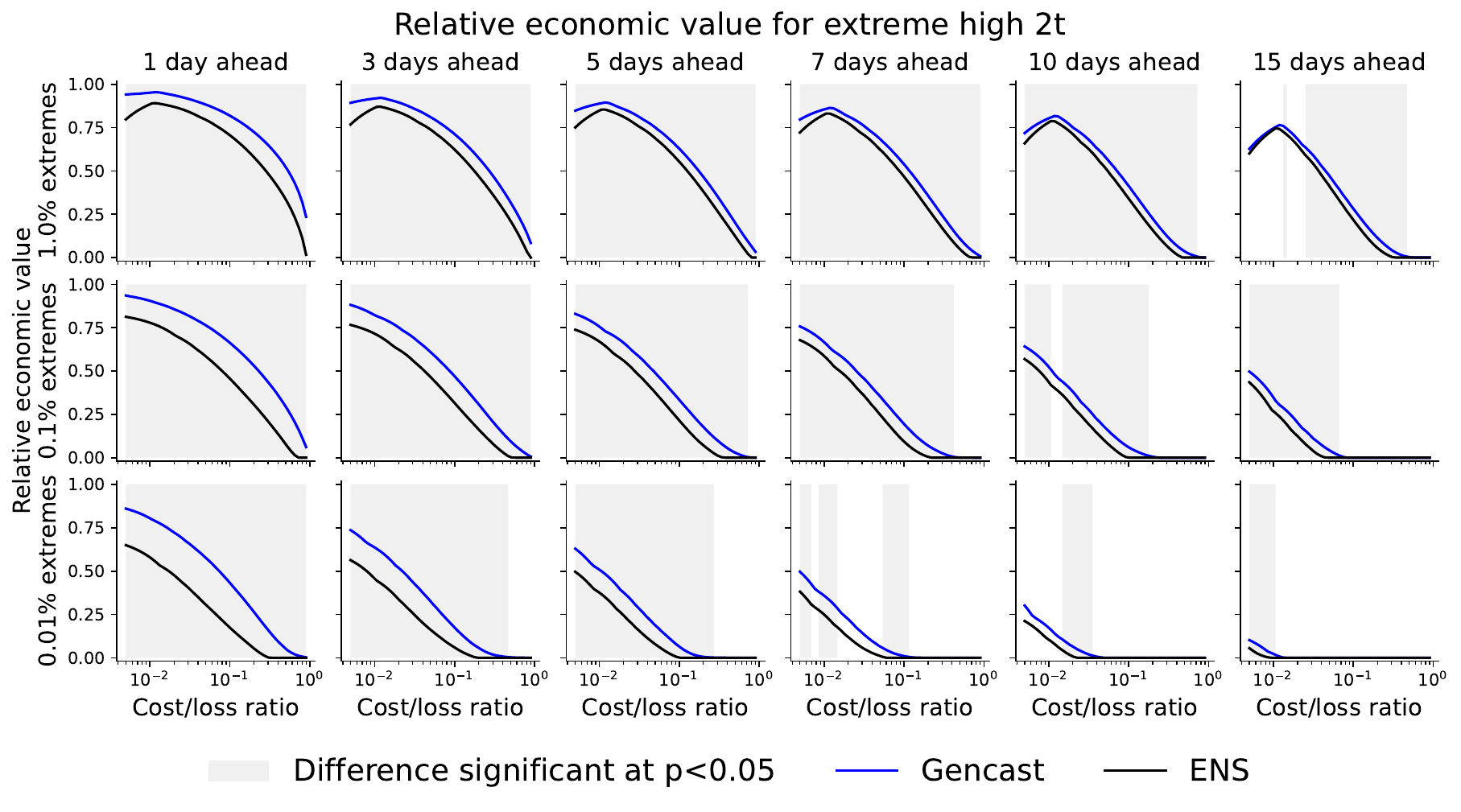}
  \caption{Relative economic value plots for extreme high temperatures. Regions for which \ourmodel is better then ENS with statistical significance are shaded in grey.}
  \label{fig:app:supplementary_rev_2t_high}
\end{figure}

\begin{figure}[H]
  \centering
  \includegraphics[width=\textwidth]{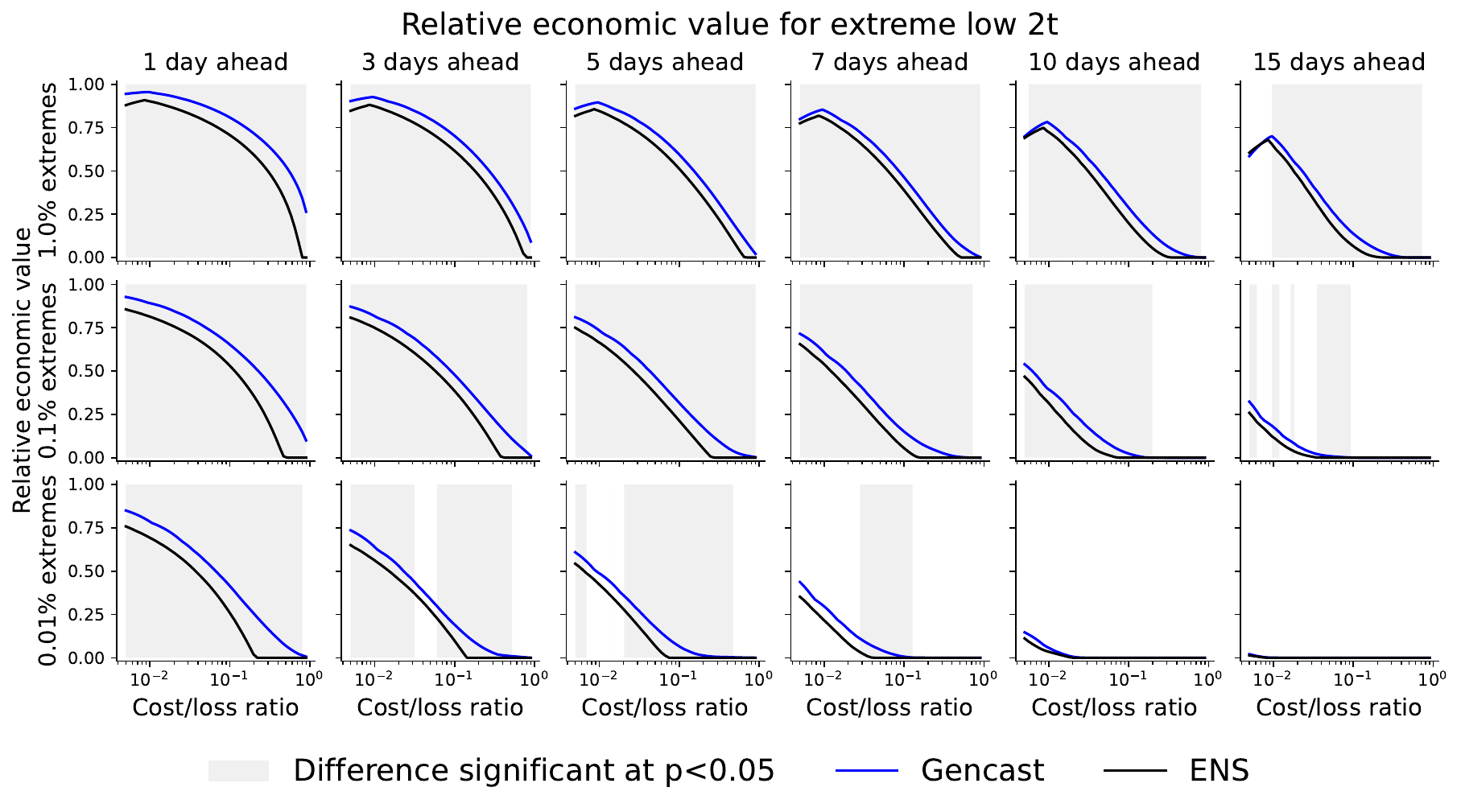}
  \caption{Relative economic value plots for extreme low temperatures. Regions for which \ourmodel is better then ENS with statistical significance are shaded in grey.}
  \label{fig:app:supplementary_rev_2t_low}
\end{figure}

\begin{figure}[H]
  \centering
  \includegraphics[width=\textwidth]{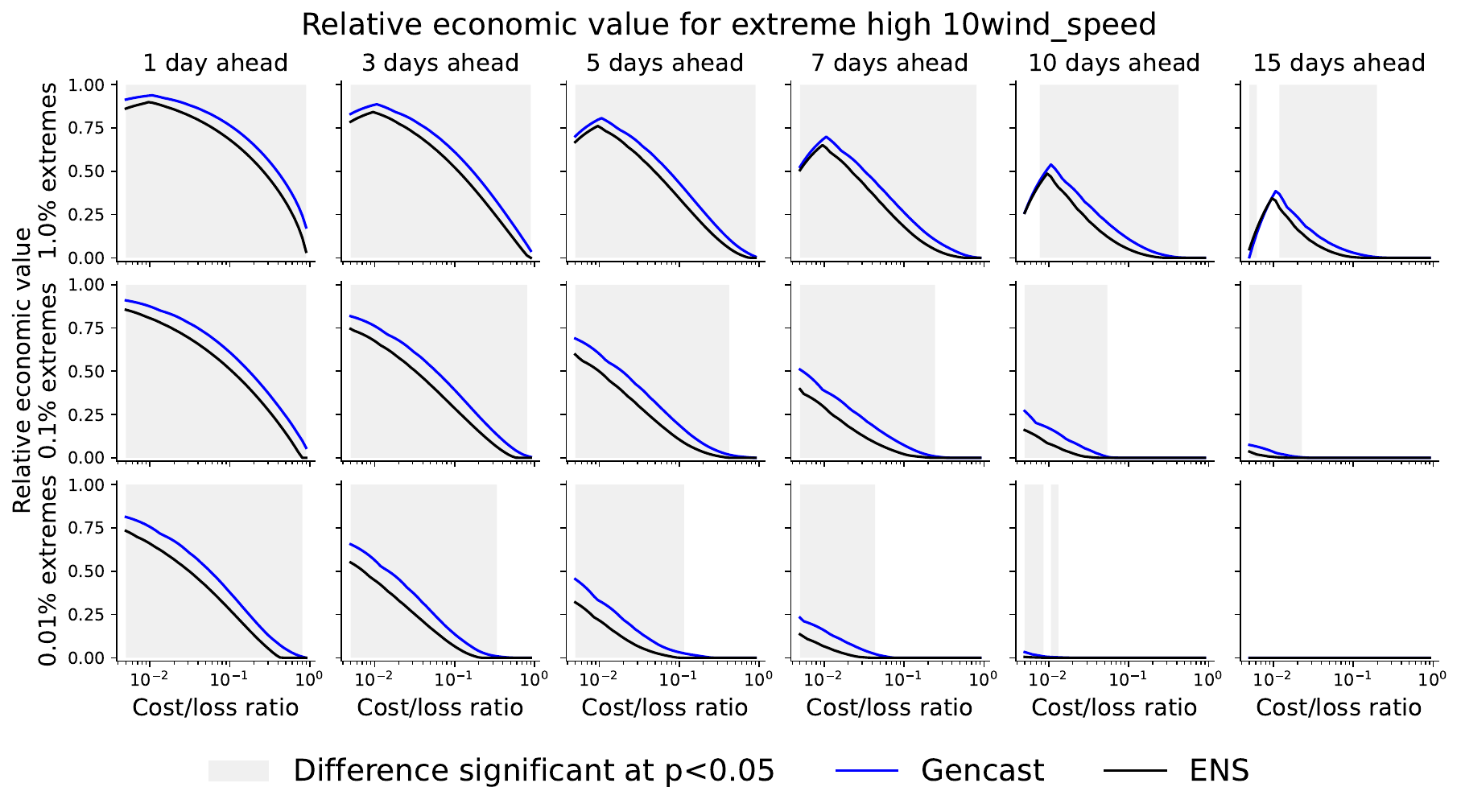}
  \caption{Relative economic value plots for extreme high wind speeds. Regions for which \ourmodel is better then ENS with statistical significance are shaded in grey.}
  \label{fig:app:supplementary_rev_wind_speed_high}
\end{figure}

\begin{figure}[H]
  \centering
  \includegraphics[width=\textwidth]{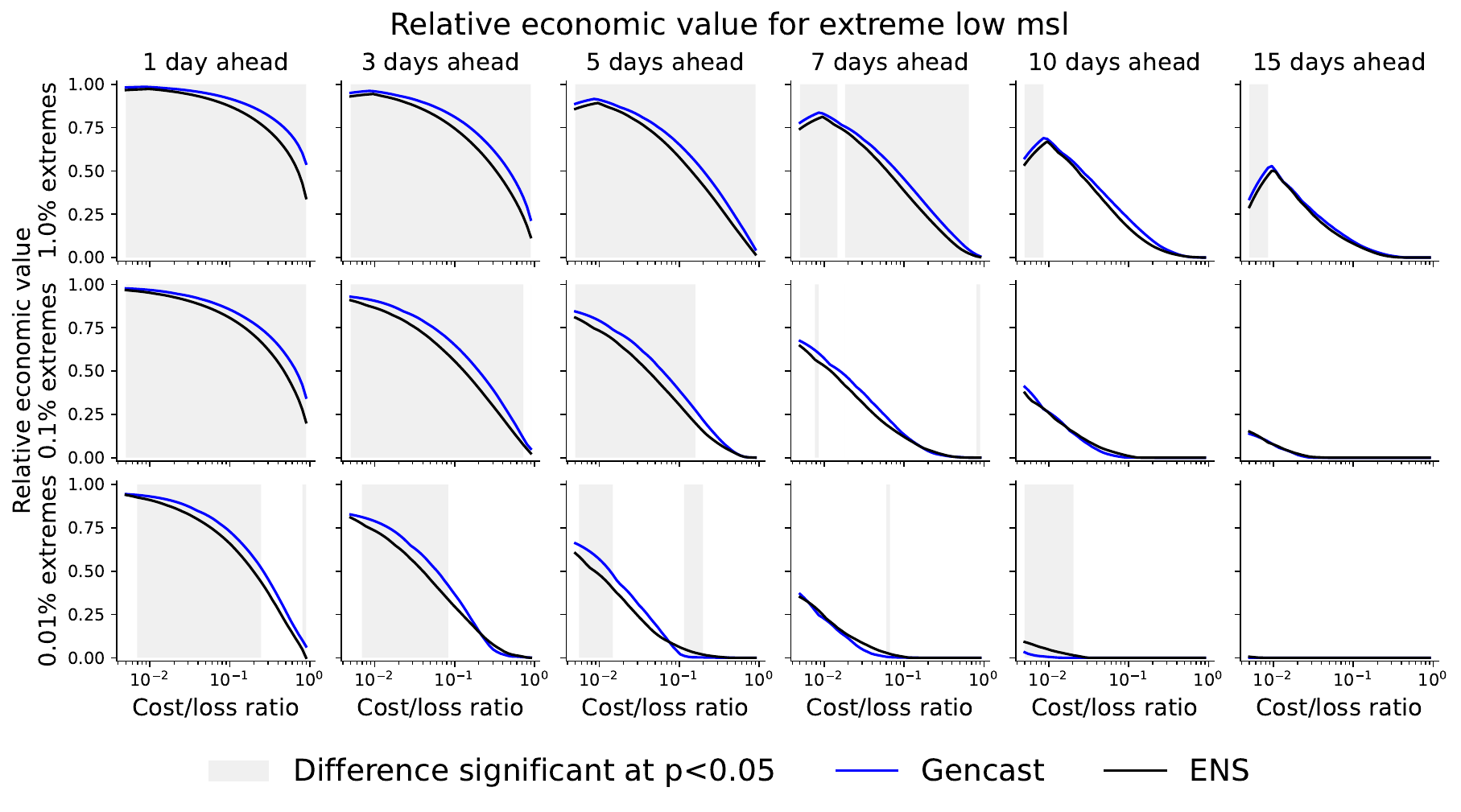}
  \caption{Relative economic value plots for extreme low mean sea level pressure. Regions for which \ourmodel is better then ENS with statistical significance are shaded in grey.}
  \label{fig:app:supplementary_rev_msl_low}
\end{figure}

We provide additional relative economic value plots, with statistical significance indicated, for various extreme thresholds, lead times and variables including extreme high temperature (\cref{fig:app:supplementary_rev_2t_high}), extreme low temperature (\cref{fig:app:supplementary_rev_2t_low}), extreme high wind speed (\cref{fig:app:supplementary_rev_wind_speed_high}), and extreme low mean sea level pressure (\cref{fig:app:supplementary_rev_msl_low}). \ourmodel is significantly better than ENS ($p < 0.05$) for up to 7 days lead time in many cases, and up to 15 days lead time in some others, while in some cases the differences are not statistically significant.

\begin{figure}
  \centering
  \includegraphics[width=\textwidth]{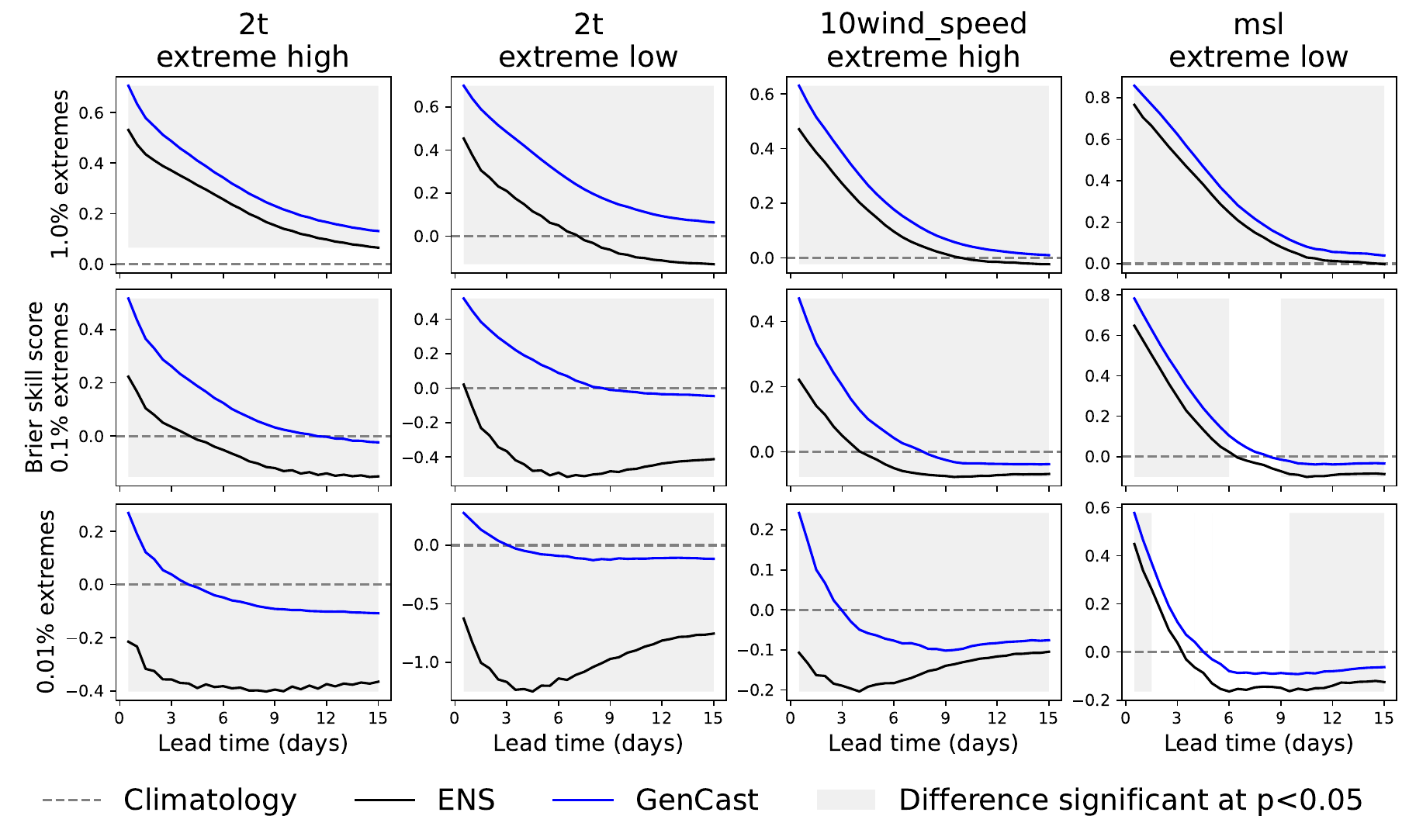}
  \caption{Brier skill scores for ENS and \ourmodel{} for extreme high temperatures, extreme low temperatures, extreme high wind speed, and extreme low mean sea level pressure, with each row computing the metric for different percentile threshold. A skill score of one represents a perfect score, a skill score of 0 represents the skill of climatology, and below zero represents a skill worse than climatology. Regions for which \ourmodel is better then ENS with statistical significance are shaded in grey.}
  \label{fig:app:supplementary_brier}
\end{figure}

We also provide comparisons on Brier skill scores, with statistical significance indicated (\cref{fig:app:supplementary_brier}). \ourmodel shows significant ($p < 0.05$) improvement over ENS for all thresholds, variables, and lead times shown, with the exception of certain lead times for <0.01 and <0.1 percentile mean sea level pressure where the improvement is not significant.

\FloatBarrier

\subsection{Tropical cyclones}\label{sec:app:cyclones_results}
\begin{figure}[H]
    \centering
    \includegraphics[width=\textwidth]{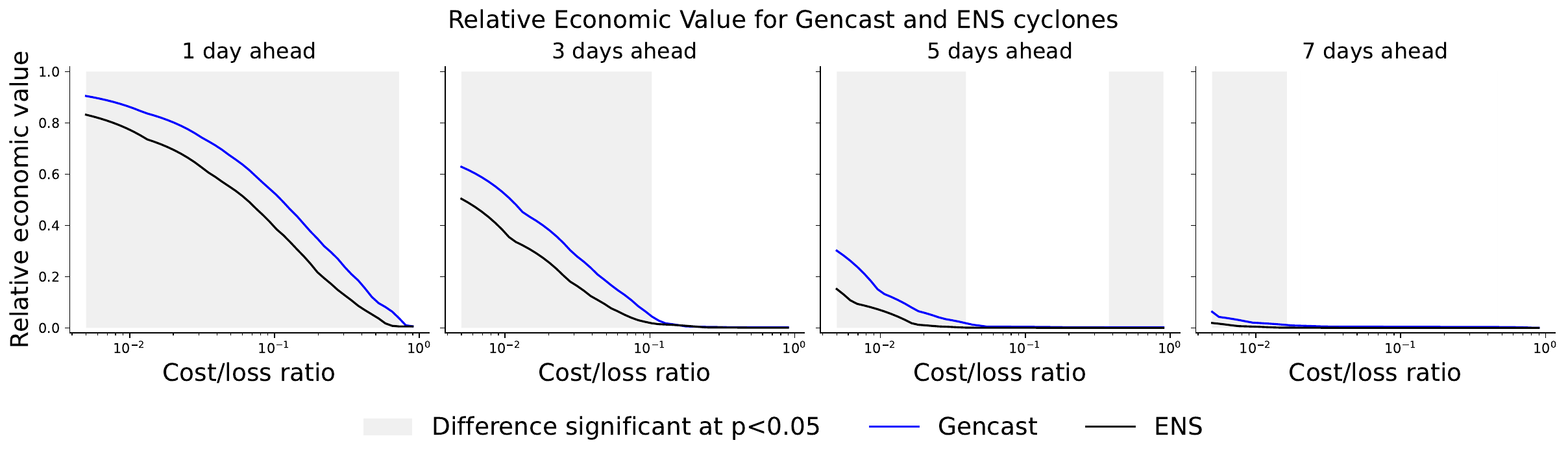}
    \caption{Relative economic value for cyclone track prediction when comparing \ourmodel and ENS to ERA5 and HRES-fc0 targets, respectively. Statistical significance $(p<0.05)$ of the improvement yielded by \ourmodel is indicated by light grey shading. The surprising statistical confidence at high cost/loss ratios at a 5 day lead time is due to GenCast having small but non-zero skill and ENS having exactly zero skill.}
    \label{fig:app:rev_cyclones_significance}
\end{figure}

This section provides supplementary results for the tropical cyclone evaluation, including statistical significance.
When tracks from \ourmodel and ENS are compared to cyclone tracks from their own ground truths (ERA5 and HRES-fc0, respectively), \cref{fig:app:rev_cyclones_significance} shows that \ourmodel significantly outperforms ENS across a broad range of cost-loss ratios at a lead time of 1 day, as well as at smaller cost-loss ratios at longer lead times up to 7 days.

\begin{figure}[H]
  \centering
  \includegraphics[width=\textwidth]{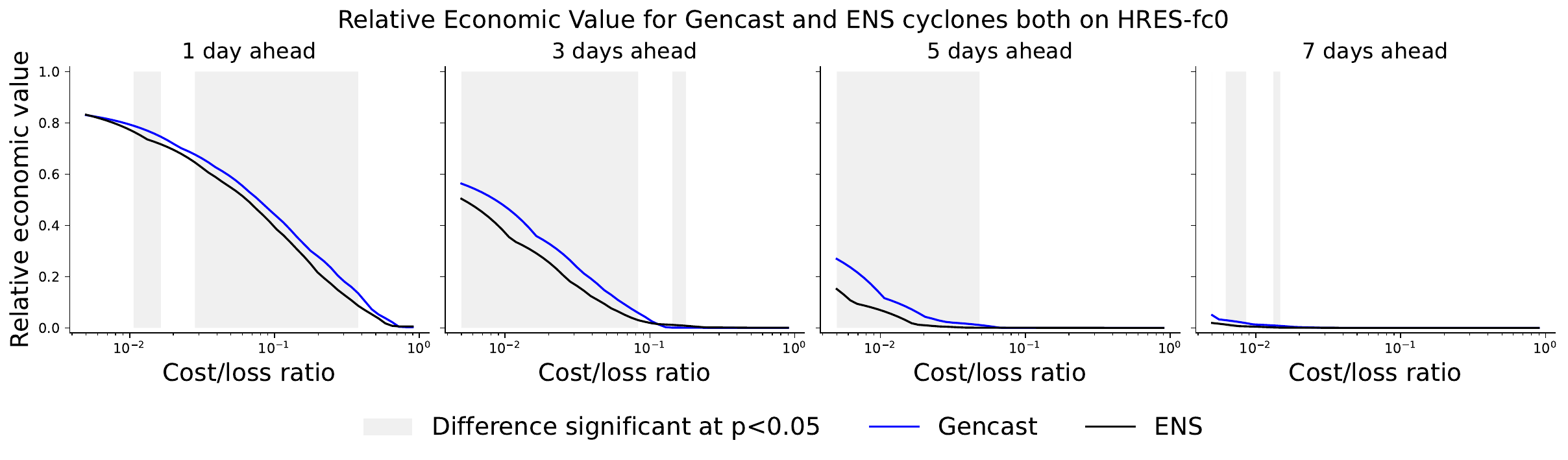} %
  \caption{Relative Economic Value for cyclone track prediction when comparing both \ourmodel and ENS to HRES-fc0 targets. Statistical significance $(p<0.05)$ of the improvement yielded by \ourmodel is indicated by light grey shading. In general, \ourmodel is still statistically significantly better than ENS despite the disadvantage of comparing against a different ground truth.}
  \label{fig:cyclones_revs_vs_fc0}
\end{figure}

For completeness, \cref{fig:cyclones_revs_vs_fc0} shows the REV obtained when both models are forced to use ENS's ground truth, HRES-fc0, giving an advantage to ENS.
The gap between ENS and \ourmodel is smaller, particularly at a 1 day lead time, because the tracker encounters a distribution shift when stitching the HRES-fc0 context window to \ourmodel's predictions (\cref{sec:app:cyclones:tempest_extremes}), which target ERA5.
There is also a second distribution shift penalty because perfectly predicting ERA5 ground truth cyclone tracks would result in non-zero error when evaluated against HRES-fc0.
Despite these two sources of penalty, \ourmodel significantly $(p<0.05)$ outperforms ENS at small cost-ratios and lead times of 3--5 days.
This shows that our results are robust to the difference in cyclone base rates between ERA5 and HRES-fc0.

\FloatBarrier

\subsection{Joint distribution}

\subsubsection{Wind speed as a derived variable}\label{sec:app:wind_speed_bias}

\gc and \gcens can exhibit significant biases when used to forecast variables that are derived non-linearly from those the model was trained to predict. This is because \gc is trained using mean-squared-error to predict an expectation, and a non-linear function of an expectation is not equal to the expectation of the non-linear function. For example, \gc and \gcens tend to predict wind speeds lower than average for long lead times, because \gc was trained to predict the expected wind vector $(u, v)$ and not wind speed $\sqrt{u^2 + v^2}$. Uncertainty about wind direction thus results in an expected wind vector closer to the origin, and a lower wind speed.

\begin{figure}
    \centering
    \includegraphics[width=\textwidth]{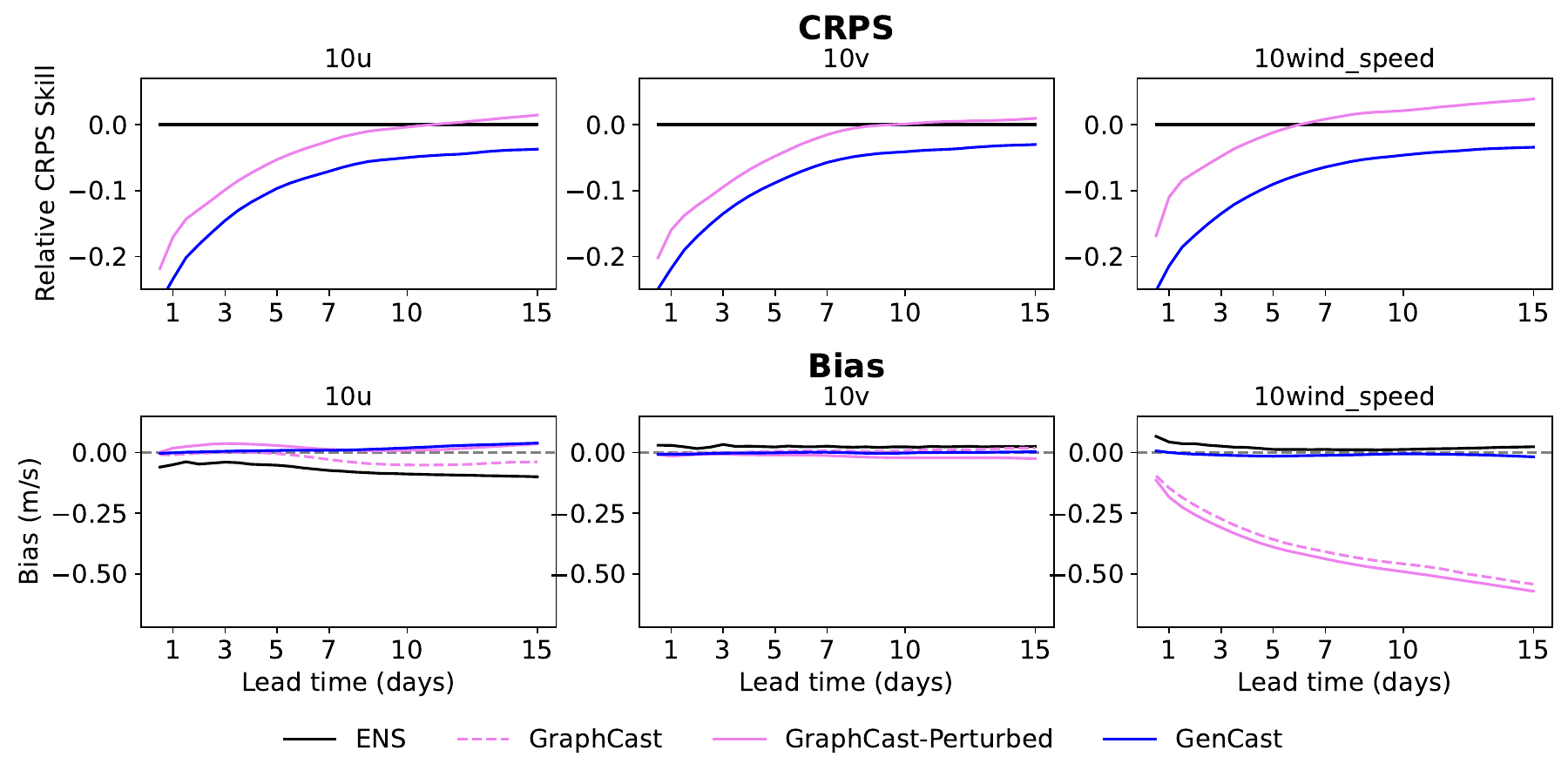}
    \caption{Comparison of the relative CRPS (lower is better) and bias (close to zero is better) performance on native surface wind components 10u and 10v, and derived wind speed. We observed how the performance of \gcens{} is relatively stronger on the variables that it was trained to predict (wind components), than it is on the derived variable (wind speed), exhibiting poorer relative CRPS and a strong negative bias for wind speed. Similarly, \gc exhibits minimal bias for 10u and 10v wind components, but a large negative bias on wind speed. On the other hand \ourmodel{} performs similarly well on both the wind components (which it was trained to predict), and the wind speed (which is derived). This seems to support the hypothesis that the states produced by \ourmodel{} are more physically plausible than \gc and \gcens{}, as are the states produced by ENS.}
    \label{fig:wind_bias}
\end{figure}

To derive a skillful predictive distribution for wind speed from $u$ and $v$ requires modelling dependence between $u$ and $v$, and so this task provides a simple test of \ourmodel's ability to model cross-variable dependence structure. In \cref{fig:wind_bias} we compare \ourmodel with \gcens (and ENS) on both CRPS and forecast bias for 10u, 10v, and \SI{10}{m} wind speed, where the wind speed variable is calculated as a function of the u and v forecasts. The most striking feature of \gcens and \gc in \cref{fig:wind_bias} is that while they exhibit minimal bias on 10u and 10v--- indeed, less than ENS---they exhibit a significant negative bias in their derived wind speed forecasts, in particular at longer lead-times. \ourmodel, on the other hand, exhibits minimal bias on the derived \SI{10}{m} wind speed, in addition to the very small bias it displays for 10u, 10v. Moreover, as shown in the top row of plots in \cref{fig:wind_bias}, while on CRPS the relative performance of \gcens against ENS is worse for \SI{10}{m} wind speed than for 10u and 10v, \ourmodel maintains the advantage it has over ENS for \SI{10}{m} wind speed as well.

\subsubsection{Pooled evaluation}\label{sec:app:pooled_eval}

\begin{figure}[H]
    \centering
    \includegraphics[width=\textwidth]{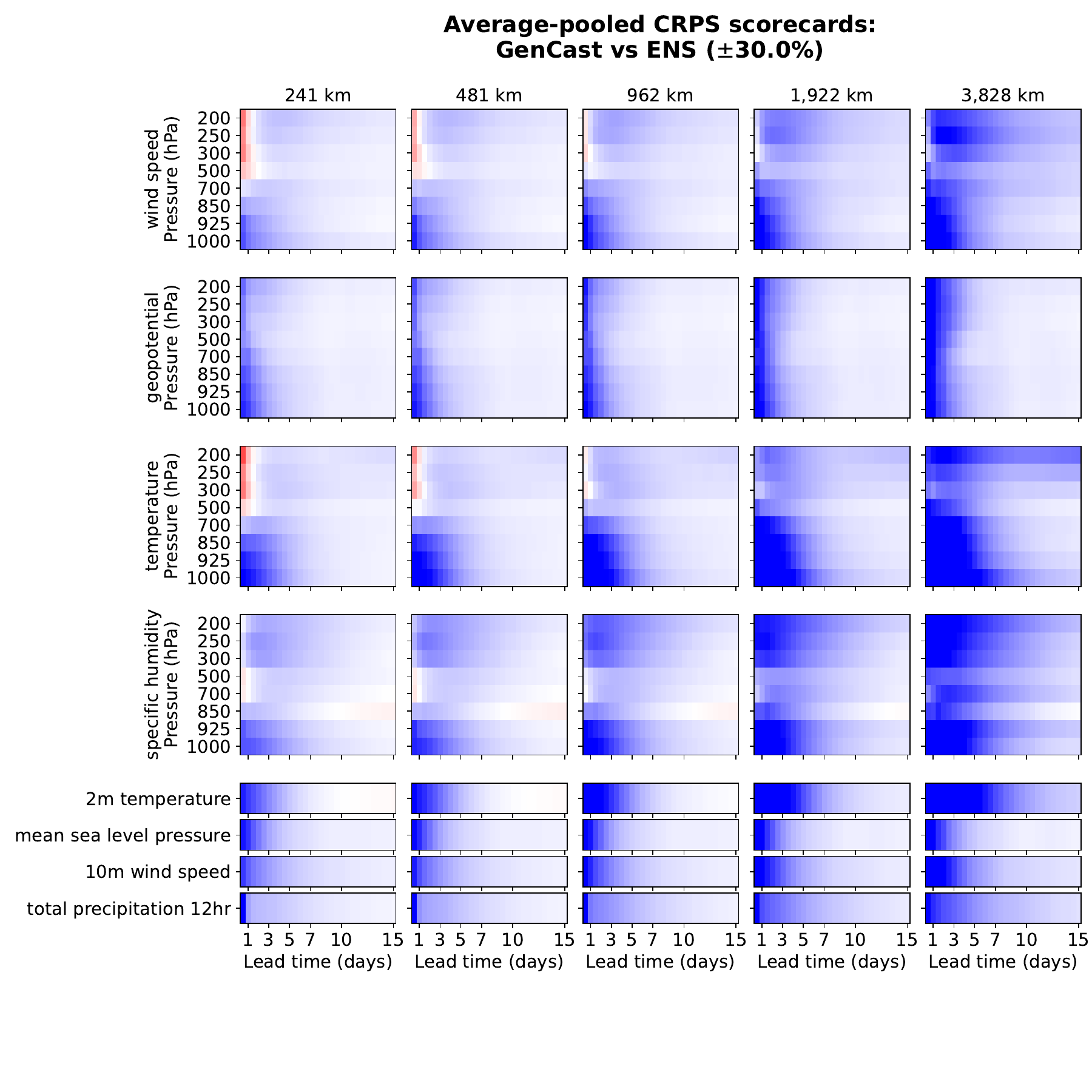}
    \caption{Average-pooled CRPS scorecard comparing \ourmodel and ENS at varying spatial scales, dark blue (resp. red) means \ourmodel is 30\% better (resp. worse) than ENS, and white means they perform equally.}
    \label{fig:pool_scorecard_gencast_average_low_res}
\end{figure}

\begin{figure}[H]
    \centering
    \includegraphics[width=\textwidth]{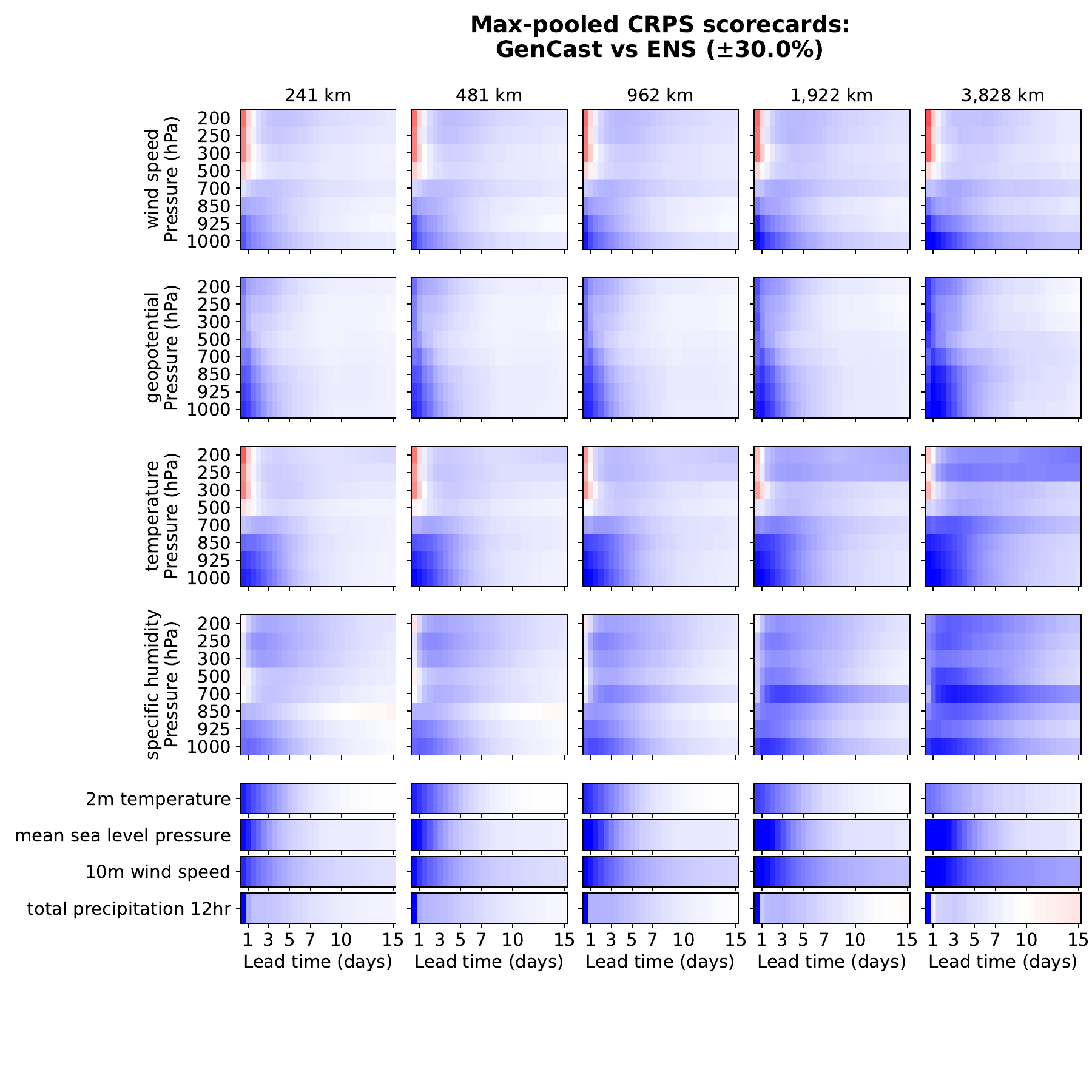}
    \caption{Max-pooled CRPS scorecard comparing \ourmodel and ENS at varying spatial scales, dark blue (resp. red) means \ourmodel is 30\% better (resp. worse) than ENS, and white means they perform equally.}
    \label{fig:pool_scorecard_gencast_max_low_res}
\end{figure}

\begin{figure}[H]
    \centering
    \includegraphics[width=\textwidth]{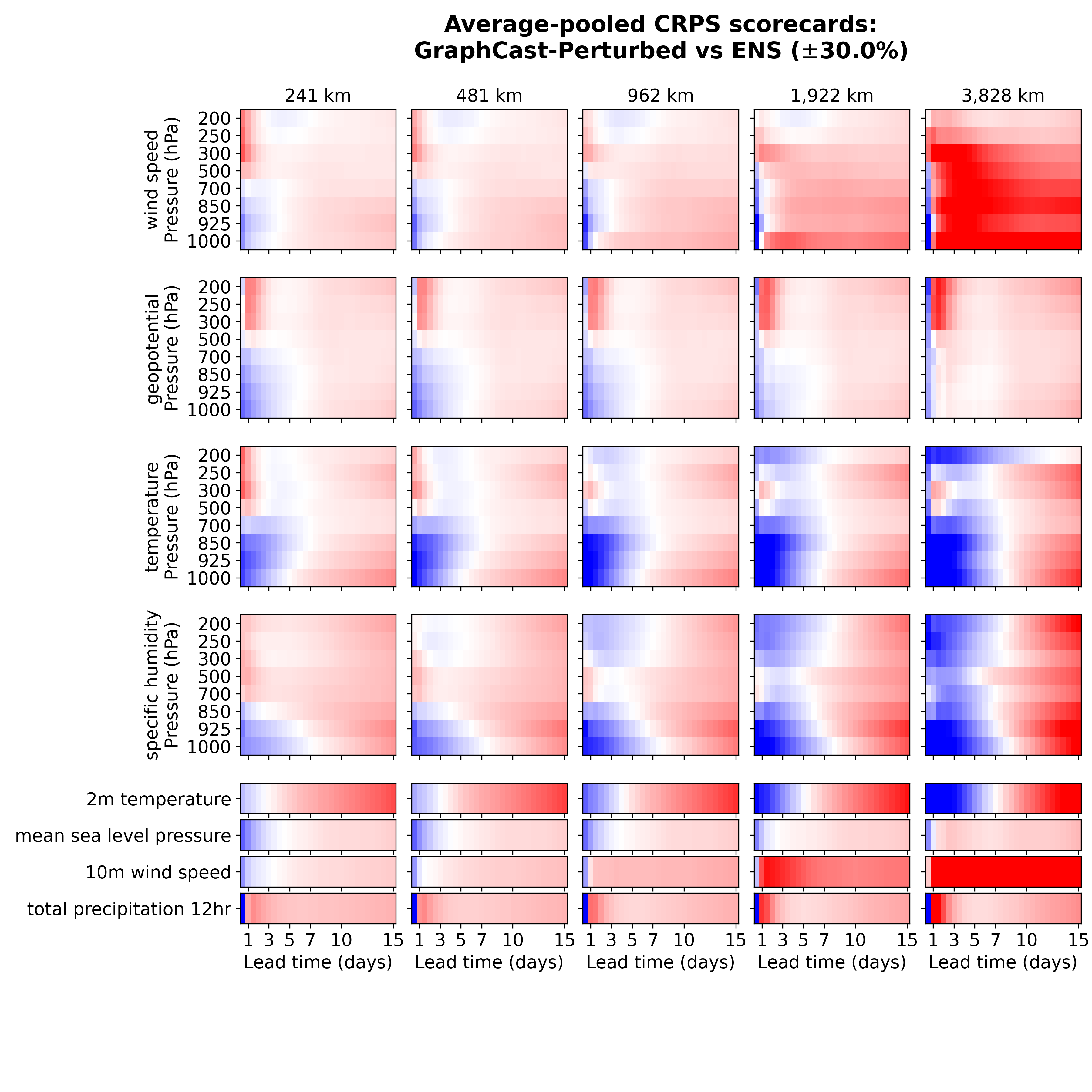}
    \caption{Average-pooled CRPS scorecard comparing \gcens and ENS at varying spatial scales, dark blue (resp. red) means \gcens is 30\% better (resp. worse) than ENS, and white means they perform equally.}
    \label{fig:pool_scorecard_graphcast_perturbed_average_low_res}
\end{figure}

\cref{fig:pool_scorecard_gencast_average_low_res} and \cref{fig:pool_scorecard_gencast_max_low_res} show averaged-pooled and max-pooled CRPS scorecards for \ourmodel relative to ENS.
\cref{fig:pool_scorecard_graphcast_perturbed_average_low_res} and \cref{fig:pool_scorecard_graphcast_perturbed_average_low_res} show averaged-pooled and max-pooled CRPS scorecards for \gcens relative to ENS.
We aggregate the u-component and v-component of wind into wind speed, and include tp12hr in our surface variables.
This results in 5400 pooled verification targets across all variables, lead times, and spatial scales.
Aggregating over all pooled verification targets, \ourmodel outperforms ENS's average-pooled CRPS in 98.1\% of cases (compared with 23.9\% for \gcens).
For max-pooled CRPS, \ourmodel outperforms ENS in 97.6\% of cases (compared with 7.0\% for \gcens, which performs especially badly on max-pooling due to blurring).

\begin{figure}[H]
    \centering
    \includegraphics[width=\textwidth]{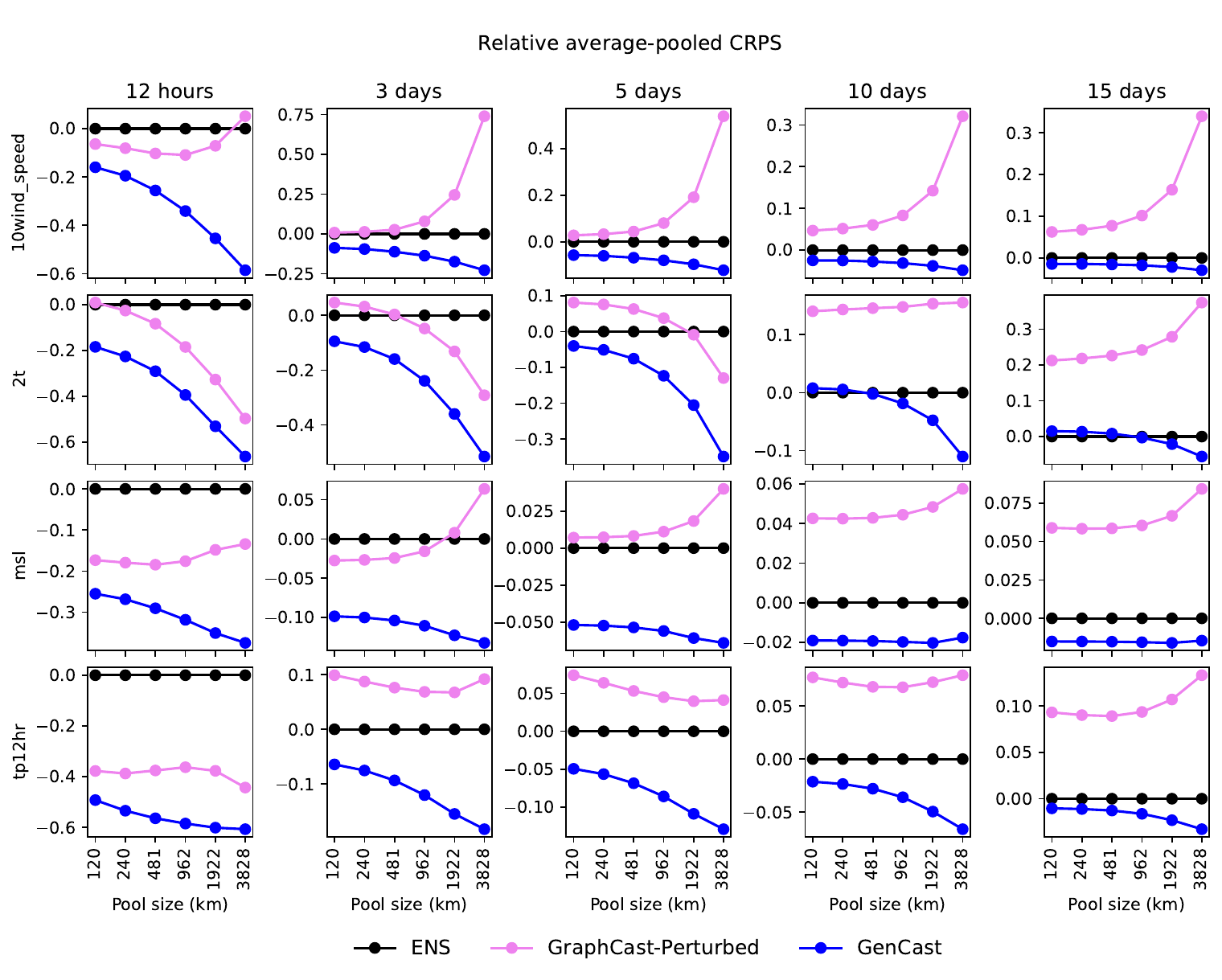}
    \caption{2019 average-pooled CRPS skill score plots for 0.25 deg surface variables. The sizes of the pooling regions can be visualized in \cref{fig:pool_sizes}.}
    \label{fig:pool_skillscore_average}
\end{figure}

\begin{figure}[H]
    \centering
    \includegraphics[width=\textwidth]{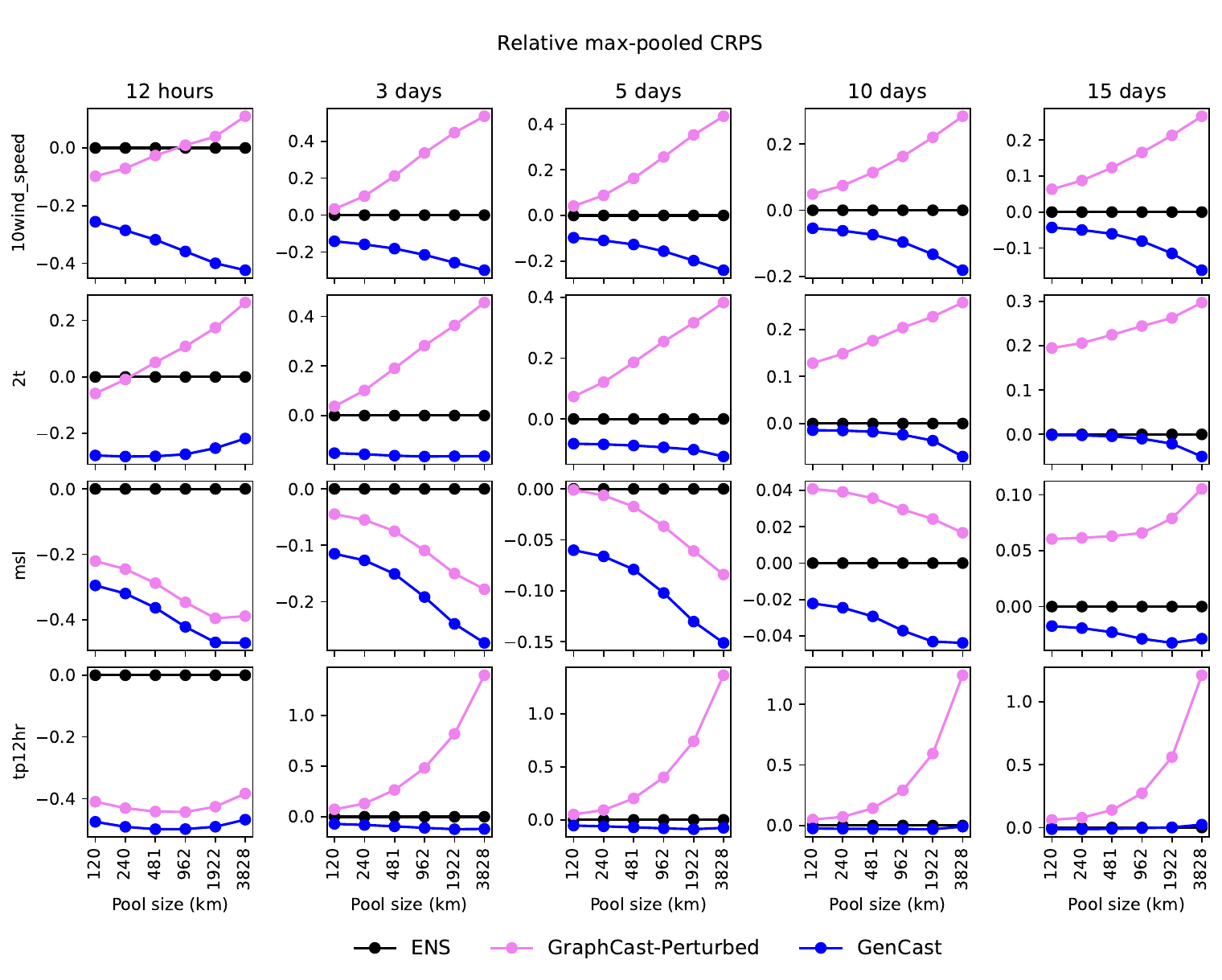}
    \caption{2019 max-pooled CRPS skill score plots for 0.25 deg surface variables. The sizes of the pooling regions can be visualized in \cref{fig:pool_sizes}.}
    \label{fig:pool_skillscore_max}
\end{figure}

These results hold up in our surface-only \ang{0.25}\xspace pooled evaluation, where \ourmodel substantially outperforms both ENS and \gcens's CRPS, often with relative performance improving as pooling size increases (\cref{fig:pool_skillscore_average} and \cref{fig:pool_skillscore_max}).

\subsection{Regional wind power forecasting statistical significance}

\begin{figure}[H]
    \centering
    \includegraphics[width=0.8\textwidth]{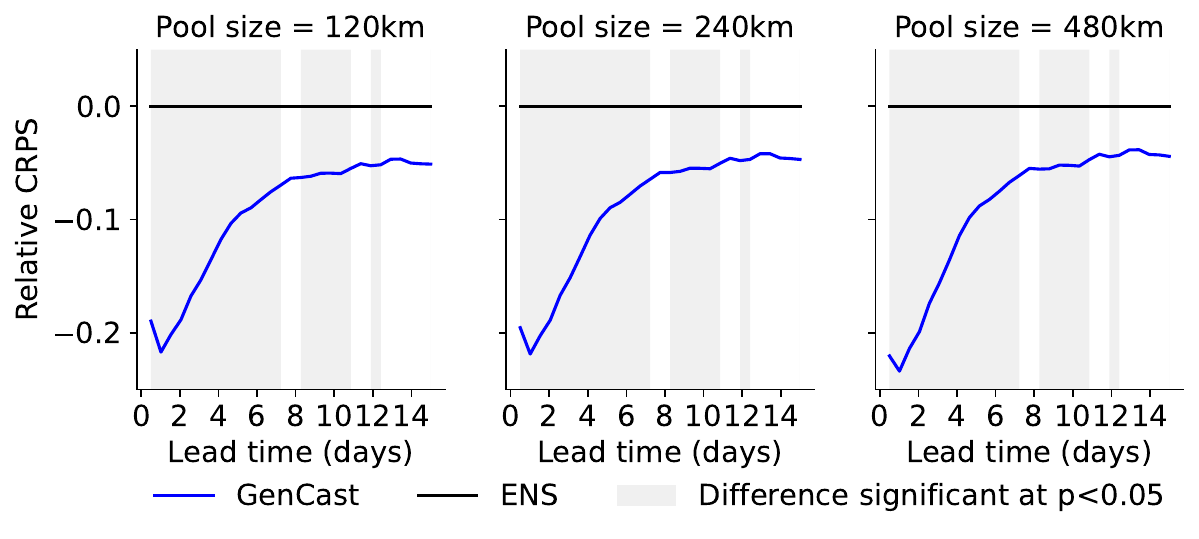}
    \caption{Regional wind power results. Grey shading indicates the lead times at which the improvement achieved by \ourmodel is statistically significant ($p < 0.05$).}
    \label{fig:app:wind_farms_significance}
\end{figure}

\cref{fig:app:wind_farms_significance} shows the relative CRPS of \ourmodel and ENS on the regional wind power forecasting task of \cref{sec:wind_farms}. Lead times at which the difference in CRPS is statistically significant ($p < 0.05$) are shaded grey.
Differences are significant up to a 7 day lead time for all pool sizes, and up to 10 days with a few exceptions.

\subsection{Ablation of EDA perturbations in \ourmodel initial conditions}\label{sec:app:eda_ablation}
\begin{figure}
    \centering
    \includegraphics[width=\textwidth]{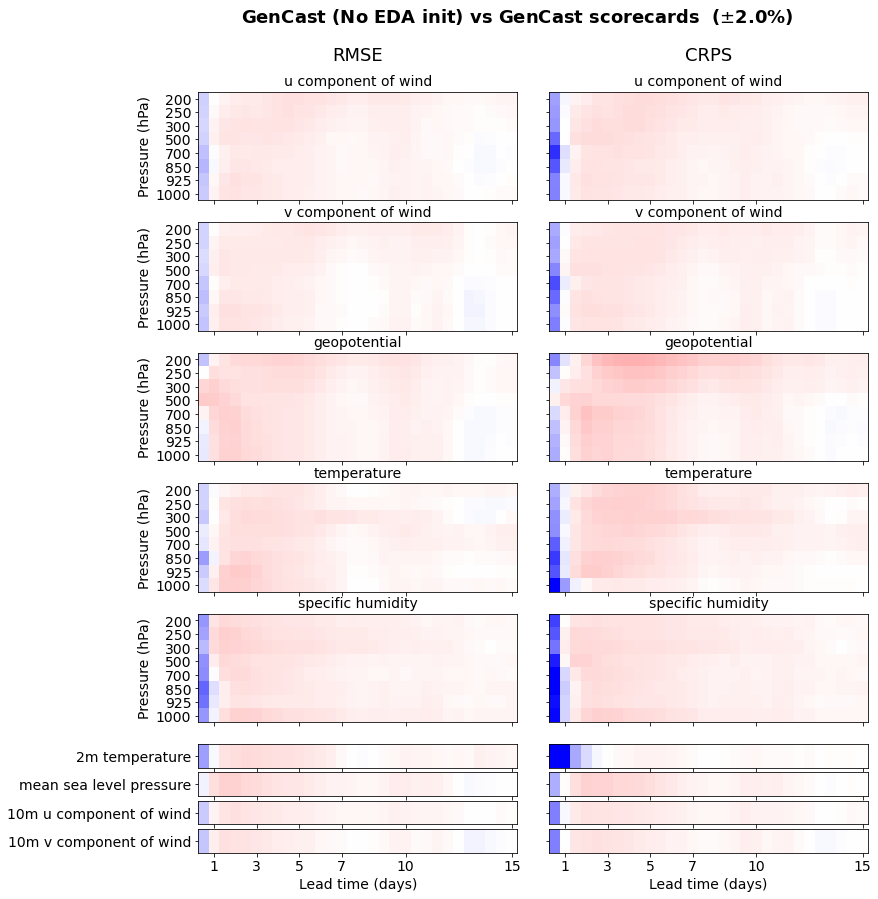}
    \caption{CRPS and RMSE scorecards comparing \ourmodel using deterministic ERA5 analysis as initial conditions to \ourmodel as presented in the main results section, using perturbed initial conditions. Performance with and without EDA-perturbed initial conditions is extremely similar, and in this scorecard maximum saturation of blue and red correspond to only a 2\% relative improvement or degradation respectively. Blue indicates \ourmodel initialised with deterministic ERA5 analysis is performing marginally better, Red indicates \ourmodel with EDA-perturbed initial conditions is performing marginally better.}
    \label{fig:no_eda_vs_with_eda_scorecards}
\end{figure}

\begin{figure}
    \centering
    \includegraphics[width=\textwidth]{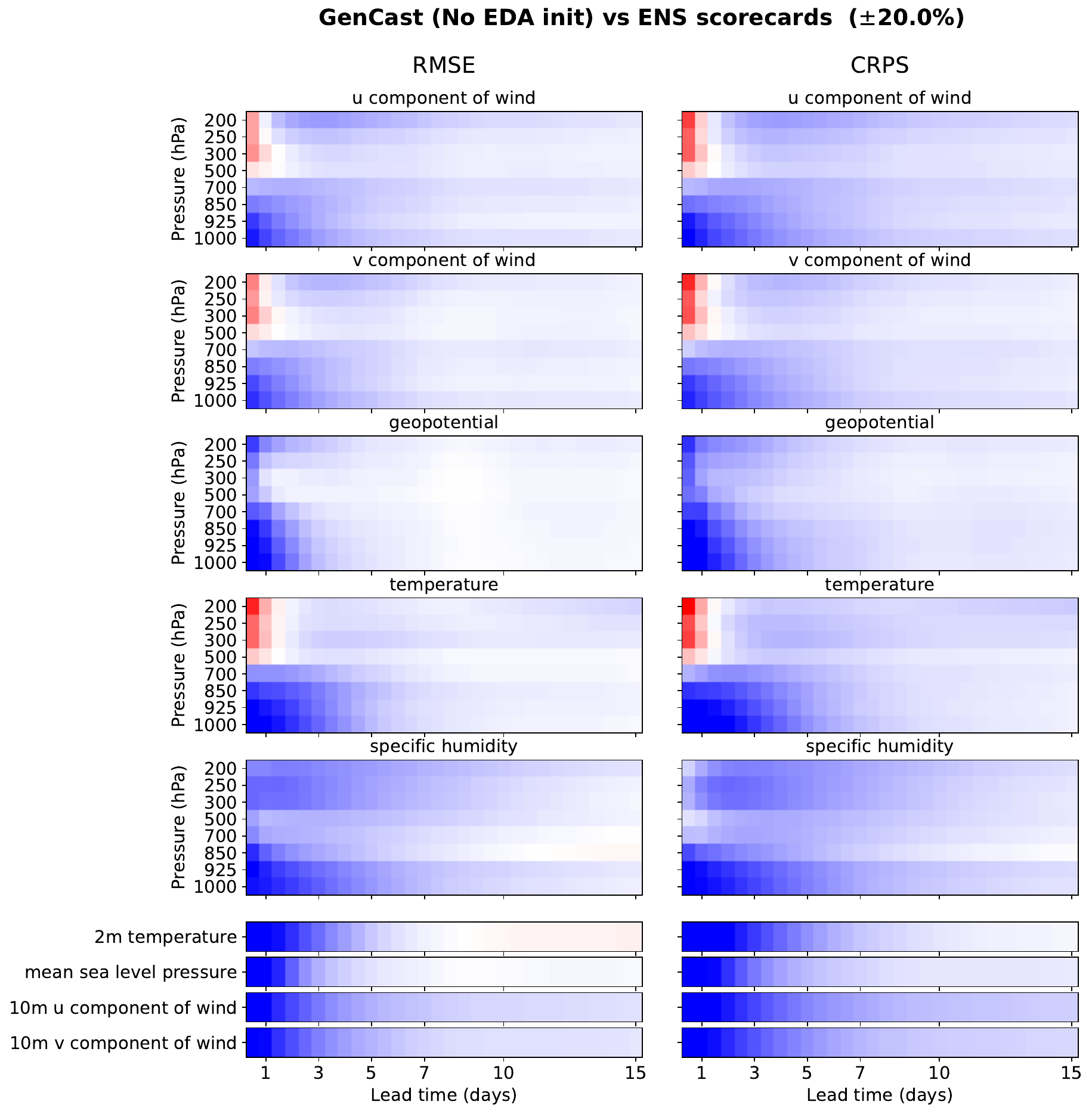}
    \caption{CRPS and RMSE scorecards comparing \ourmodel using deterministic ERA5 analysis as initial conditions to ENS. \ourmodel still outperforms ENS on 96\% and 97\% of ensemble-mean RMSE and CRPS targets respectively.}
    \label{fig:no_eda_vs_ens_scorecards}
\end{figure}

\begin{figure}
    \centering
    \includegraphics[width=\textwidth]{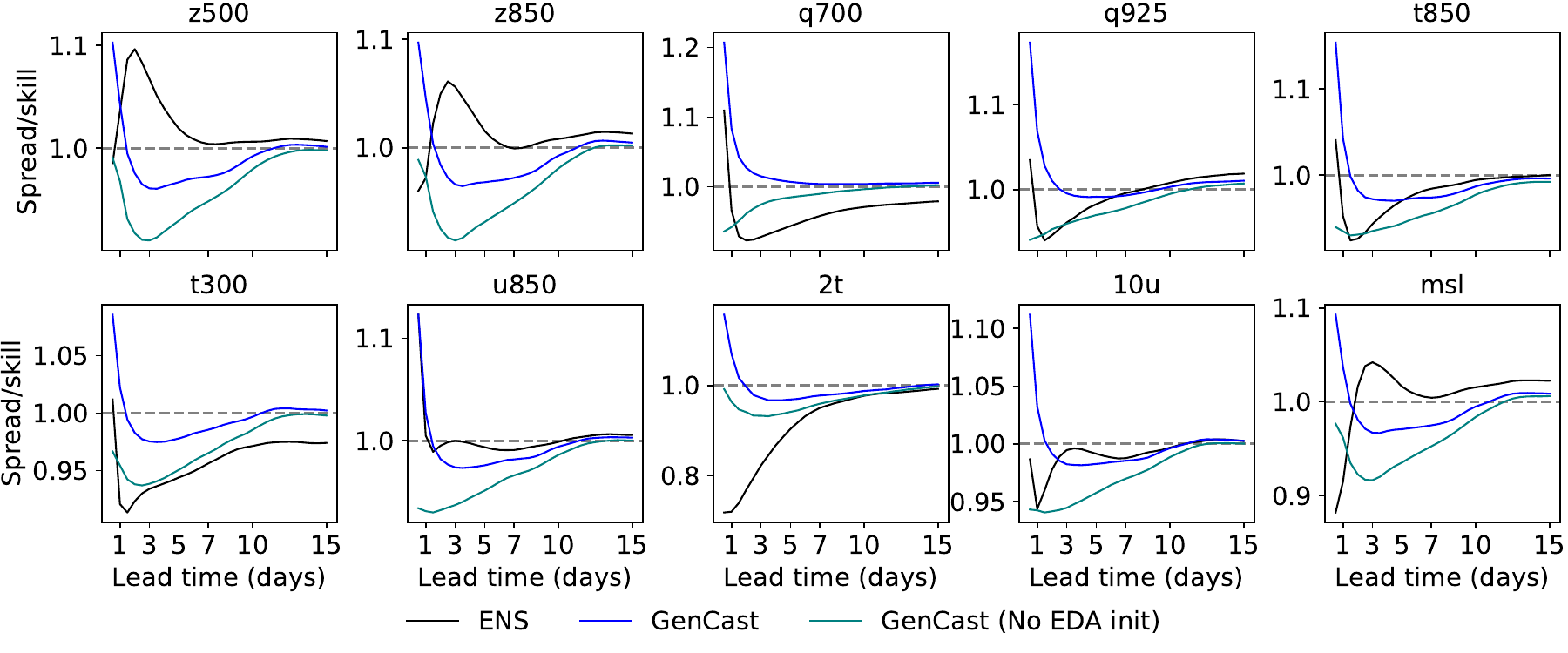}
    \caption{Spread-skill ratios of ENS, \ourmodel, and \ourmodel but without EDA-perturbed initial conditions. Initial spread-skill is closer to 1 for the first forecast step, but overall \ourmodel without EDA-perturbed initial conditions is somewhat under-dispersed. compared to the default \ourmodel.}
    \label{fig:eda_ablation_spread_skill}
\end{figure}

Here we ablate the use of perturbed initial conditions when generating ensemble forecasts with \ourmodel. \cref{fig:no_eda_vs_with_eda_scorecards} (colour range only $\pm 2$\%) shows that on CRPS and RMSE, EDA perturbations have very little effect, and \cref{fig:no_eda_vs_ens_scorecards} confirms that \ourmodel still outperforms ENS on RMSE and CRPS in the same percentage of cases even when initialising all ensemble members from deterministic ERA5 analysis. \cref{fig:eda_ablation_spread_skill} plots the spread skill ratios across lead times for a subset of variables and pressure levels, showing that the primary impact of the perturbed initial conditions is on the dispersion of the ensemble. Initialising with deterministic analysis alone generally improves the spread-skill ratio achieved on the first step or two. However we expect this is an artifact of evaluating against deterministic analysis, which will artificially reward under dispersion at short lead times, rather than being a meaningful improvement. In addition, the ensemble-initialised \ourmodel has generally better spread skill from approximately 2-3 days lead time onward.
 
\newpage

\subsection{Empirical support of initialisation time and evaluation time choices for verification}\label{sec:app:initialisation_choices}

\subsubsection{Initialisations used for \ourmodel evaluation}
\begin{figure}[ht!]
    \centering
    \includegraphics[width=\textwidth]{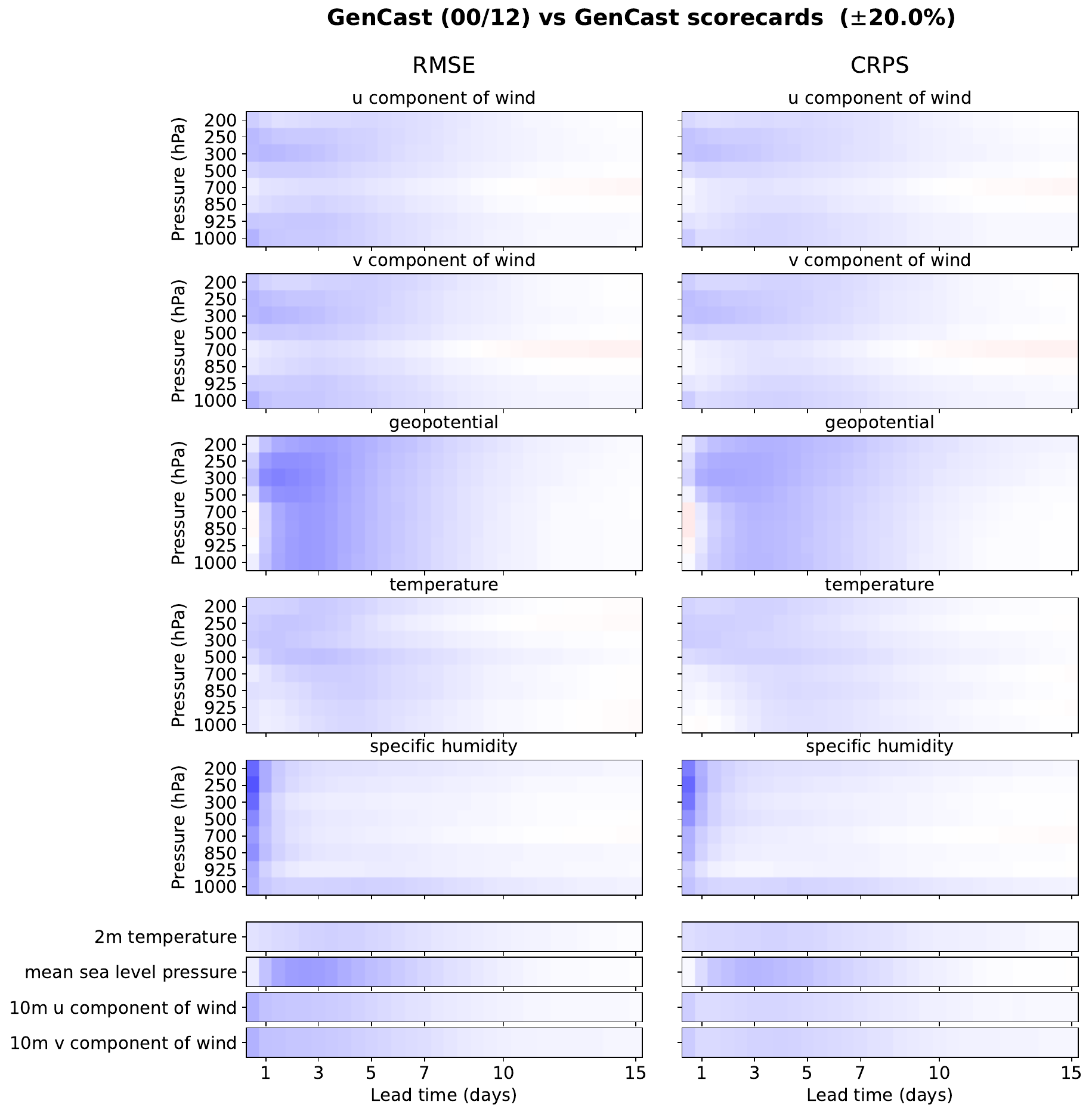}
    \caption{Scorecard comparing the performance of \ourmodel evaluated on 00/12 UTC initialisation times, to \ourmodel evaluated on the 06/18 UTC initialisation times (the latter is used in the rest of the evaluation in the paper). The 9 hour assimilation look-ahead of the 00/12 UTC initialisations gives the model a consistent advantage in CRPS and Ensemble-Mean RMSE over the 3 hour assimilation look-ahead of the 06/18 UTC initialisations. This highlights the importance of accounting for assimilation window look-ahead when comparing metrics across model evaluations.}
    \label{fig:scorecard_00_12_init}
\end{figure}

Our main results always use 06/18 UTC initialisation times when evaluating \ourmodel. \cref{fig:scorecard_00_12_init} shows a scorecard comparing \ourmodel initialised at 00/12 and 06/18 UTC, showing how the 00/12-initialised forecasts have a systematic advantage due to the ERA5 look-ahead at those initialisation times. This motivates the use of 06/18 UTC initialisations for the evaluations of \ourmodel as the most conservative approach.

\subsubsection{Lead time interpolation in regional wind farm evaluation}\label{sec:app:lead_time_averaging}

\begin{figure}
    \centering
    \includegraphics[width=\textwidth]{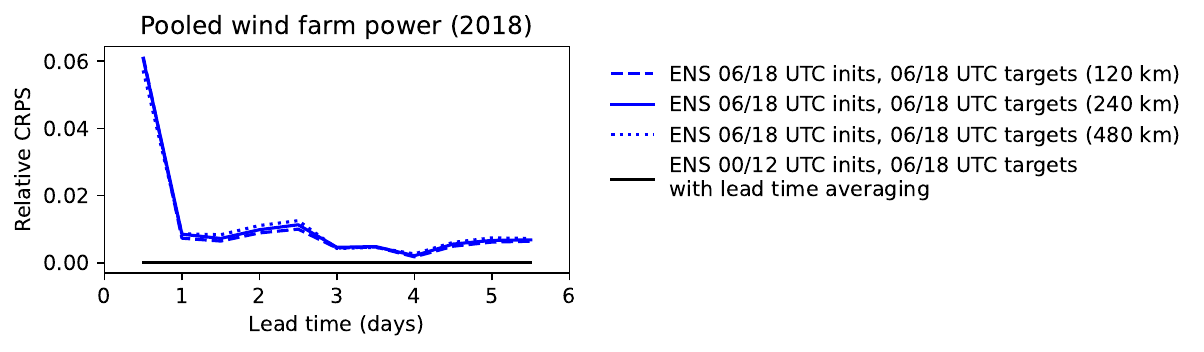}
    \caption{Relative CRPS of ENS (06/18 UTC initialisations)  compared to ENS 00/12 UTC initialisations with the lead time averaging applied to 06/18 UTC targets, on the regional wind power forecasting task.}
    \label{fig:ens_0618_sfc}
\end{figure}

\cref{sec:app:ens_init_times} motivates and describes the lead time interpolation method for evaluating ENS and \ourmodel on the same set of validity times for regional wind power forecasting. We were able to download 06/18 UTC initialised ENS forecasts for some surface variables for 2018 to validate and justify this approach. \cref{fig:ens_0618_sfc} shows that on the regional wind power forecasting task, lead time interpolation with 00/12-initialised ENS forecasts actually overestimates the performance of 06/18-initialised ENS forecasts on 06/18 UTC targets. This is particularly the case at a 12-hour lead time, where 06/18-initialised ENS performs 6\% worse than 00/12-initialised ENS with lead time interpolation. This suggests that the lead time interpolation is in fact advantaging ENS in \cref{sec:wind_farms}, and explains the non-monotonicity in \ourmodel's relative performance at the shortest lead times in \cref{fig:panel_applications}d.

\FloatBarrier

\clearpage

\section{Forecast visualisations}\label{sec:app:visualizations}

\subsection{Tropical cyclone tracks}\label{sec:app:cyclone_tracks}

\begin{figure}[!ht]
    \centering
    \hspace*{-0.8cm}
    \includegraphics[width=\textwidth]{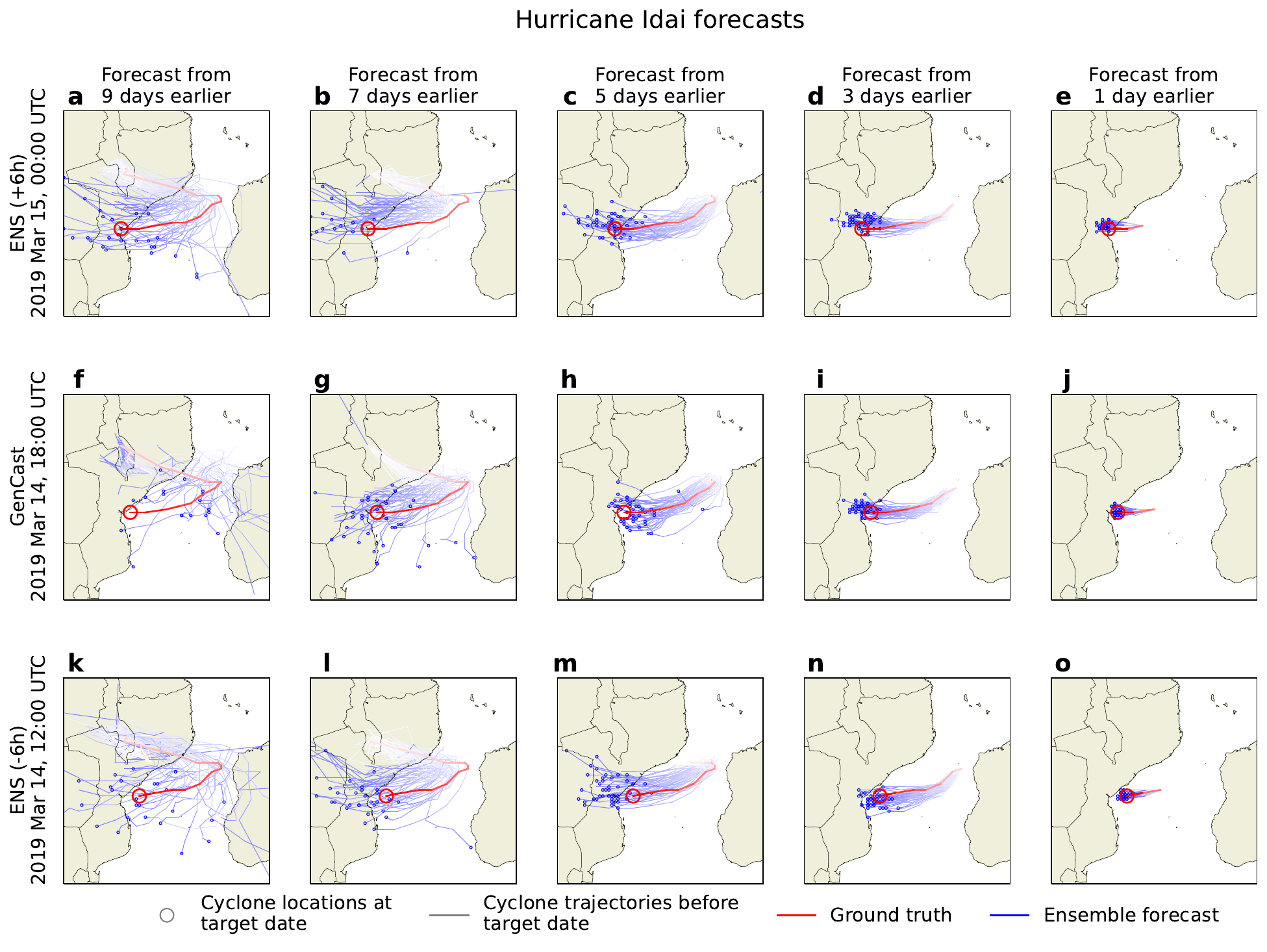}
    \caption{Visualisation of Cyclone Idai trajectory forecasts.}
    \label{fig:app:visualization_cyclone_idai}
\end{figure}

\begin{figure}[!ht]
    \centering
    \hspace*{-0.8cm}
    \includegraphics[width=\textwidth]{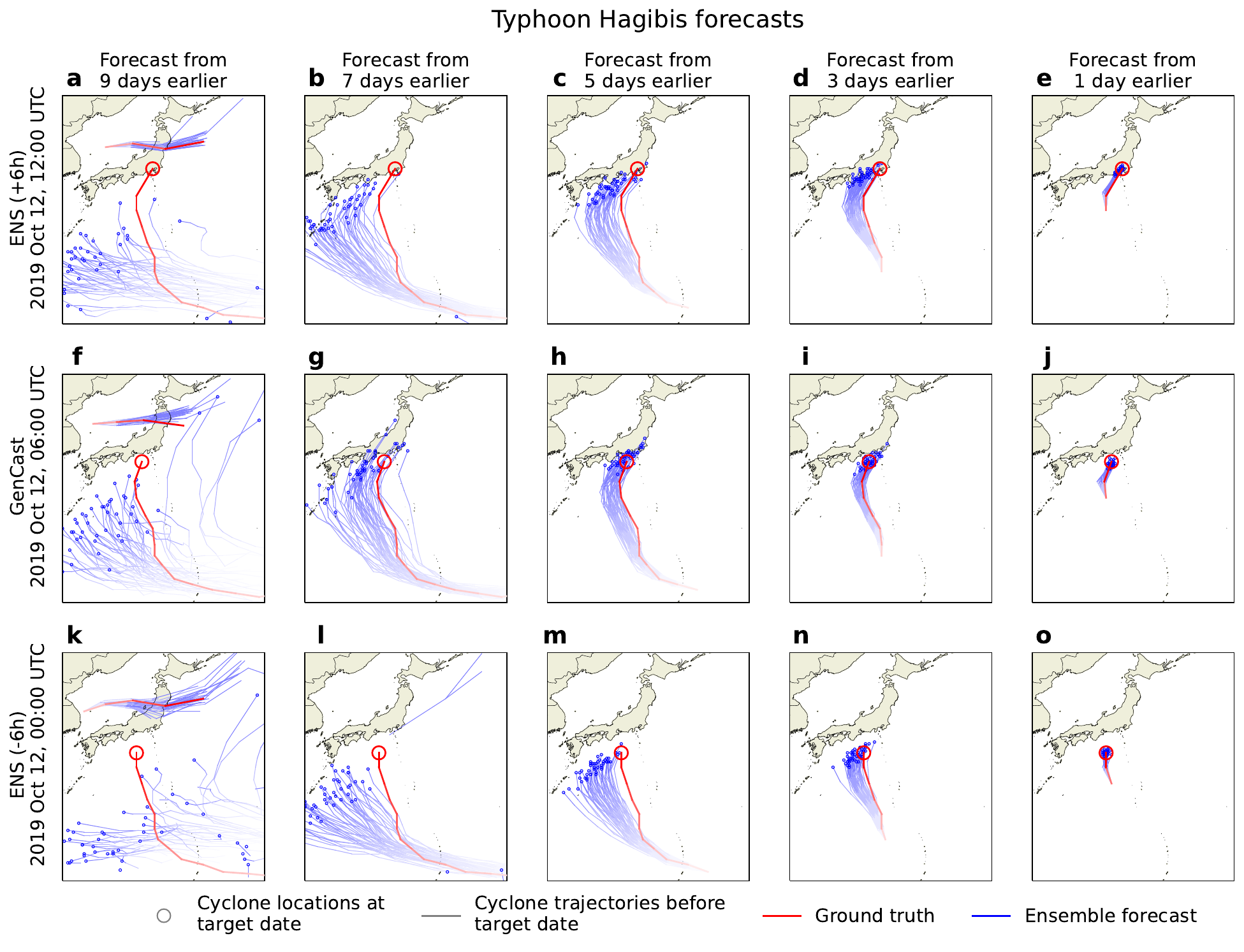}
    \caption{Visualisation of Typhoon Hagibis trajectory forecasts.}
    \label{fig:app:visualization_cyclone_hagibis}
\end{figure}

\begin{figure}[!ht]
    \centering
    \hspace*{-0.8cm}
    \includegraphics[width=\textwidth]{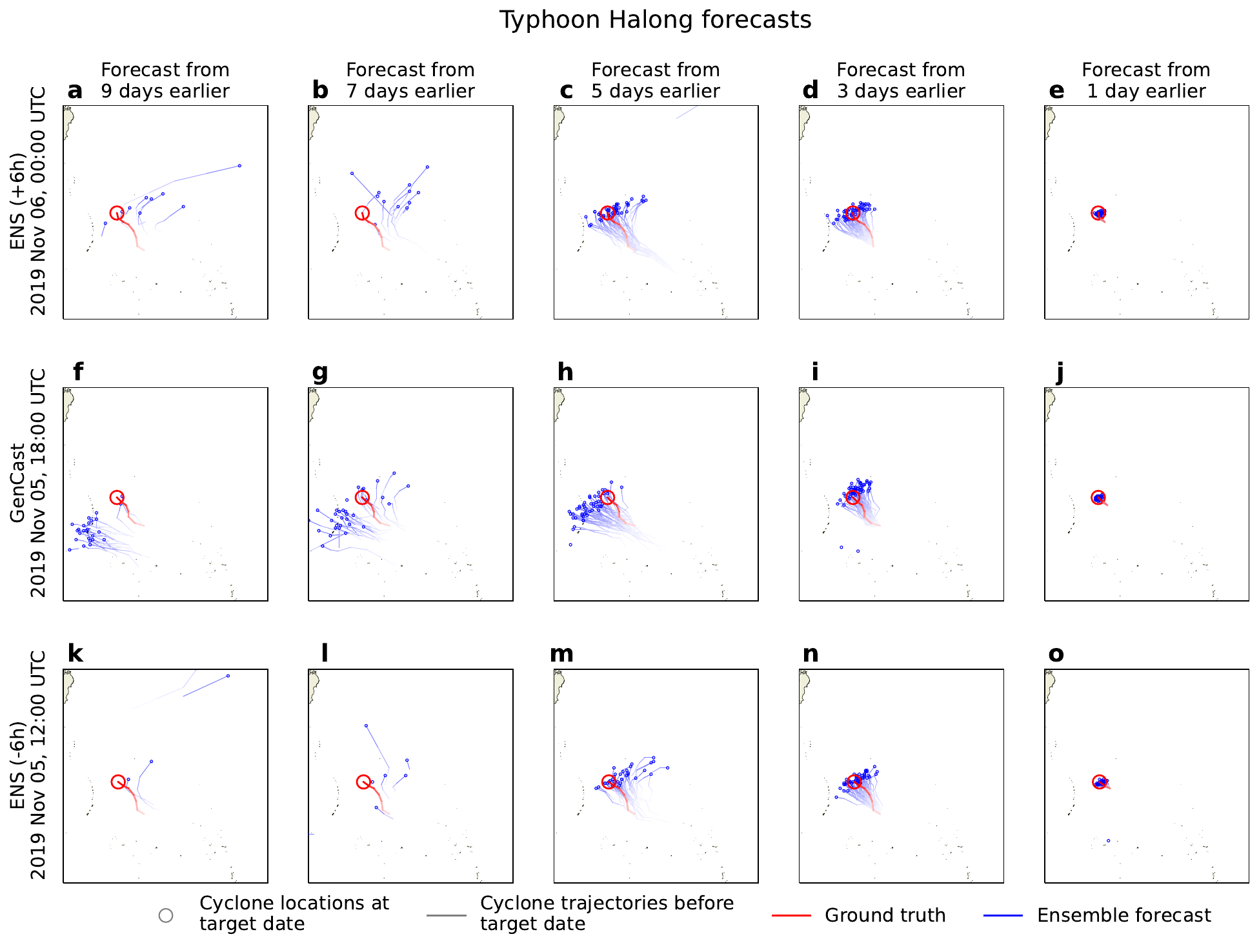}
    \caption{Visualisation of Typhoon Halong trajectory forecasts.}
    \label{fig:app:visualization_cyclone_halong}
\end{figure}

\begin{figure}[!ht]
    \centering
    \hspace*{-0.8cm}
    \includegraphics[width=\textwidth]{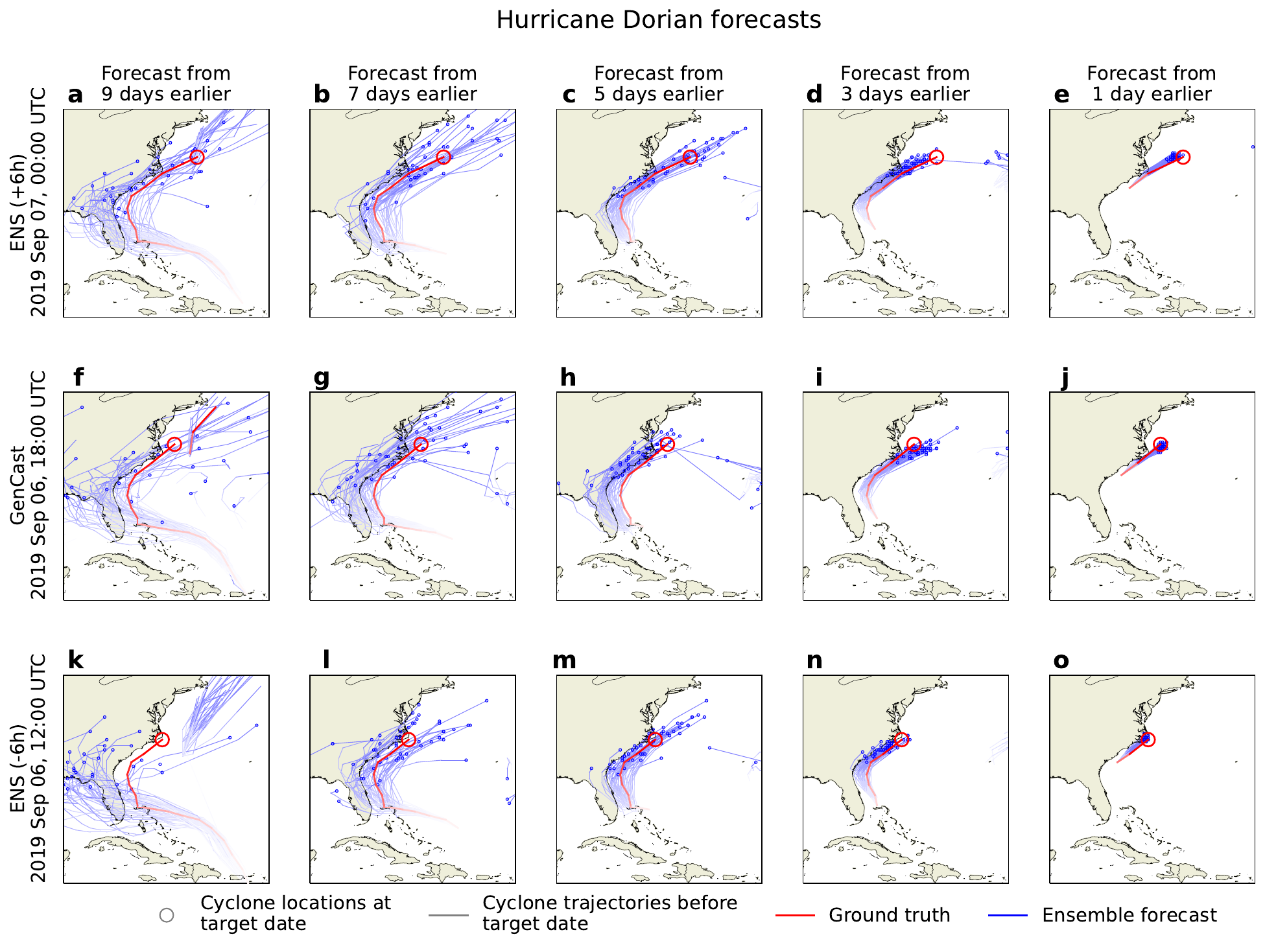}
    \caption{Visualisation of Hurricane Dorian trajectory forecasts.}
    \label{fig:app:visualization_cyclone_dorian}
\end{figure}

Here we provide further cyclone track visualisations. We picked some of the most extreme cyclones in 2019, based on several measures: the deadliest, the costliest, the strongest globally, and the strongest in the Atlantic basin. The deadliest cyclone was Cyclone Idai, which killed over 1500 people and caused a humanitarian crisis in Mozambique, Zimbabwe, and Malawi (\cref{fig:app:visualization_cyclone_idai}). The costliest cyclone was Typhoon Hagibis, with struck Japan and caused damages of 17.3 billion USD in 2019 (\cref{fig:app:visualization_cyclone_hagibis}). The strongest cyclone was Typhoon Halong, with a minimum barometric pressure of 905 hPa and maximum 1-minute sustained winds of 305 km/h (85 m/s) (\cref{fig:app:visualization_cyclone_halong}). The strongest cyclone in the Atlantic basin was Hurricane Dorian, which caused catastrophic damage in the Bahamas as a Category-5 hurricane before moving northwards along the coast of the United States, where it caused further damage and power outages (\cref{fig:app:visualization_cyclone_dorian}). 

In our visualisations, we set the validity dates as the day of landfall for Idai and Hagibis. For Halong, which did not make landfall, we set the validity date as the day the cyclone reached Category-5 status. For Dorian, we picked a time when the cyclone had weakened to Category-2 status, to also capture capture its effects on the East coast of the US.

Due to the 6-hour offset in initialisation and validity times between \ourmodel and ENS in our cyclone analysis (06/18 UTC and 00/12 UTC, respectively), in our visualisations we show ENS forecasts initialised both 6 hours before and 6 hours after \ourmodel's (with the same lead time). A 12-hour later initialisation time can give a model information on a cyclone that is 12 hours further into its development, which can have a material effect on cyclone forecasts in some cases. For example, when initialised 12 hours later, ENS performs substantially better at predicting Hurricane Dorian's curve northward along the eastern coast of the United States after passing the Bahamas (\cref{fig:app:visualization_cyclone_dorian}a vs \cref{fig:app:visualization_cyclone_dorian}k).

These visualisations are for illustration purposes only and have not been chosen as representative of differences between GenCast and ENS.
We refer the reader to our systematic REV evaluation for a rigorous and principled comparison of model performance (\cref{fig:panel_applications}b, \cref{fig:app:rev_cyclones_significance}, \cref{sec:app:cyclones:evaluation}).
We note that in some cases the trajectory of the main cyclone being visualised coincides with the trajectory of another cyclone nearby (e.g.~\cref{fig:app:visualization_cyclone_hagibis}a, \cref{fig:app:visualization_cyclone_dorian}a).
Finally, whereas tracks with short lead times branch out from an initial point, some tracks for long lead times \textit{start} from different places (e.g.~\cref{fig:app:visualization_cyclone_halong}a,f,k). This is because the forecasts are initialised before the cyclone started and the model is uncertain about whether cyclogenesis will occur and, if so, where it will occur.~\looseness=-1

\FloatBarrier

\subsection{Visualising forecasts of additional variables during Typhoon Hagibis}\label{sec:app:visualizations_hagibis}

\begin{figure}[H]
    \centering
    \includegraphics[width=\textwidth]{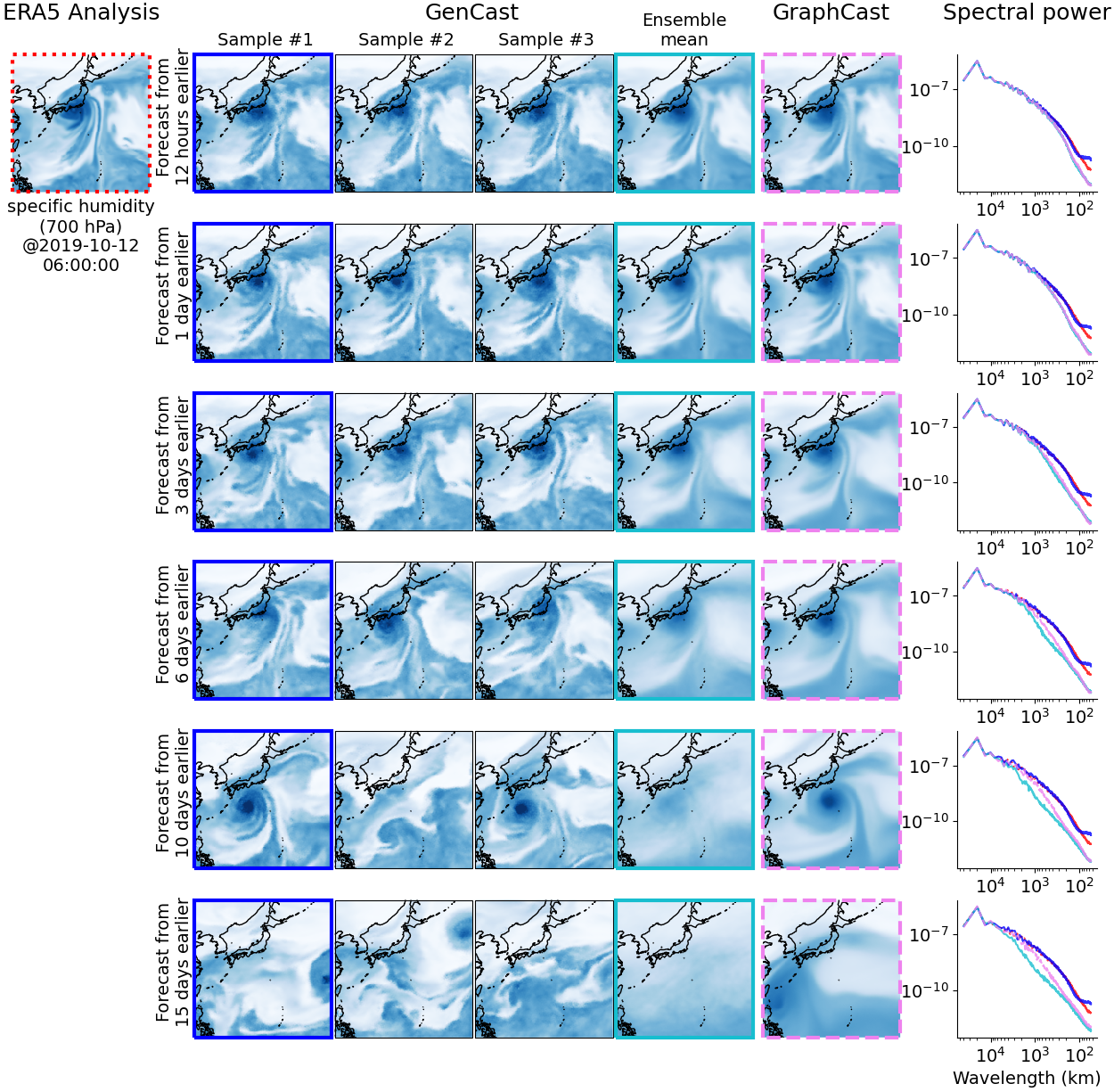}
    \caption{Visualisation of specific humidity at 700 hPa.}
    \label{fig:visualization_q700}
\end{figure}

\begin{figure}[H]
    \centering
    \includegraphics[width=\textwidth]{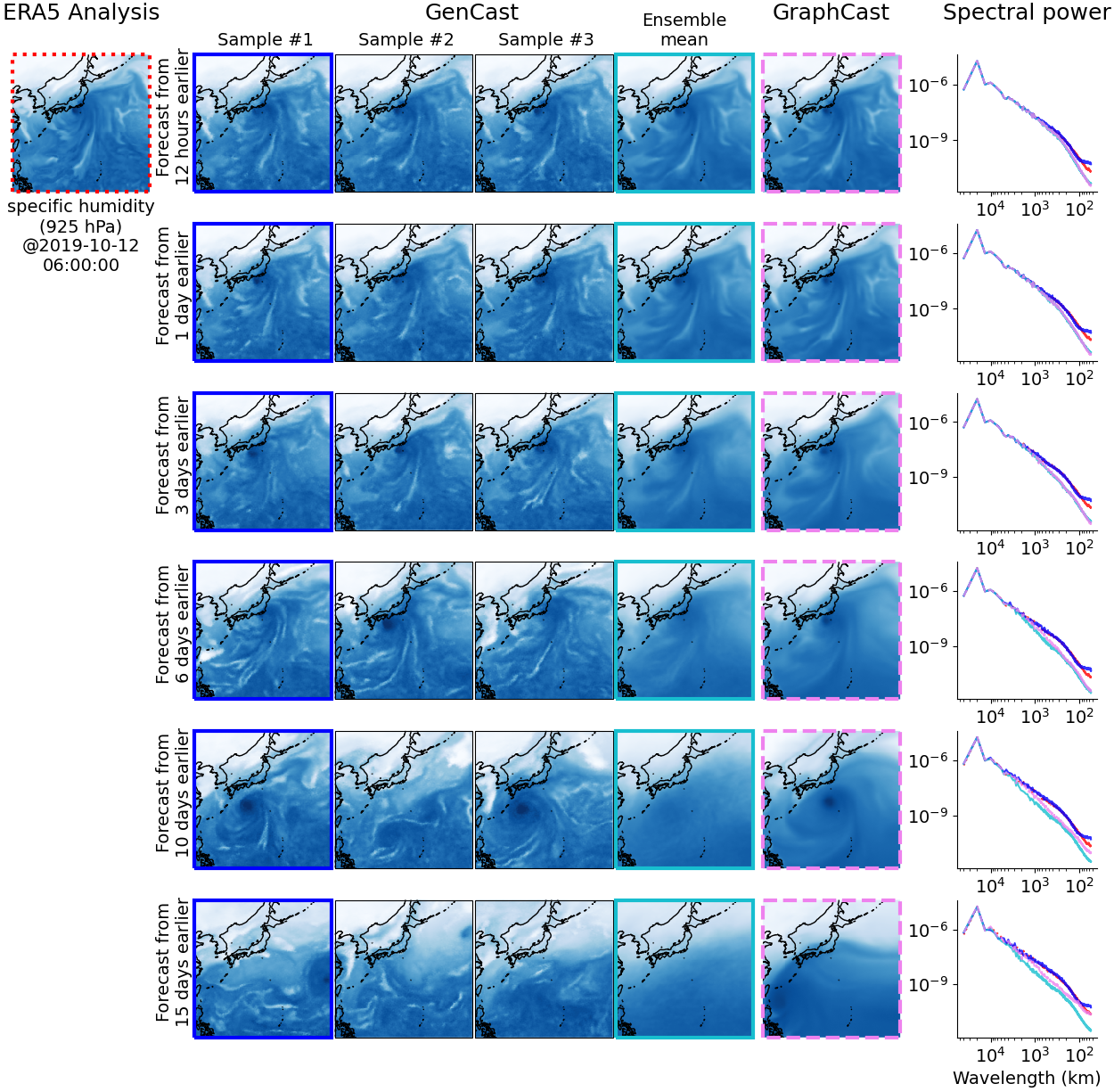}
    \caption{Visualisation of specific humidity at 925 hPa.}
    \label{fig:visualization_q925}
\end{figure}

\begin{figure}[H]
    \centering
    \includegraphics[width=\textwidth]{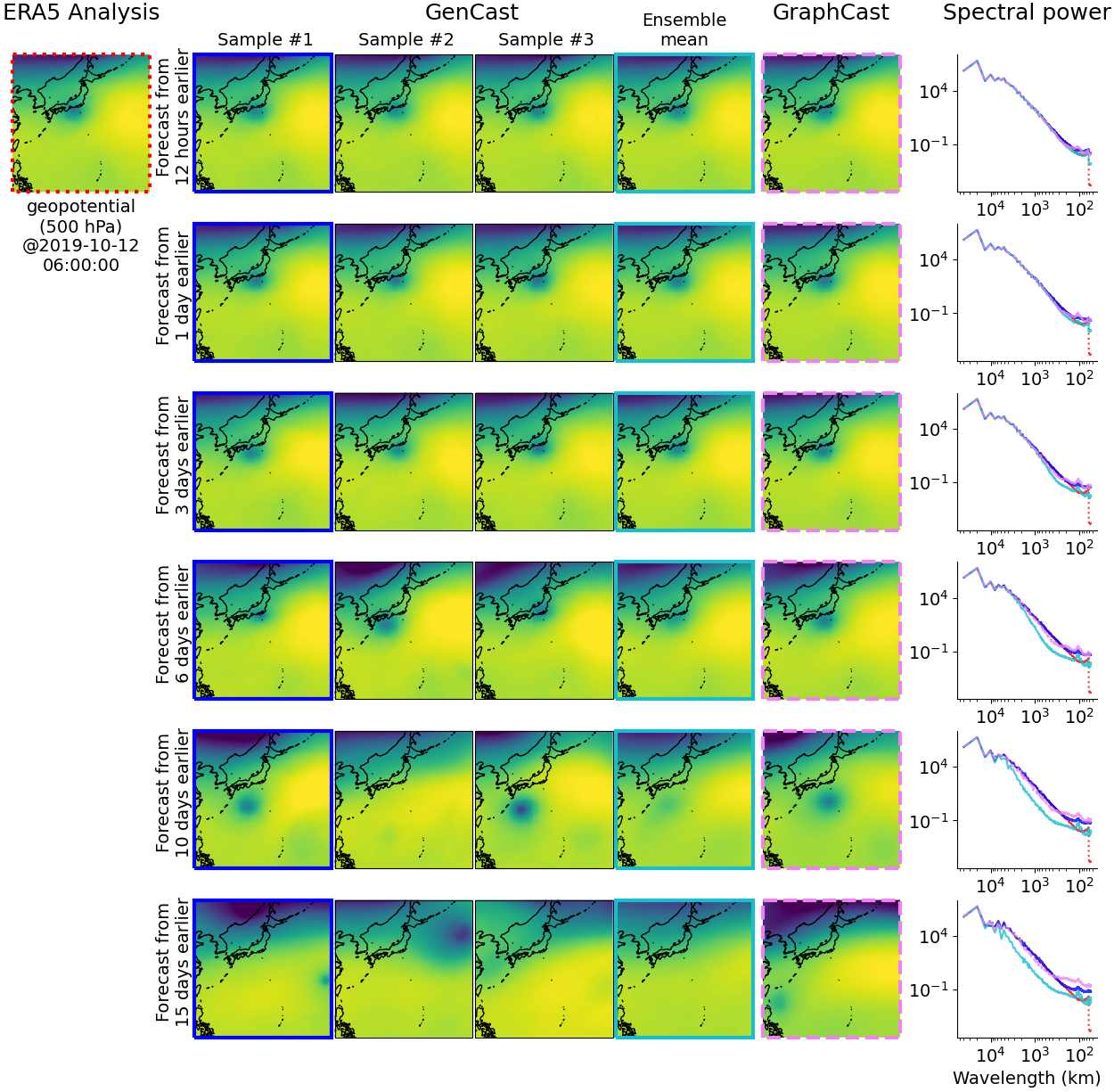}
    \caption{Visualisation of geopotential at 500 hPa.}
    \label{fig:visualization_z500}
\end{figure}

\begin{figure}[H]
    \centering
    \includegraphics[width=\textwidth]{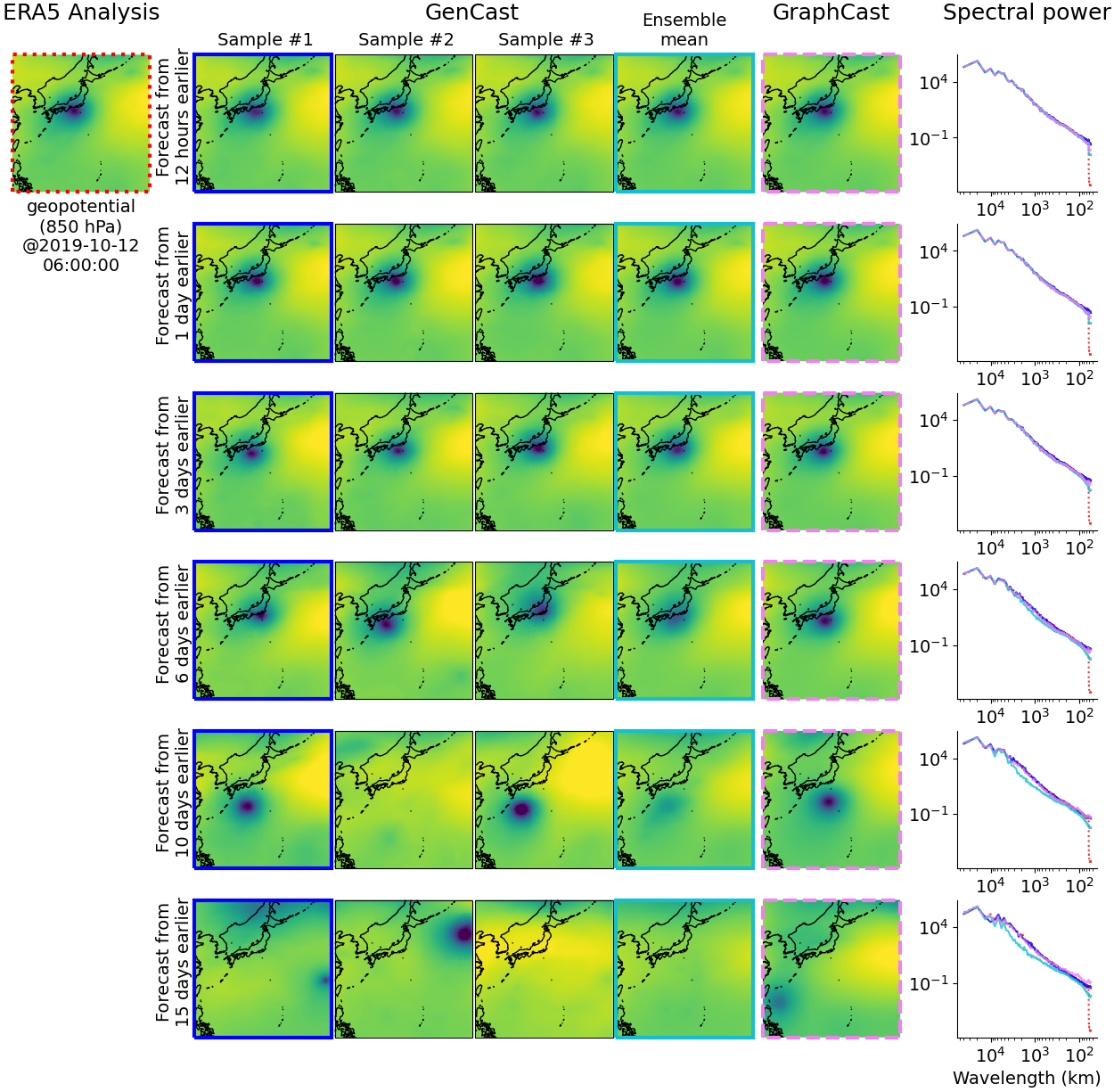}
    \caption{Visualisation of geopotential at 850 hPa.}
    \label{fig:visualization_z850}
\end{figure}

\begin{figure}[H]
    \centering
    \includegraphics[width=\textwidth]{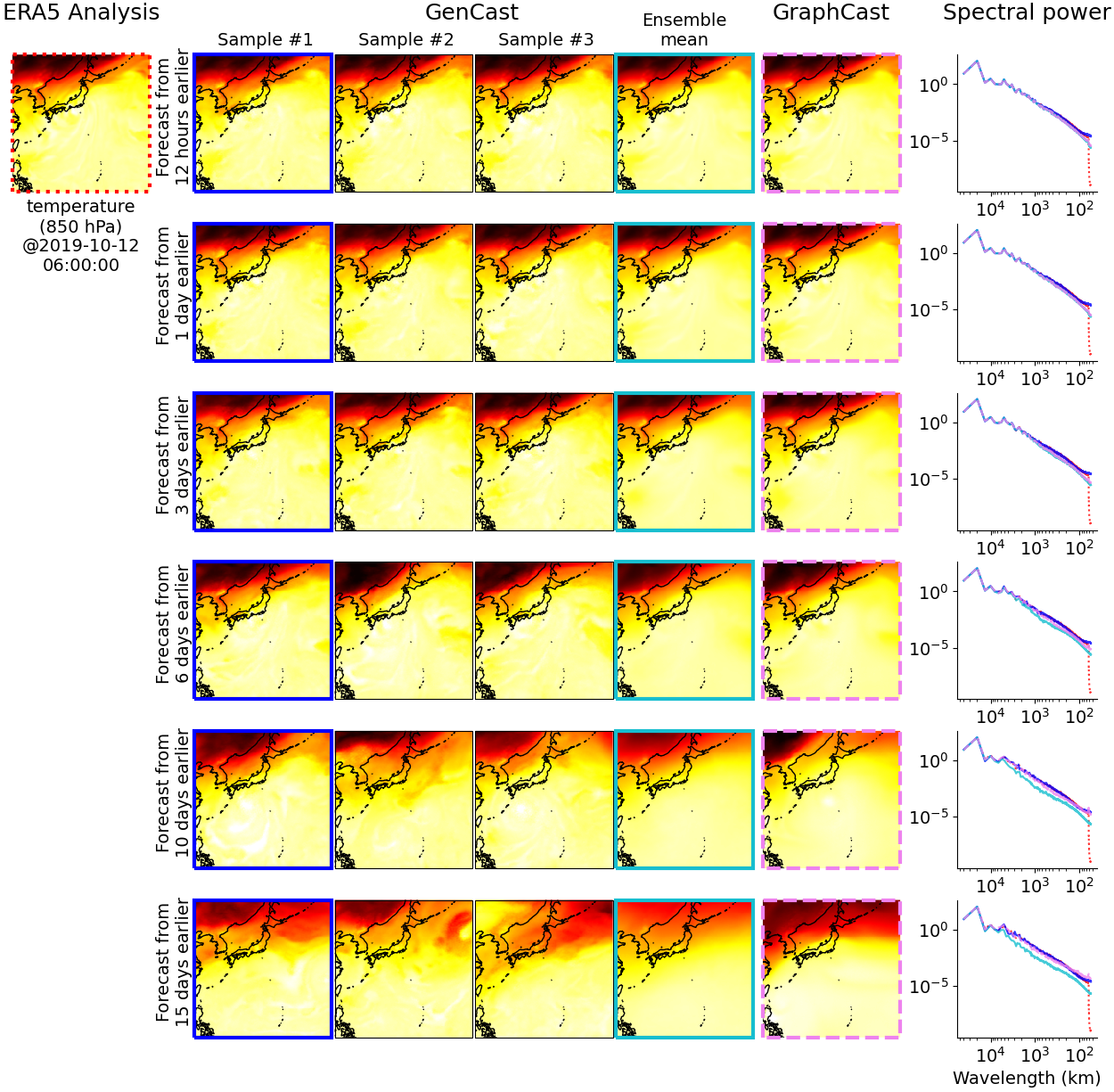}
    \caption{Visualisation of temperature at 850 hPa.}
    \label{fig:visualization_t850}
\end{figure}

\begin{figure}[H]
    \centering
    \includegraphics[width=\textwidth]{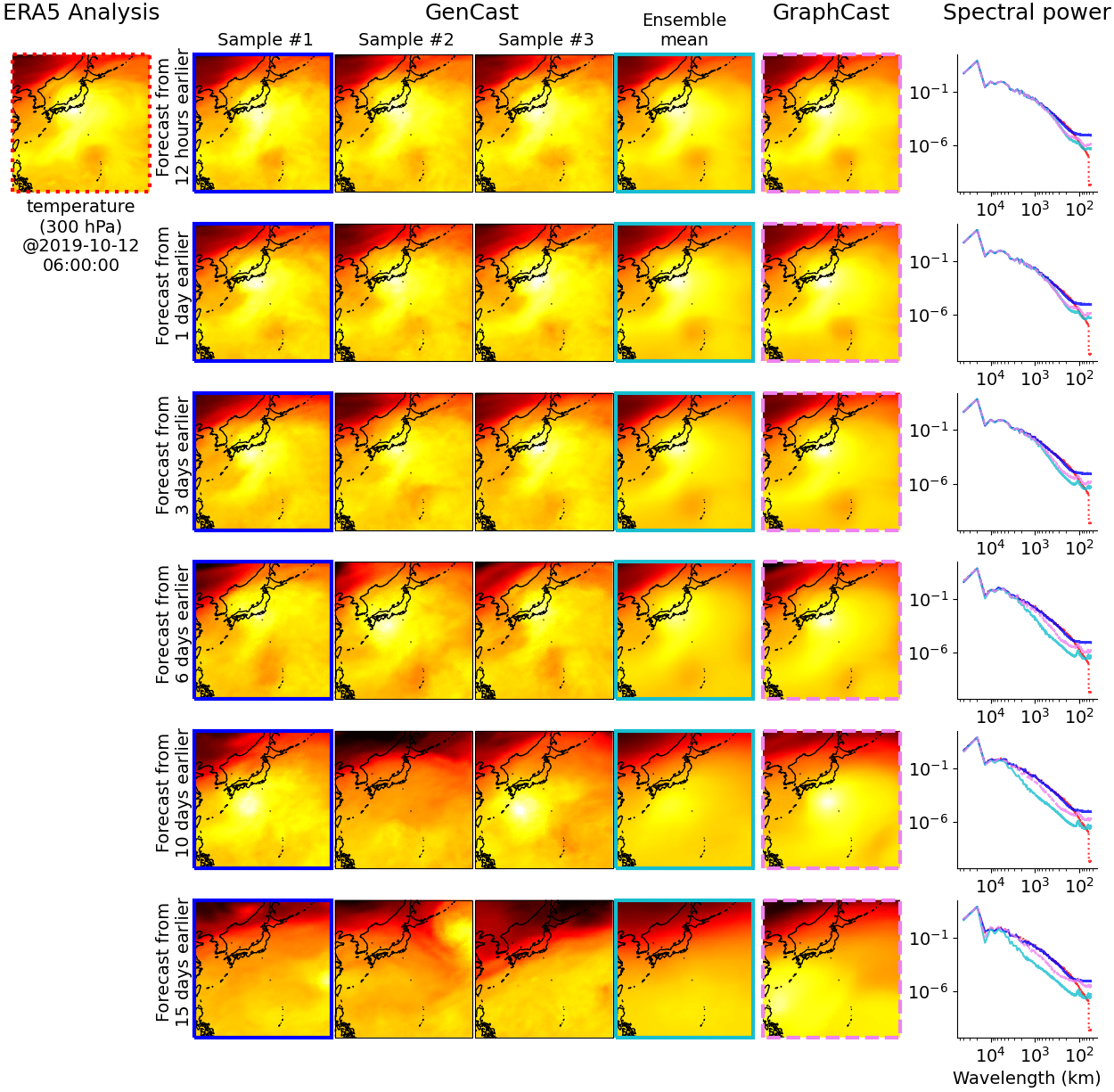}
    \caption{Visualisation of temperature at 300 hPa.}
    \label{fig:visualization_t300}
\end{figure}

\begin{figure}[H]
    \centering
    \includegraphics[width=\textwidth]{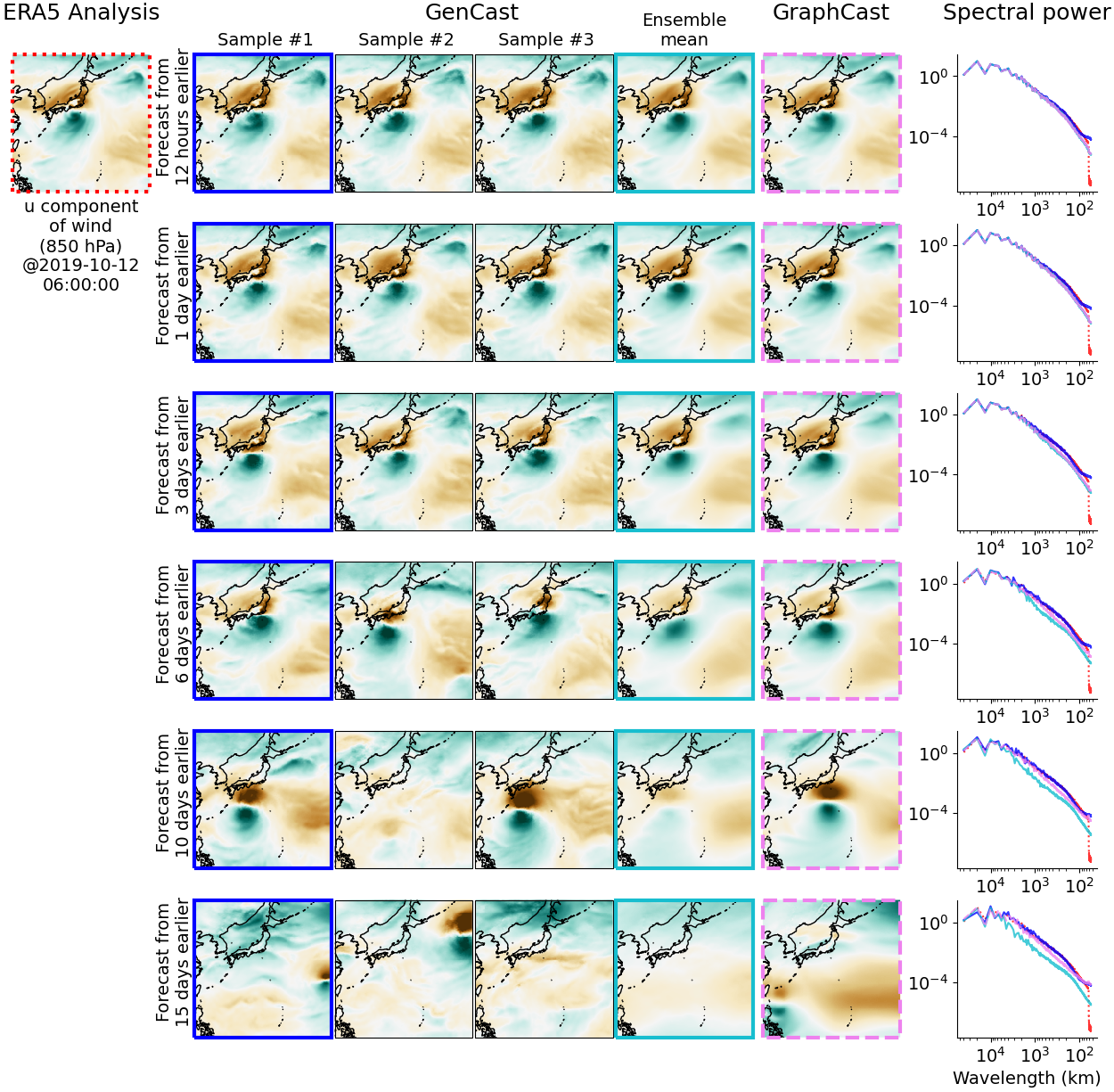}
    \caption{Visualisation of u component of wind at 850 hPa.}
    \label{fig:visualization_u850}
\end{figure}

\begin{figure}[H]
    \centering
    \includegraphics[width=\textwidth]{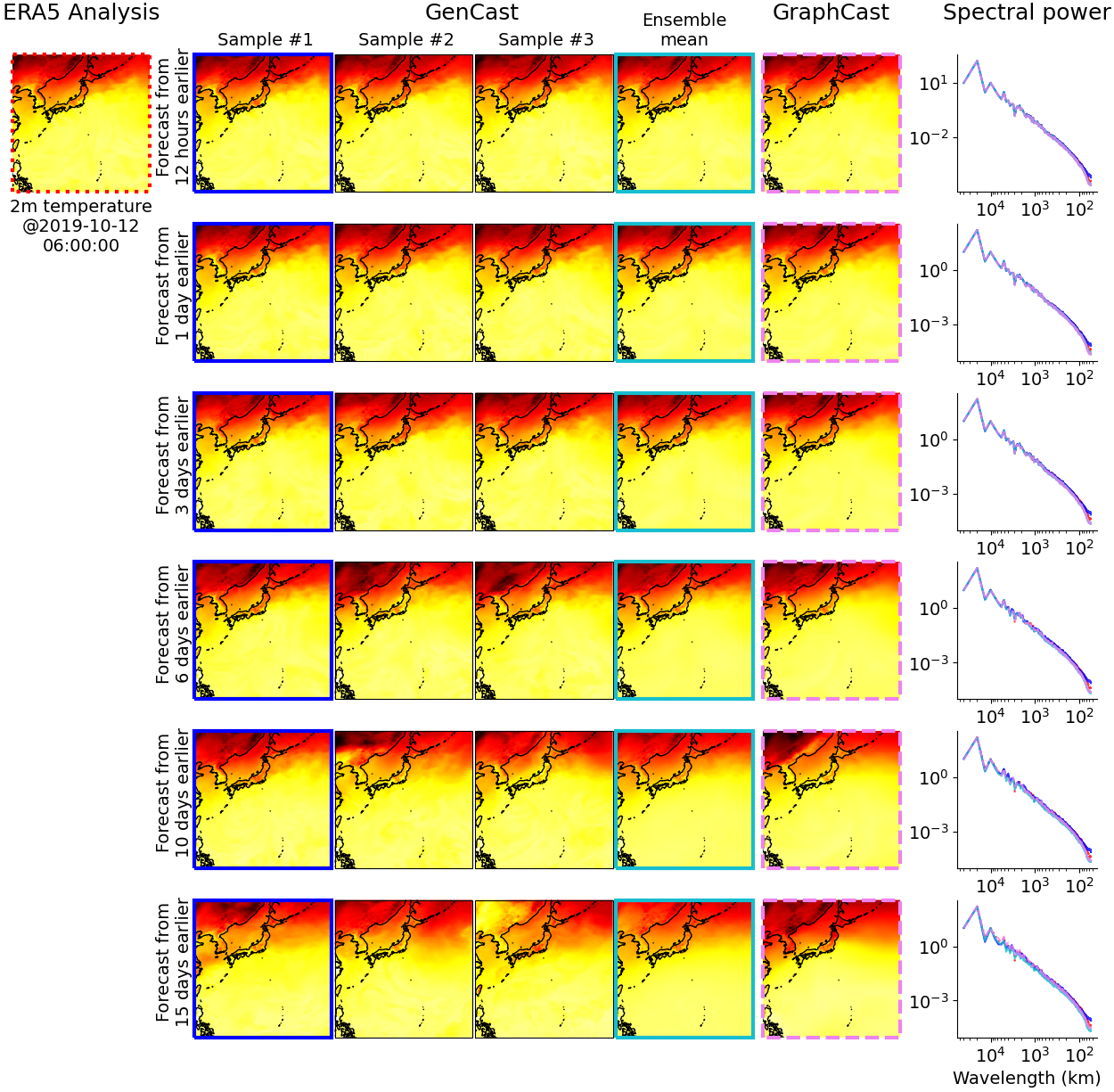}
    \caption{Visualisation of 2 meter temperature.}
    \label{fig:visualization_2t}
\end{figure}

\begin{figure}[H]
    \centering
    \includegraphics[width=\textwidth]{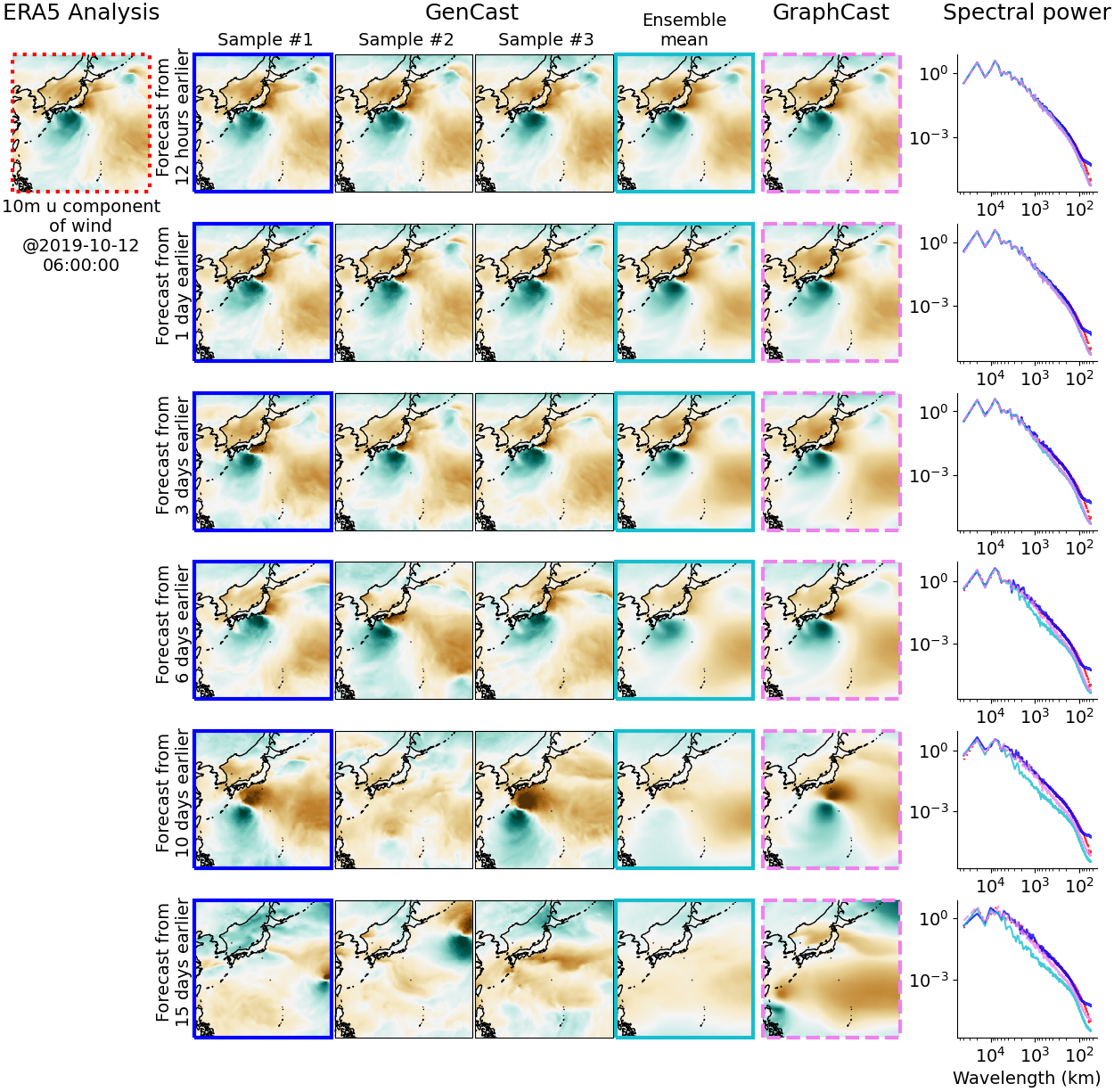}
    \caption{Visualisation of 10 meter u component of wind.}
    \label{fig:visualization_10u}
\end{figure}

\begin{figure}[H]
    \centering
    \includegraphics[width=\textwidth]{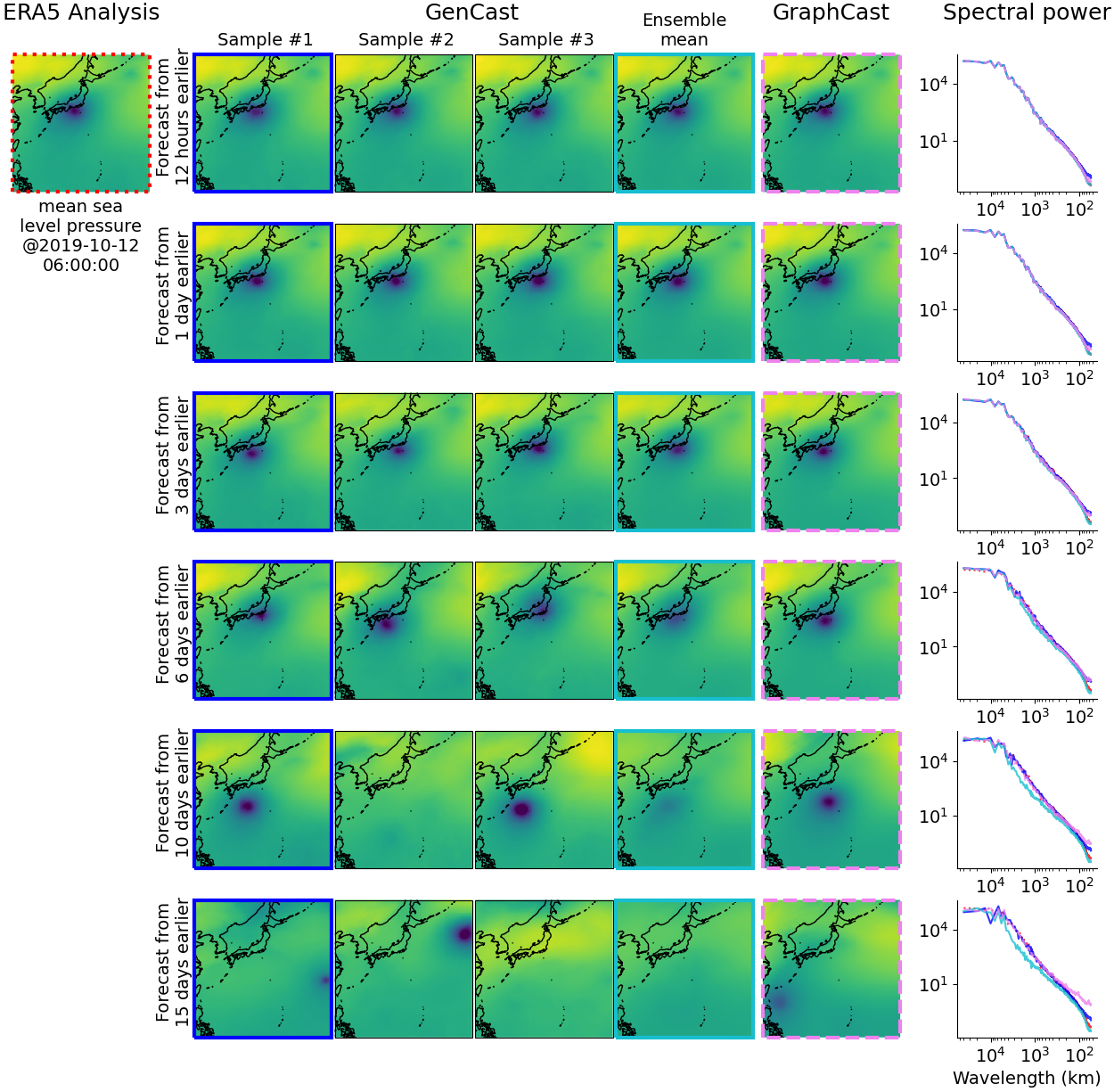}
    \caption{Visualisation of mean sea level pressure.}
    \label{fig:visualization_msl}
\end{figure}

We provide additional visualisations of Typhoon Hagibis similar to \cref{fig:visualization} for additional lead times, and 10 representative variables including specific humidity at 700 hPa (\cref{fig:visualization_q700}), specific humidity at 925 hPa (\cref{fig:visualization_q925}), geopotential at 500 hPa (\cref{fig:visualization_z500}), geopotential at 850 hPa (\cref{fig:visualization_z850}), temperature at 850 hPa (\cref{fig:visualization_t850}), temperature at 300 hPa (\cref{fig:visualization_t300}), u component of wind at 850 hPa (\cref{fig:visualization_u850}), 2 meter temperature (\cref{fig:visualization_2t}),  10 meter u component of wind (\cref{fig:visualization_10u}), and mean sea level pressure (\cref{fig:visualization_msl}).

\newpage

\subsection{Global forecasts}\label{sec:app:visualizations_global}

\begin{figure}[!ht]
    \centering
    \vspace*{-1.5cm}
    \includegraphics[width=\textwidth]{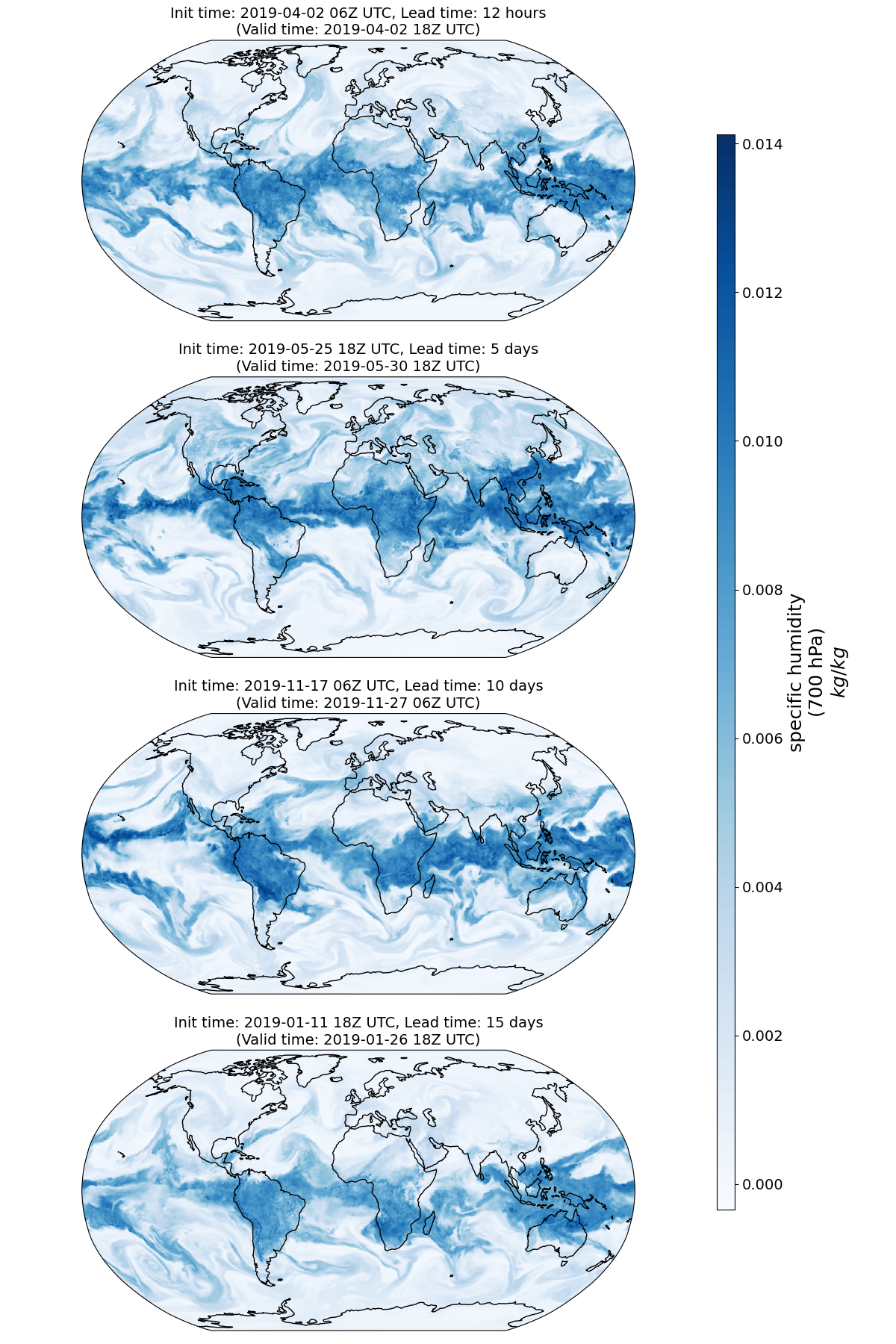}
    \caption{Visualisation of specific humidity at 700 hPa.}
    \label{fig:global_visualization_q700}
\end{figure}

\begin{figure}[!ht]
    \centering
    \vspace*{-1.5cm}
    \includegraphics[width=\textwidth]{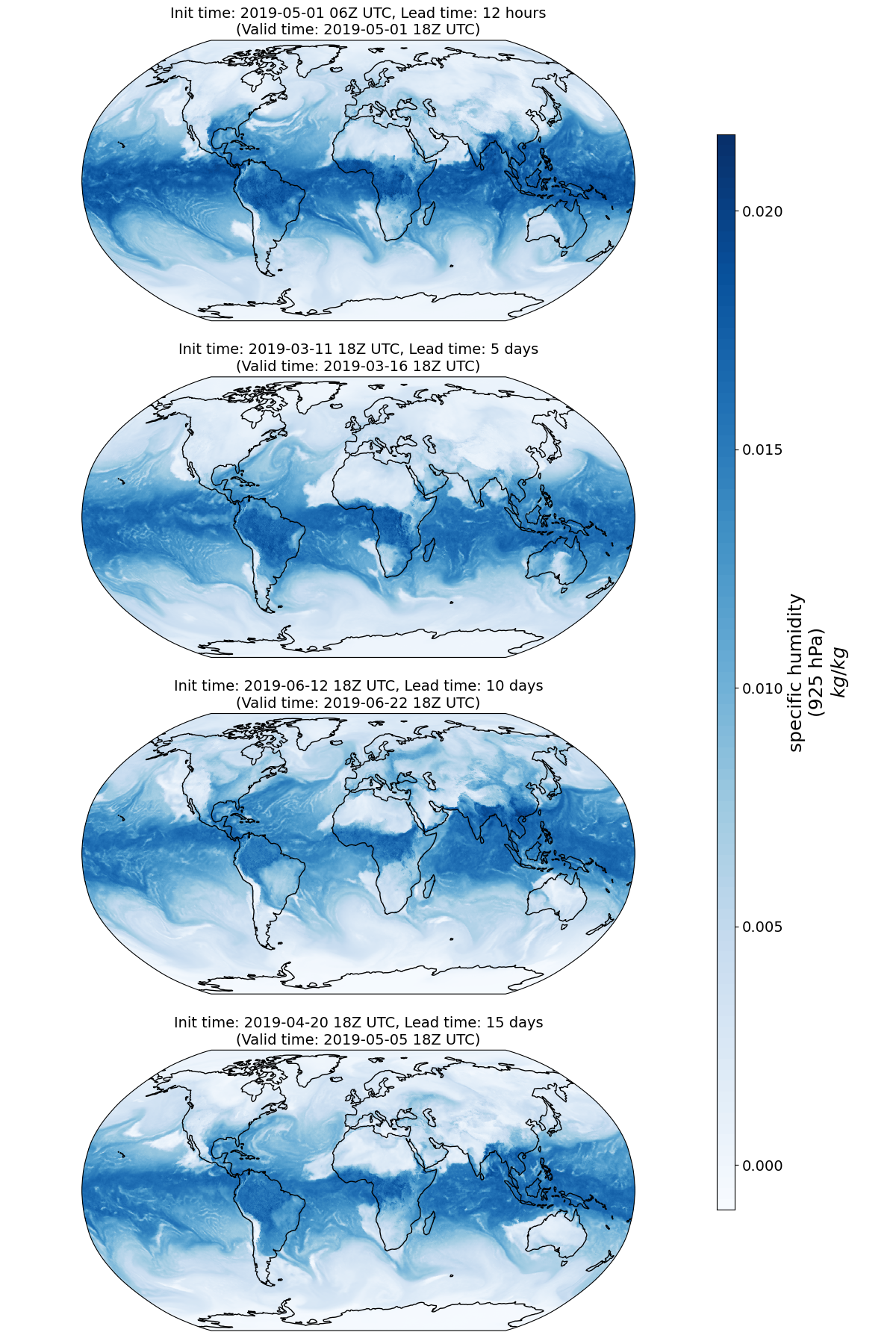}
    \caption{Visualisation of specific humidity at 925 hPa.}
    \label{fig:global_visualization_q925}
\end{figure}

\begin{figure}[!ht]
    \centering
    \vspace*{-1.5cm}
    \includegraphics[width=\textwidth]{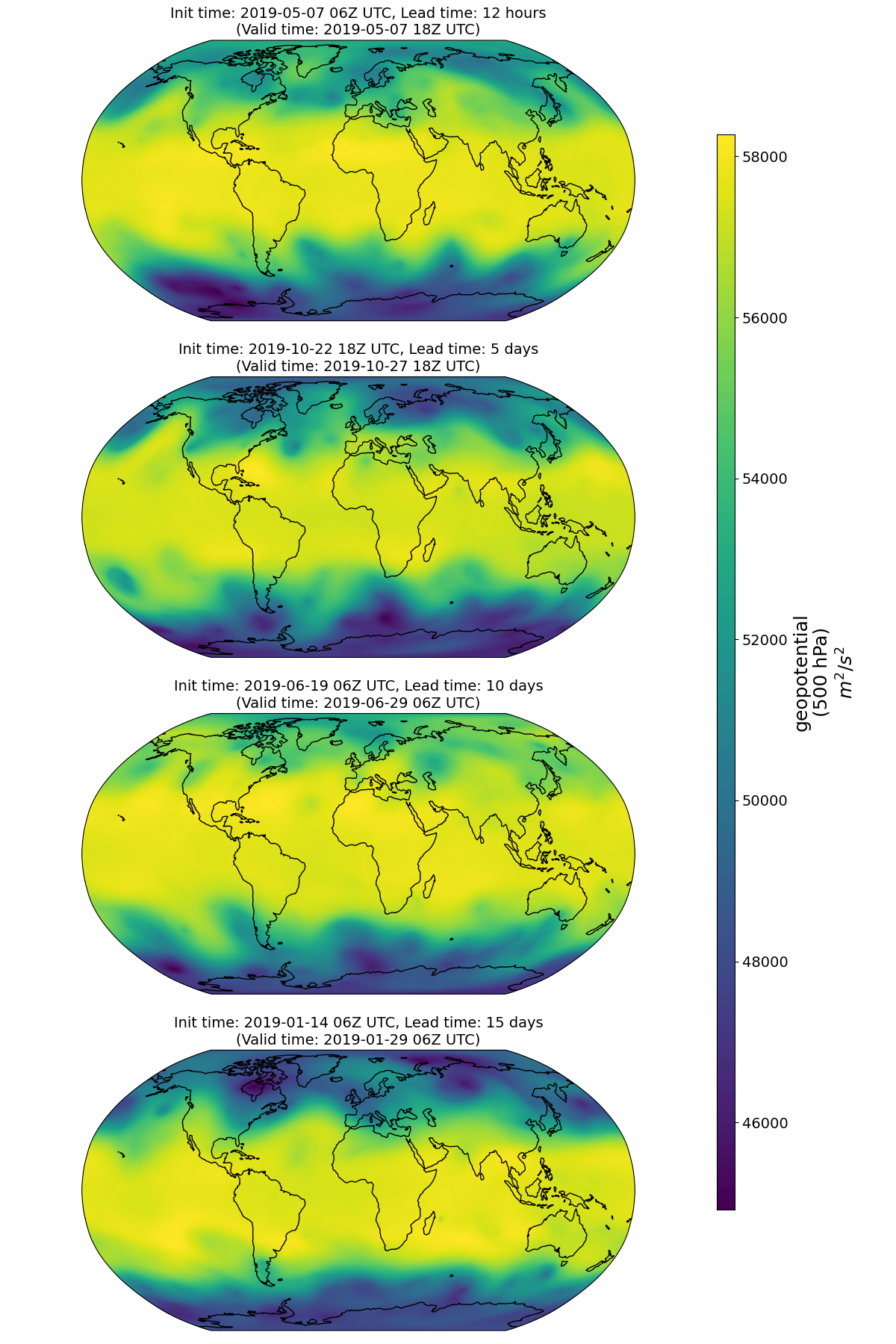}
    \caption{Visualisation of geopotential at 500 hPa.}
    \label{fig:global_visualization_z500}
\end{figure}

\begin{figure}[!ht]
    \centering
    \vspace*{-1.5cm}
    \includegraphics[width=\textwidth]{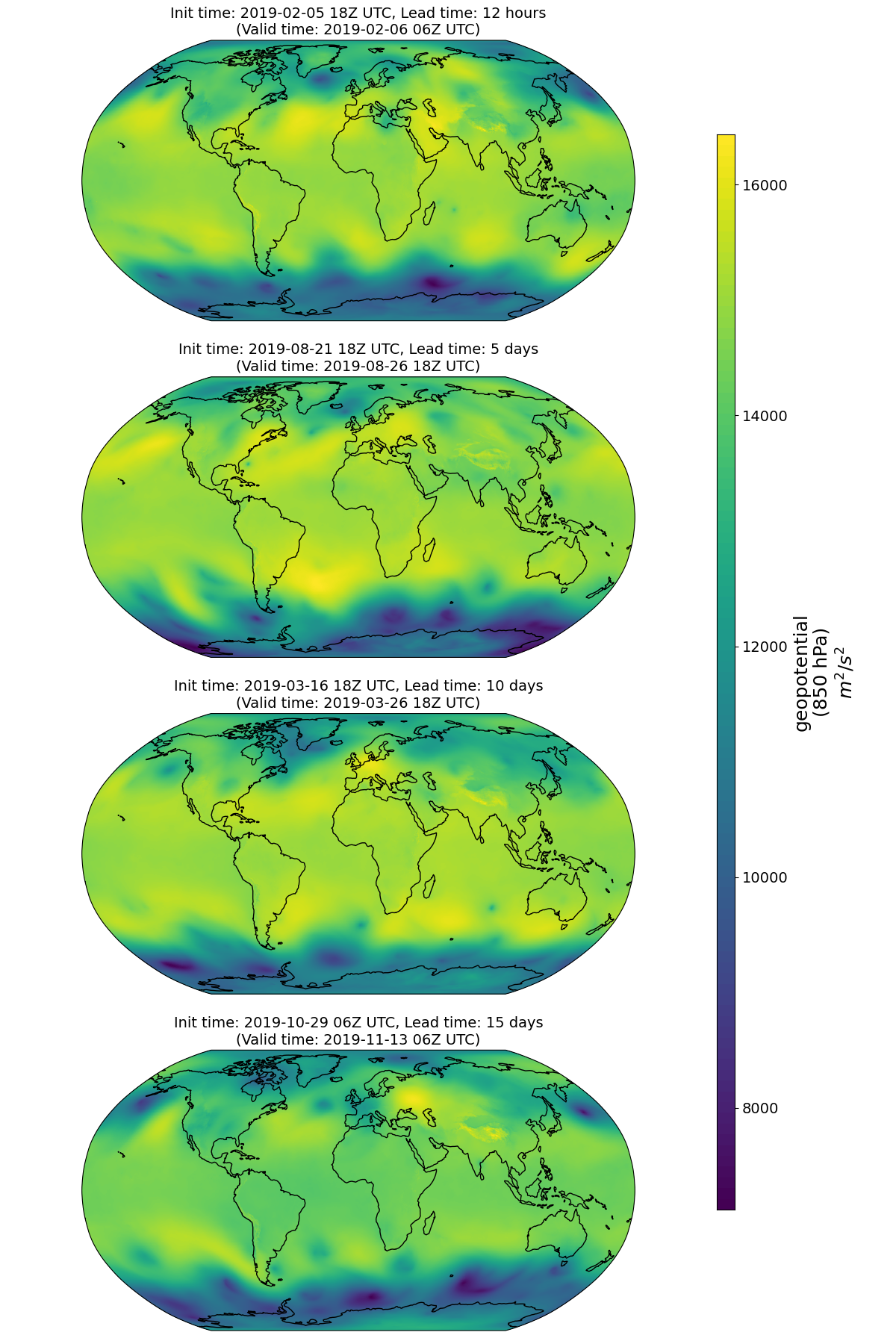}
    \caption{Visualisation of geopotential at 850 hPa.}
    \label{fig:global_visualization_z850}
\end{figure}

\begin{figure}[!ht]
    \centering
    \vspace*{-1.5cm}
    \includegraphics[width=\textwidth]{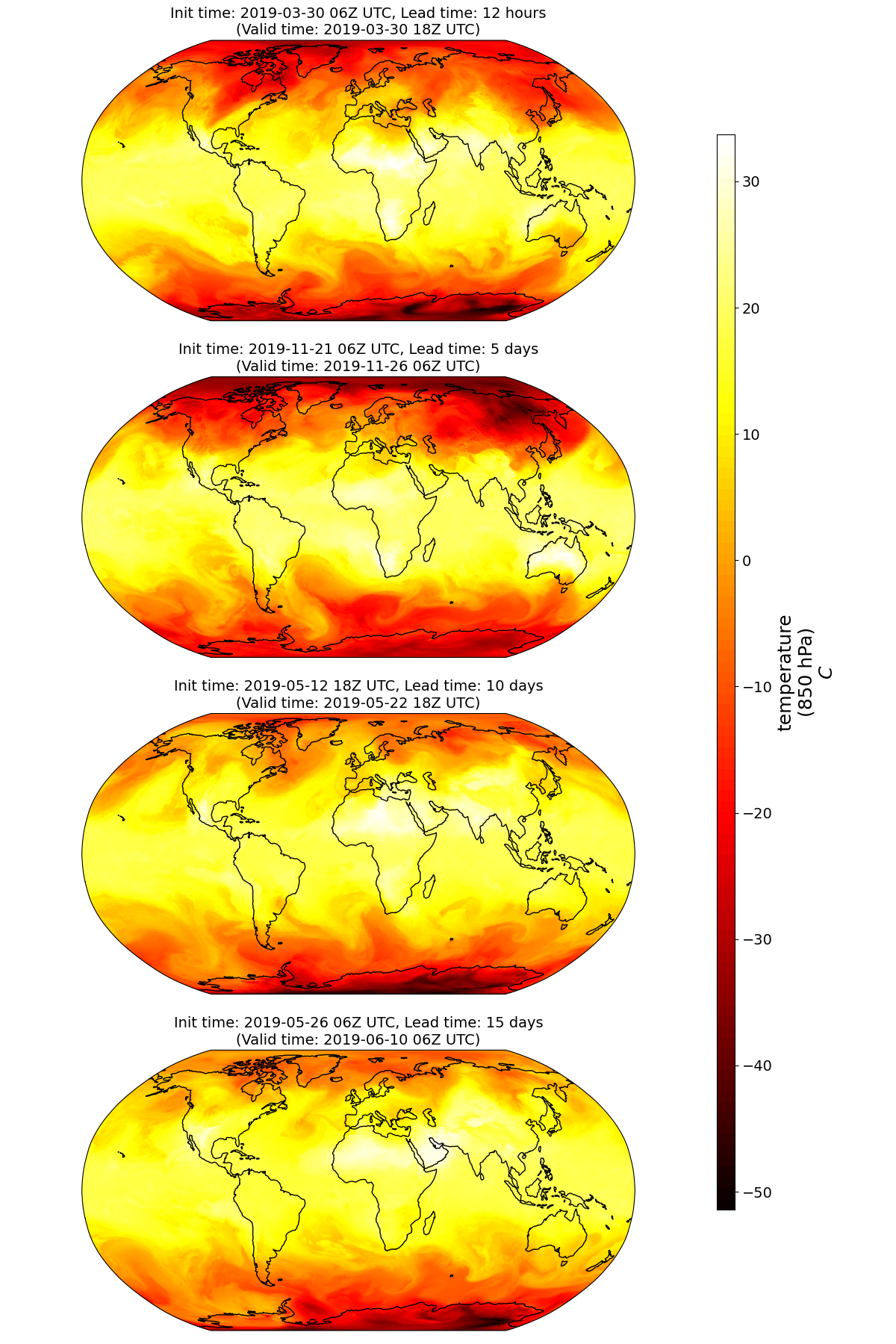}
    \caption{Visualisation of temperature at 850 hPa.}
    \label{fig:global_visualization_t850}
\end{figure}

\begin{figure}[!ht]
    \centering
    \vspace*{-1.5cm}
    \includegraphics[width=\textwidth]{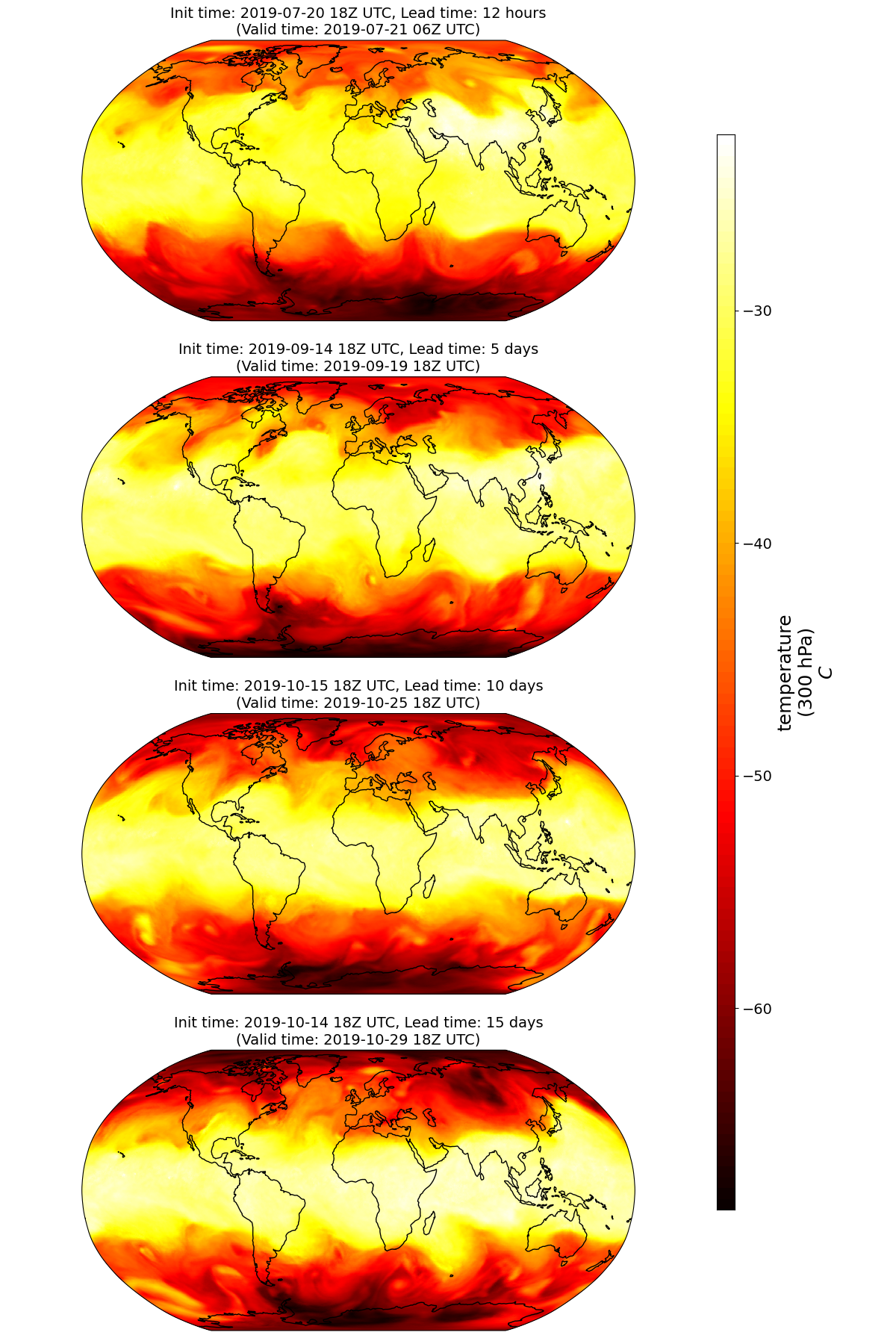}
    \caption{Visualisation of temperature at 300 hPa.}
    \label{fig:global_visualization_t300}
\end{figure}

\begin{figure}[!ht]
    \centering
    \vspace*{-1.5cm}
    \includegraphics[width=\textwidth]{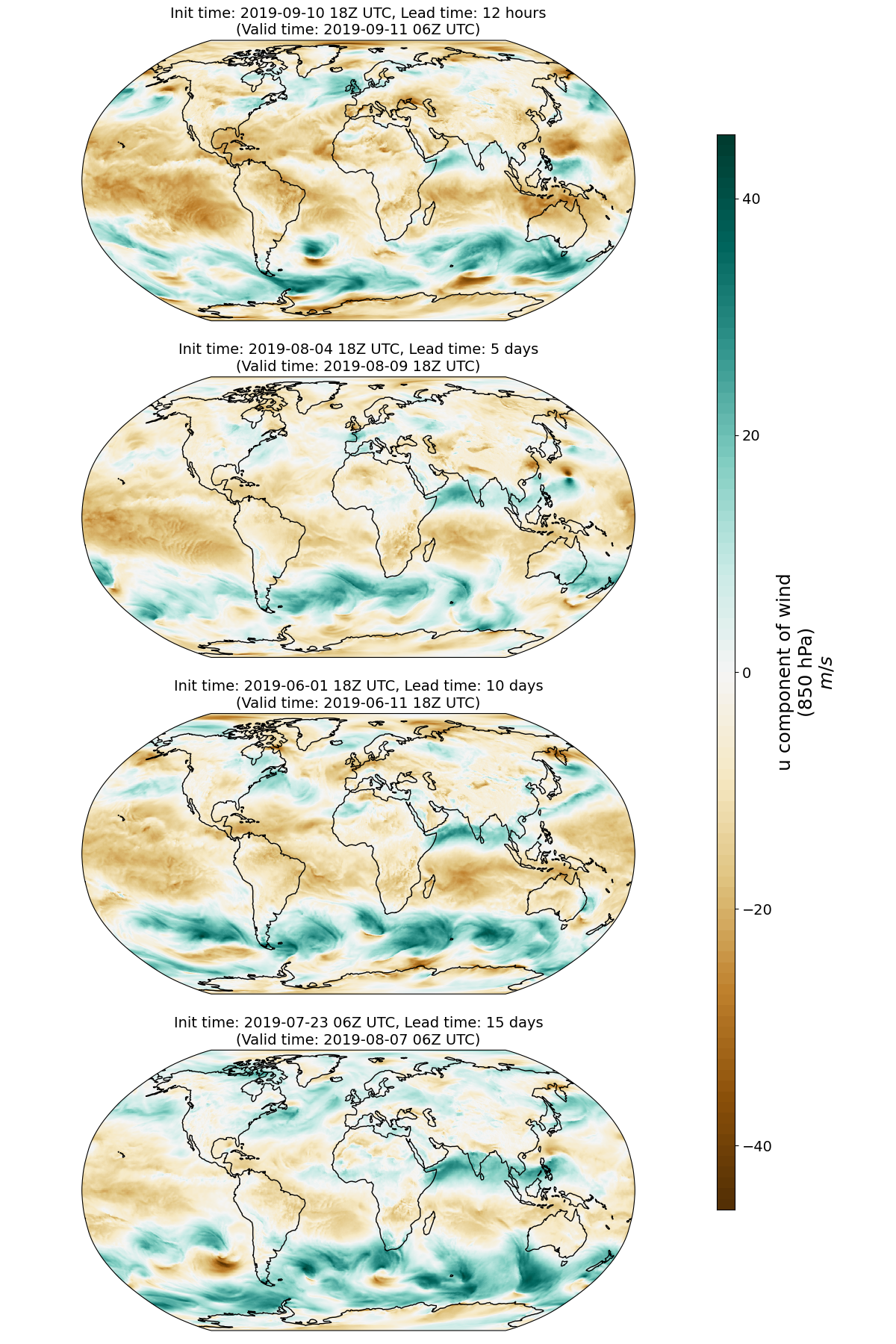}
    \caption{Visualisation of u component of wind at 850 hPa.}
    \label{fig:global_visualization_u850}
\end{figure}

\begin{figure}[!ht]
    \centering
    \vspace*{-1.5cm}
    \includegraphics[width=\textwidth]{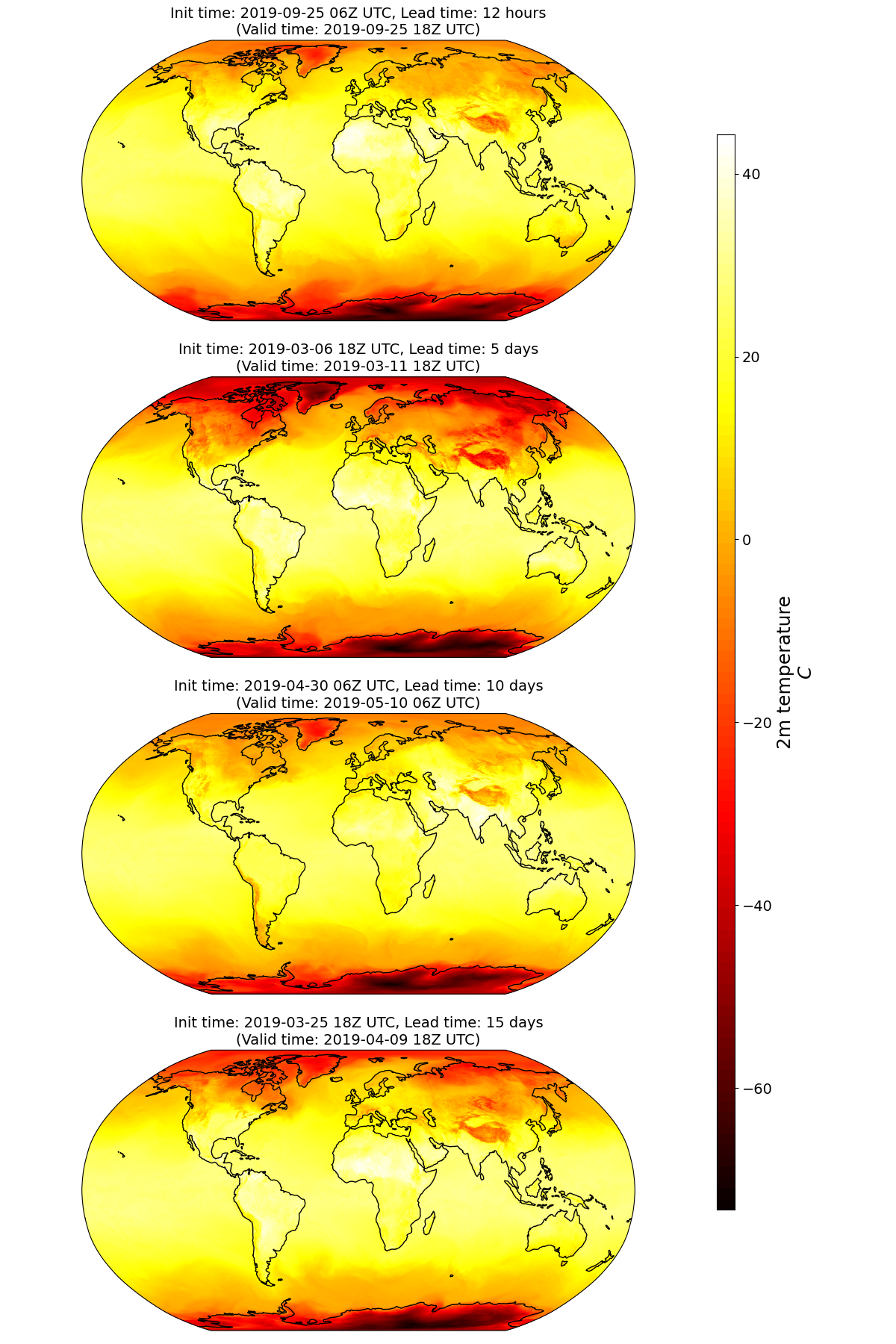}
    \caption{Visualisation of 2 meter temperature.}
    \label{fig:global_visualization_2t}
\end{figure}

\begin{figure}[!ht]
    \centering
    \vspace*{-1.5cm}
    \includegraphics[width=\textwidth]{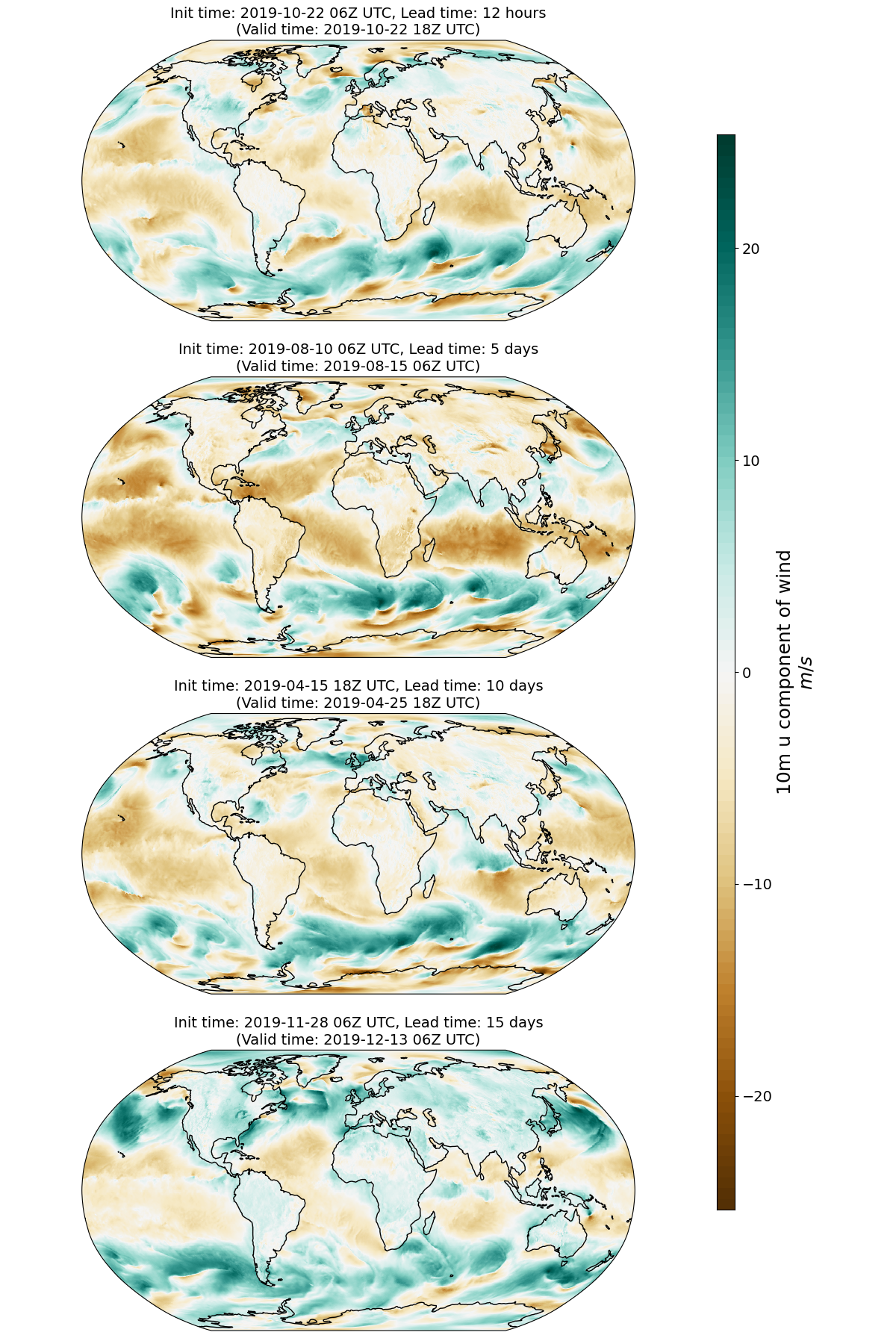}
    \caption{Visualisation of 10 meter u component of wind.}
    \label{fig:global_visualization_10u}
\end{figure}

\begin{figure}[!ht]
    \centering
    \vspace*{-1.5cm}
    \includegraphics[width=\textwidth]{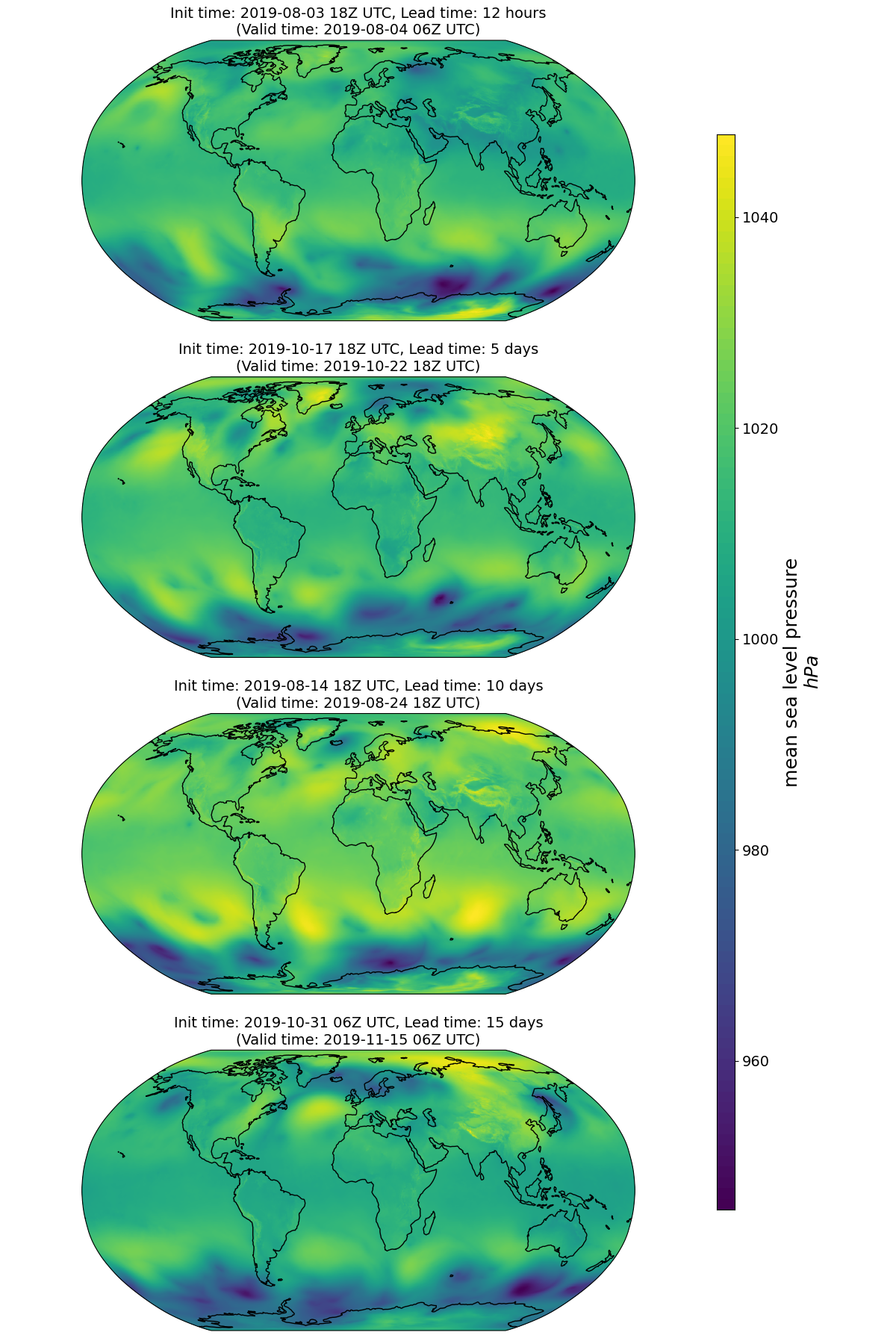}
    \caption{Visualisation of mean sea level pressure.}
    \label{fig:global_visualization_msl}
\end{figure}

We provide representative global visualisations for various lead times, and 10 representative variables including specific humidity at 700 hPa (\cref{fig:global_visualization_q700}), specific humidity at 925 hPa (\cref{fig:global_visualization_q925}), geopotential at 500 hPa (\cref{fig:global_visualization_z500}), geopotential at 850 hPa (\cref{fig:global_visualization_z850}), temperature at 850 hPa (\cref{fig:global_visualization_t850}), temperature at 300 hPa (\cref{fig:global_visualization_t300}), u component of wind at 850 hPa (\cref{fig:global_visualization_u850}), 2 meter temperature (\cref{fig:global_visualization_2t}), 10 meter u component of wind (\cref{fig:global_visualization_10u}), and mean sea level pressure (\cref{fig:global_visualization_msl}). For each variable-lead time combination we choose the 2019 initialization time with median CRPS error. This means that each figure shows a snapshot from a different forecast trajectory at each lead time. In each case we plot the first ensemble member.

\FloatBarrier
\newpage
\section{Author Contributions}

In alphabetical order by surname for each category of work:

Conceptualisation:  P.B., S.M., I.P., M.W.
Project management: P.B.,  R.L., I.P., J.S., M.W. 
Data Curation: P.B., A.E.K., I.P., A.S-G.
Model Development: F.A., A.E.K., R.L.,  D.M., I.P.,  A.S-G., M.W. 
Code: F.A., T.R.A.,  T.E., A.E.K., R.L., D.M., I.P., A.S-G., M.W.  
Experiments and Evaluation:	F.A., T.R.A., A.E.K., R.L.,  D.M., I.P., A.S-G., M.W.
Paper writing (original draft):	F.A., T.R.A., P.B., R.L., I.P.,  A.S-G., M.W.
Paper writing (reviewing and editing): F.A., T.R.A., P.B.,  A.E.K., D.M, S.M., I.P., A.S-G., M.W. 

\end{appendices}

\end{document}